\documentclass[10pt,twocolumn,letterpaper]{article}

\usepackage[pagenumbers]{cvpr}
\usepackage[T1]{fontenc}
\usepackage{xspace}
\usepackage{multirow}
\AtBeginDocument{%
  \crefname{table}{Table}{Tables}%
  \Crefname{table}{Table}{Tables}%
  \crefname{figure}{Fig.}{Figs.}%
  \Crefname{figure}{Fig.}{Figs.}%
}
\newcommand{\method}{SENSA\xspace}
\newcommand{\subjectstate}{Subject State\xspace}

\newcommand{\tableformat}{%
  \fontsize{8.5}{10.2}\selectfont
  \setlength{\tabcolsep}{4pt}%
  \renewcommand{\arraystretch}{1.12}%
}

\definecolor{linkblue}{rgb}{0.21,0.49,0.74}
\usepackage[pagebackref,breaklinks,colorlinks,allcolors=linkblue]{hyperref}

\title{SENSA: Visual-to-Brain Generation with Individualized\\
Subject-State Transfer and Adaptation}
\author{%
    \large
    Yihang Zou$^{1}$ \quad
    Chunmei Qing$^{1}$\thanks{Corresponding author.} \quad
    Ruixi Liang$^{1}$ \quad
    Haichuan Wei$^{1}$ \quad
    Junpeng Tan$^{2}$\\[3pt]
    \large
    Qifan Yang$^{3}$ \quad
    Wanxiang Luo$^{4}$ \quad
    Haodong Li$^{5}$ \quad
    Xiangmin Xu$^{1}$ \quad
    Yefeng Zheng$^{2}$\\[6pt]
    \normalsize $^{1}$South China University of Technology \quad
    $^{2}$Westlake University\\
    \normalsize $^{3}$National University of Singapore \quad
    $^{4}$Harbin Institute of Technology\\
    \normalsize $^{5}$The Chinese University of Hong Kong
}
\hypersetup{
    pdfauthor={Yihang Zou, Chunmei Qing, Ruixi Liang, Haichuan Wei,
        Junpeng Tan, Qifan Yang, Wanxiang Luo, Haodong Li,
        Xiangmin Xu, Yefeng Zheng}
}

\begin{document}
\maketitle
\begin{abstract}
Visual-to-brain encoding predicts neural responses to visual stimuli, supporting visual cognition research and neural decoding. However, challenges in disentangling shared from image-specific activity, together with inter-subject distribution shifts and representational heterogeneity, limit accurate encoding and cross-subject generalization with limited calibration data.
We introduce \method, a structured continuous neural response generation framework
that separates shared activity from image-dependent responses and
calibrates a compact subject state. An independent response prior captures shared activity, while complementary visual descriptors drive separable sensor-time bases
and a shared Response Refiner. Subject-Adaptive Factor Transport adjusts the spatiotemporal
expression of shared bases, while Source-Grounded Residual Memory predicts image-dependent corrections in source-response coordinates. Together, they personalize a frozen generator using limited paired support.
Experiments demonstrate that the generation backbone outperforms recent generators across three datasets, with image-identity perturbations confirming stimulus specificity. Using only 400 paired images, electroencephalography (EEG) and the Natural Scenes Dataset (NSD) calibration achieve residual Pearson correlations of 0.1091 and 0.2995, respectively, demonstrating effective brain-response generation and cross-subject adaptation.
\end{abstract}

\begin{figure}[t]
    \centering
    \includegraphics[width=\columnwidth]{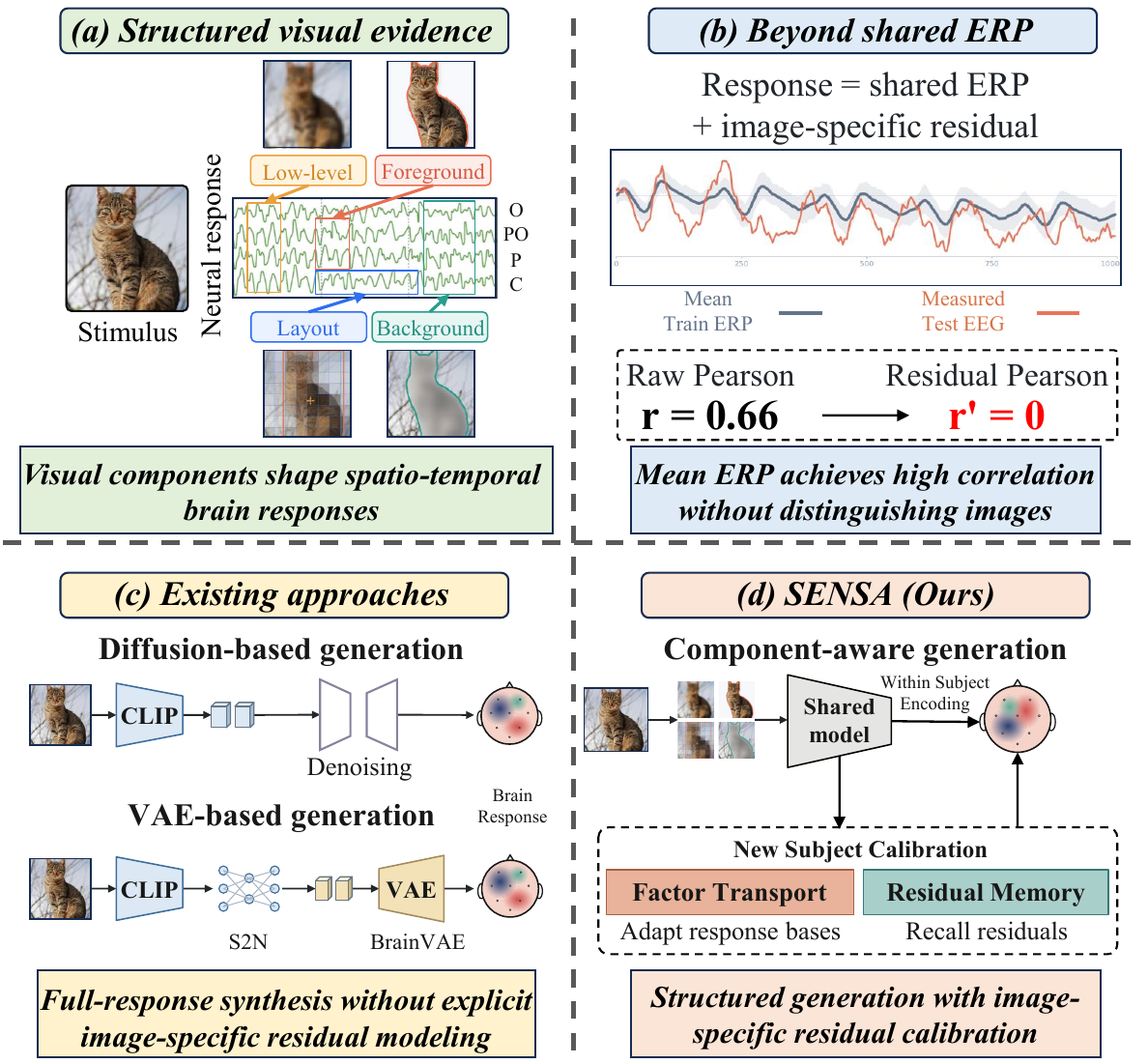}
    \caption{\textbf{Motivation and overview of SENSA.}
    (a) Complementary visual cues motivate structured response modeling;
    component-to-response arrows are schematic.
    (b) A training-mean ERP can match shared activity without image specificity.
    (c) Full-response synthesis motivates evaluating image-dependent prediction
    separately.
    (d) SENSA calibrates a frozen shared generator through Factor Transport
    and Residual Memory.}
    \label{fig:teaser}
\end{figure}

\section{Introduction}
\label{sec:intro}

Visual-to-brain encoding predicts neural responses to visual stimuli, establishing a quantitative mapping between visual information and neural representations that supports visual cognition research and neural decoding~\cite{naselaris2011encoding}. Image--neural response pairs provide explicit supervision and candidate response templates for neural decoding~\cite{kay2008identifying}, expand training data under limited neural recordings, and enable the investigation of shared neural representations underlying visual perception and mental imagery through controllable stimulus generation~\cite{concept2brain2026, Wadia2026SharedCode}. Thus, joint image--neural response generation may advance neural decoding from passive signal analysis toward active interrogation of cognitive representations.

These applications require predictions that distinguish images. Deep-feature encoders~\cite{yamins2014performance,guclu2015deep,styves2023brain} and generators including CLIP-guided diffusion, Concept2Brain, and SynBrain~\cite{xu2026image,concept2brain2026,mai2025synbrain} map images to responses (\cref{fig:teaser}(c)). However, full-response similarity can reflect stimulus-shared activity rather than image discrimination, as the mean event-related potential (ERP) example illustrates (\cref{fig:teaser}(b)). Encoding must recover image-dependent variation beyond this shared structure. This requires modeling image contributions across space and time~\cite{cichy2014resolving,graumann2022location}. Complementary low-level, foreground, background, and layout cues (\cref{fig:teaser}(a)) provide visual evidence for these response patterns.

Even accurate image-specific predictions must be expressed in the target individual's response space. Shared stimulus representations coexist with individual cortical topographies~\cite{haxby2011common,chen2015srm,haxby2020hyperalignment}; transferring visual information therefore need not transfer its measured response pattern. Target mean and scale describe response statistics, but do not identify how responses vary with images. Few-shot synthesis supports source reuse~\cite{mai2025synbrain}, while scarce target pairs constrain relearning this mapping. Calibration must retain shared visual knowledge while adapting its subject-specific expression and remaining image-dependent errors.

To this end, we introduce \method, combining structured generation with
individualized calibration as outlined in \cref{fig:teaser}(d).
An independent prior captures shared activity, while complementary visual
descriptors drive separable sensor-time bases and a shared Response Refiner.
Calibration must first align shared response structure to the target subject.
With the generator frozen, \emph{Subject-Adaptive Factor Transport (FT)}
uses a small paired support set to adjust the spatial mixing, timing,
and amplitudes of shared response bases.
The preceding stage aligns the overall neural response structure.
To further refine image-specific response details, we freeze FT and fit
\emph{Source-Grounded Residual Memory (RM)}, which maps image keys to
the remaining errors in fixed source-response coordinates, providing
image-conditioned corrections for unseen stimuli.
Our contributions are:
\begin{itemize}
    \item A structured adaptive continuous neural signal generator is employed to restore the specific image-stimulated brain activity that is masked due to shared responses. By combining independent response priors, complementary visual descriptors, and separable response bases, the framework enables the synthesis of additional electroencephalography (EEG) responses paired with images.
    \item A progressive self-calibration strategy can be employed to address the inter-subject response differences while keeping the generator frozen: FT can adapt the overall structure of the shared neural response signal, and RM predicts the remaining image-dependent residual errors in the source-response coordinate system. The generated responses can then be used to complete EEG and enhance decoding of real EEG.
    \item We evaluate on four EEG, magnetoencephalography (MEG), and functional magnetic resonance imaging (fMRI) benchmarks, probing stimulus specificity and calibration mechanisms through image identity perturbation and graded memory key corruption, and use the generated response templates to complete EEG and enhance decoding of real neural responses, demonstrating their downstream utility.
\end{itemize}

\section{Related Work}
\label{sec:related}

\paragraph{Structured visual encoding.}
Feature-weighted receptive fields and factorized readouts separate
feature selectivity from spatial pooling~\cite{styves2018fwrf,klindt2017neural},
making the preferred visual content and its receptive-field location
separately interpretable. Brain-optimized networks further learn
representations from neural supervision~\cite{styves2023brain}.
Scene statistics, surface geometry, and object location explain complementary response
properties~\cite{stansbury2013scene,lescroart2019surfaces,cichy2017scene,graumann2022location}.
These findings motivate retaining multiple visual cues, but selecting
features and pooling locations leaves their composition across sensors and
time to be specified. SENSA addresses this response construction with
separable sensor-time bases driven by complementary descriptors. An
independent prior captures shared activity, assigning the visual pathway
the distinct role of predicting image-dependent variation.

\paragraph{Visual-to-brain generation.}
CLIP-guided diffusion aligns visual conditioning with neural denoising
through cross-attention~\cite{xu2026image}. Concept2Brain maps CLIP concepts
into an EEG variational latent space~\cite{concept2brain2026}.
SynBrain models fMRI variability under visual semantic
constraints~\cite{mai2025synbrain}. NeuroFlow and Mind-Omni broaden generation
to unified encoding and decoding~\cite{mai2026neuroflow,lu2026mindomni}.
These approaches connect visual semantics to response synthesis, but
full-response fidelity and semantic consistency do not quantify how much
image-specific variation is recovered beyond shared activity. SENSA
explicitly decomposes these components for continuous prediction of
repetition-averaged responses. Residual scores and image-identity
perturbations then test whether the predicted variation depends on the
correct stimulus, complementing full-response evaluation.

\paragraph{Shared representations and subject adaptation.}
Functional alignment links common stimulus representations to
subject-specific topographies~\cite{haxby2011common,chen2015srm,haxby2020hyperalignment}.
Anatomical conditioning parameterizes responses by
location~\cite{chen2025nrf}, while MindEye2 transfers shared representations
for brain-to-image decoding~\cite{scotti2024mindeye2}.
EEG alignment corrects distribution shifts~\cite{zanini2018riemannian,he2020euclidean};
test-time training and entropy minimization adapt models using unlabeled
inputs~\cite{sun2020ttt,wang2021tent}. For visual-to-brain transfer,
response statistics can be estimated without image pairing, but they do
not identify which target response an image should evoke. SENSA separates
these information sources: response-only supports estimate target
statistics; paired supports fit FT to shared response structure and RM
to remaining image-conditioned errors. This preserves the shared generator
while calibrating its predictions to individual subjects.

\section{Method}
\label{sec:method}

\begin{figure*}[t]
    \centering
    \includegraphics[width=\textwidth]{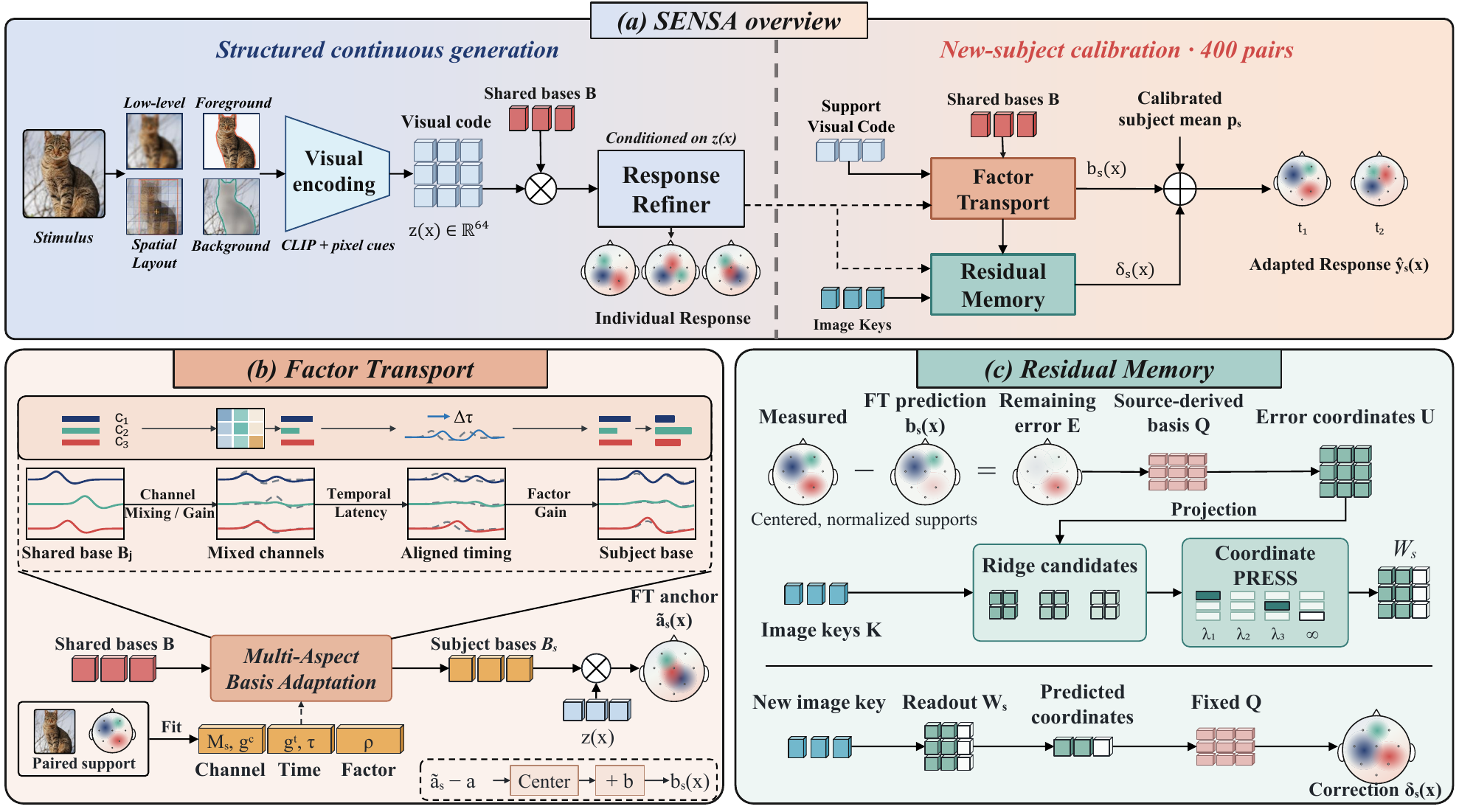}
    \caption{\textbf{SENSA architecture for sensor-time responses.}
    (a) Visual coefficients weight shared response bases; the Response Refiner
    yields the population prediction.
    (b) FT adapts channel mixing, timing, and factor amplitudes.
    (c) RM maps image keys to error coordinates through coordinate-wise
    PRESS-selected Ridge, including zero correction.
    The target support mean $p_s$ is added separately.
    Waveforms and transformations are schematic.}
    \label{fig:architecture}
\end{figure*}

\subsection{Task and target information}

To render shared visual knowledge in individual response spaces,
SENSA predicts the repetition-averaged response to image $x$ for subject $s$
in modality $m\in\{\mathrm{EEG},\mathrm{MEG},\mathrm{fMRI}\}$:
\begin{equation}
    \hat y_s^m(x)=G_m(x;S_s),\qquad S_s=A_m(\mathcal D_s).
    \label{eq:sensa}
\end{equation}
Here $G_m$, $A_m$, and $S_s$ denote the generator, calibration procedure, and
\subjectstate, respectively.
EEG/MEG use $C$ sensors and $T$ samples, $y_s\in\mathbb R^{C\times T}$;
fMRI uses $y_s\in\mathbb R^{V_s}$ for $V_s$ native voxels.
Response-only support $\mathcal D_s^{r}=\{y_{s,i}\}_{i=1}^{K}$ gives
$p_s=K^{-1}\sum_i y_{s,i}$ and residual root-mean-square (RMS) magnitude
$\sigma_s$ for calibration.
Paired support $\mathcal D_s^{p}=\{(x_i,y_{s,i})\}_{i=1}^{K}$
additionally fits FT parameters $\Theta_s^{\mathrm{FT}}$ and RM readout $W_s$.
The sensor state is
$S_s=(p_s,\Theta_s^{\mathrm{FT}},\overline\Delta_s,\bar\kappa_s,W_s)$,
with FT correction mean $\overline\Delta_s$ and key mean
$\bar\kappa_s$. Shared weights and bases stay fixed.
Below, we omit $m$ for sensors.

\subsection{Structured continuous generation}

\paragraph{Visual encoding.}
To retain complementary visual cues, descriptors $f_g(x)$ are indexed by
$\mathcal G=\{\mathrm{spatial},\mathrm{region},\mathrm{blur},\mathrm{low}\}$:
patch grids, joint foreground/background pooling, multi-scale blur
features, and pixel-level color, contrast, and Gabor statistics, respectively.
Frozen CLIP ViT-L/14~\cite{radford2021clip} layers 8, 12, and 16 encode
$g\in\{\mathrm{spatial},\mathrm{region},\mathrm{blur}\}$;
$f_{\mathrm{low}}$ bypasses CLIP. Projectors $P_g$ and learned gains
$\alpha_g$ fuse descriptors into 64 response coefficients:
\begin{equation}
    z(x)=\frac{1}{\sqrt{4}}\sum_{g\in\mathcal G}\alpha_gP_g f_g(x)
    \in\mathbb R^{64}.
    \label{eq:visual-code}
\end{equation}
\paragraph{Shared bases.}
To express visual content through reusable response structure, each
coefficient $z_j(x)$ weights a basis $B_j\in\mathbb R^{C\times T}$:
\begin{equation}
    a(x)=\sum_{j=1}^{64}z_j(x)B_j,\qquad
    B_j=\beta_j u_jv_j^\top.
    \label{eq:continuous-factors}
\end{equation}
Learned unit-norm spatial $u_j\in\mathbb R^C$ and temporal
$v_j\in\mathbb R^T$ factors have amplitudes $\beta_j$.
An independent prior captures shared activity; the bank supplies FT's
deformable structure and RM's fixed response coordinates.

\paragraph{Response Refiner.}
To expand the anchor's fixed linear span, the Response Refiner adds
a visual-conditioned correction:
\begin{equation}
    b(x)=a(x)+\mathcal R_\theta(a(x),\bar z(x)).
    \label{eq:population}
\end{equation}
Here $\bar z(x)=(z(x)-\mu_z)\oslash d_z$ uses source-fitted means
$\mu_z$ and standard deviations $d_z$, with elementwise division $\oslash$.
It modulates local/coarse temporal convolutions to produce the shared
prediction for calibration.

\subsection{Subject-Adaptive Factor Transport}

FT adapts channel and timing patterns by partitioning the 64 bases into
response groups indexed by $h$ on the EEG datasets
(THINGS-EEG2 and Alljoined-1.6M):
\begin{equation}
    \widetilde B_{s,h,j}
    =\rho_{s,h,j}\operatorname{Shift}_{\tau_{s,h}}
    \!\left[(M_sB_{h,j})\odot
    \left(g^c_{s,h}(g^t_{s,h})^\top\right)\right].
    \label{eq:continuous-transport}
\end{equation}
Parameters $\Theta_s^{\mathrm{FT}}$ comprise identity-initialized
channel mixing $M_s$, channel/temporal gains $g^c/g^t$, latency $\tau$,
and factor gain $\rho$; $\odot$ denotes elementwise multiplication.
Indices $h$, $j$, and $g$ denote response groups, within-group bases,
and visual descriptors, respectively.
MEG uses native factor updates (\cref{sec:supp-method-details}).

Reusing $z(x)$ with transported bases gives $\widetilde a_s(x)$.
With $\Delta_s(x)=\widetilde a_s(x)-a(x)$, FT adds the centered anchor change:
\begin{equation}
    b_s(x)=b(x)+\Delta_s(x)-\overline\Delta_s,
    \label{eq:transport-output}
\end{equation}
where $\overline\Delta_s=K^{-1}\sum_i\Delta_s(x_i)$.
Centering preserves $b(x)$'s support mean; FT changes response geometry
while preserving $z(x)$ and the shared refiner.

\subsection{Source-Grounded Residual Memory}

To correct errors remaining after FT, RM maps image keys to fixed
source-response coordinates.
Orthonormalizing flattened bases gives
$Q\in\mathbb R^{64\times CT}$. Matrices $Y_s,F_s\in\mathbb R^{K\times CT}$
stack vectorized $y_{s,i}$ and $b_s(x_i)$, respectively.
The operator $\mathcal C_s$ subtracts column means across supports.
With fixed source-residual RMS $\sigma_0$, errors and their coordinates are
\begin{align}
    E&=\sigma_0^{-1}\mathcal C_s[Y_s-F_s], \nonumber\\
    U&=EQ^\top(QQ^\top+\eta I)^{-1}.
    \label{eq:memory-target}
\end{align}
Here $\eta=.05$ stabilizes projection. To retrieve errors by image content, RM reuses
the refiner context as bounded keys $\kappa(x)=\tanh(\bar z(x)/4)$;
the fixed divisor controls saturation.
With $\bar\kappa_s=K^{-1}\sum_i\kappa(x_i)$,
rows $(K_s)_{i:}=\kappa(x_i)-\bar\kappa_s$ form the centered key matrix.
Ridge regression~\cite{hoerl1970ridge}
fits keys to error coordinates:
\begin{equation}
    W_s(\lambda)=(K_s^\top K_s+\lambda I)^{-1}K_s^\top U.
    \label{eq:memory-fit}
\end{equation}
To select regularization without reserving supports,
PRESS~\cite{allen1974press} gives omitted-fit predictions $q_{ij}(\lambda)$;
native coordinates $EQ^\top$ with omitted-fit intercept centering give
targets $t_{ij}$. Each direction selects its penalty:
\begin{equation}
    \lambda_{s,j}=\arg\min_{\lambda\in\Lambda\cup\{\infty\}}
       \sum_i\bigl(q_{ij}(\lambda)-t_{ij}\bigr)^2.
    \label{eq:memory-selection}
\end{equation}
Here $i$ and $j$ index support samples and error directions, respectively;
infinity suppresses a direction.
Selected columns form $W_s\in\mathbb R^{64\times64}$, conditional on frozen
keys, $Q$, and FT. The fixed grid $\Lambda$ and centering are specified in
\cref{sec:supp-method-details}.

For a query key $\kappa_c(x)=\kappa(x)-\bar\kappa_s$,
\begin{align}
    \delta_s(x)&=\sigma_0\kappa_c(x)W_sQ,
    \label{eq:memory-output}\\
    \hat y_s(x)&=p_s+b_s(x)+\operatorname{reshape}(\delta_s(x)).
    \label{eq:renderer}
\end{align}
Here $\sigma_0$ restores response units. RM transfers support image--error
associations to unseen images without updating the generator.

\subsection{Training and calibration strategy}

\paragraph{Source training.}
Source training learns a reusable generator, excluding target $s$ under
leave-one-subject-out (LOSO) training.
For the EEG datasets, we fit an image-independent prior, freeze it, then
learn visual projectors and bases; MEG jointly fits mean-subtracted source
responses. For training-scale targets $Y$ and predictions $\hat Y$
(prior-plus-anchor on EEG), we minimize
\begin{equation}
    \mathcal L_{\mathrm{gen}}
    =\operatorname{MSE}(\hat Y,Y)
     +\lambda_c\mathcal L_{\mathrm{corr}}(\hat Y,Y)
     +\lambda_t\mathcal R_{\mathrm{time}}.
    \label{eq:generation-loss}
\end{equation}
Here MSE denotes mean squared error;
$\mathcal L_{\mathrm{corr}}=1-\overline r$ averages sample-wise Pearson;
$\mathcal R_{\mathrm{time}}$ penalizes adjacent-sample changes and changes
in slope of normalized temporal factors, encouraging smooth response
profiles. We use $(\lambda_c,\lambda_t)=(.05,.001)$ on EEG and $(.03,0)$
on MEG. The Response Refiner then minimizes
\begin{equation}
    \mathcal L_{\mathrm{ref}}
    =\operatorname{MSE}\!\left(b/\sigma_0,
                        Y_{\mathrm{src}}^{\mathrm{res}}/\sigma_0\right),
    \label{eq:refiner-loss}
\end{equation}
where $Y_{\mathrm{src}}^{\mathrm{res}}$ averages mean-subtracted source
responses. Its RMS $\sigma_0$ normalizes refiner training and RM coordinates.

\paragraph{Target calibration.}
To accommodate a new subject's response statistics and geometry, target
calibration estimates $p_s$ and fits $\Theta_s^{\mathrm{FT}}$ from supports,
keeping source weights frozen:
\begin{equation}
    \mathcal L_{\mathrm{FT}}
    =\operatorname{MSE}\!\left(\mathcal C_sF_s/\sigma_s,
                              \mathcal C_sY_s/\sigma_s\right)
     +\lambda_{\mathrm{id}}\mathcal R_{\mathrm{id}}.
    \label{eq:transport-loss}
\end{equation}
Here $\sigma_s=[(KCT)^{-1}\sum_i\|y_{s,i}-p_s\|_F^2]^{1/2}$
is the residual RMS over $K$ supports, $C$ sensors, and $T$ samples.
Centering removes shared activity; $\sigma_s$ normalizes response magnitude.
$\mathcal R_{\mathrm{id}}$ discourages departures from identity transport,
weighted by $\lambda_{\mathrm{id}}=.001$ on EEG and $0$ on MEG.
Fixing FT then defines the errors for RM
(\cref{eq:memory-fit,eq:memory-selection}); removing FT requires refitting RM.

\paragraph{Native fMRI.}
For individual voxel layouts, fixed anatomical maps transport source heads
into a native bank $B_s^0\in\mathbb R^{64\times V_s}$.
With the frozen source latent $\phi(x)$, spatial FT fits a Ridge-regularized
basis update $D_s$ from centered supports:
\begin{equation}
    \hat y_s^{\mathrm{fMRI}}(x)
    =p_s+\bigl(\phi(x)-\bar\phi_s\bigr)(B_s^0+D_s)+\delta_s(x).
    \label{eq:native-fmri}
\end{equation}
Here $\bar\phi_s$ is the support latent mean. RM uses fixed coordinates from
$B_s^0$ and source-standardized latent keys, replacing $CT$ by $V_s$ above.
The Natural Scenes Dataset (NSD) has no temporal transport; its closed-form spatial update and matched
controls are in \cref{sec:supp-nsd-implementation,tab:supp-nsd-factorial}.

\section{Experiments}
\label{sec:experiments}

\begin{table*}[t]
    \centering
    \tableformat
    \setlength{\tabcolsep}{2.2pt}
    \caption{Within-subject visual-to-brain generation on four benchmarks.
    Raw and Resid.\ denote raw and training-response-residual Pearson
    correlation, respectively.
    Per-metric sources and simple controls are in
    \cref{sec:supp-protocol,tab:supp-local-within}.
    NSD evaluates all methods against responses averaged over three repeats;
    SynBrain's released predictions are rescored under this same convention.}
    \label{tab:within}
    \begin{tabular*}{\textwidth}{@{\extracolsep{\fill}}l*{12}{r}@{}}
        \toprule
        & \multicolumn{3}{c}{THINGS-EEG2~\cite{gifford2022things}}
        & \multicolumn{3}{c}{THINGS-MEG~\cite{hebart2023things}}
        & \multicolumn{3}{c}{Alljoined-1.6M~\cite{xu2025alljoined}}
        & \multicolumn{3}{c}{NSD~\cite{allen2022nsd}} \\
        \cmidrule(lr){2-4}\cmidrule(lr){5-7}\cmidrule(lr){8-10}\cmidrule(lr){11-13}
        Method
        & Raw $\uparrow$ & Resid.\ $\uparrow$ & MSE $\downarrow$
        & Raw $\uparrow$ & Resid.\ $\uparrow$ & MSE $\downarrow$
        & Raw $\uparrow$ & Resid.\ $\uparrow$ & MSE $\downarrow$
        & Raw $\uparrow$ & Resid.\ $\uparrow$ & MSE $\downarrow$ \\
        \midrule
        G{\"u}{\c{c}}l{\"u} et al.~\cite{guclu2015deep}
            & .2399 & .0276 & .1601
            & .2430 & .0457 & .8310
            & .0755 & .0065 & .3975
            & .7338 & .1826 & .0925 \\
        Yamins et al.~\cite{yamins2014performance}
            & .2355 & .0208 & .1748
            & .2690 & .0381 & .7960
            & .0742 & .0056 & .4284
            & .7301 & .1787 & .0943 \\
        Xu et al.~\cite{xu2026image}
            & .4250 & .0080 & .1090
            & .3790 & .0194 & .6630
            & .0302 & .0005 & .4326
            & .8091 & .2708 & .0611 \\
        SynBrain~\cite{mai2025synbrain}
            & .2760 & .0143 & .0644
            & .3020 & .0259 & .7270
            & .1627 & .0117 & .2987
            & .8228 & \textbf{.3564} & .0623 \\
        Concept2Brain~\cite{concept2brain2026}
            & .6184 & .0623 & .0234
            & .5143 & .0963 & .5904
            & .3610 & .0306 & .0866
            & .8168 & .1103 & .0571 \\
        \midrule
        \method (ours)
            & \textbf{.6798} & \textbf{.2140} & \textbf{.0202}
            & \textbf{.5621} & \textbf{.2047} & \textbf{.5444}
            & \textbf{.3838} & \textbf{.0836} & \textbf{.0856}
            & \textbf{.8400} & .3480 & \textbf{.0507} \\
        \bottomrule
    \end{tabular*}
\end{table*}

\subsection{Datasets and protocols}

We evaluate THINGS-EEG2 (ten subjects, 63 channels $\times$ 250 time
samples)~\cite{gifford2022things}, THINGS-MEG (four subjects)~\cite{hebart2023things},
Alljoined-1.6M (twenty subjects, 32 channels $\times$ 250 time samples), and NSD
(subjects 1/2/5/7, 12,682--15,724 native voxels)~\cite{allen2022nsd}.

EEG/MEG evaluations use all official training images
(16,540 per subject for EEG2 and Alljoined-1.6M, and 19,848 for MEG) and
all 200 official test images. NSD uses all 9,000 training and 1,000 test images.
Concept2Brain selects sensor budgets on training-only concept splits,
then refits on all training images. Test responses are not used for fitting
or checkpoint selection (\cref{sec:supp-protocol}).

LOSO excludes the target subject from source training. Calibration uses
$K=400$ distinct target-training images; conditions share frozen sources,
supports and complete official tests. For EEG, this support set corresponds
to approximately ten minutes of stimulus-presentation data, with four
repeats per image.

\subsection{Evaluation}

The primary metric is sample-wise Pearson after subtracting the target's
outer-training response mean from predictions and measurements.
This residual score isolates image variation; raw Pearson and MSE assess
full responses. MSE is compared only within each dataset.
EEG spatiotemporal predictivity correlates across all 200 images at each
electrode and time. Curves average subjects; uncertainty uses 95\%
subject-bootstrap intervals. The training mean is an evaluator statistic,
not a zero-shot input or a test-derived fit. Support draws vary calibration
images, independently of source seeds.
Shuffled controls refit the target calibration path with incorrect
image--response correspondence and identical budgets.

\paragraph{Comparisons.}
Baseline configurations, training budgets, and inference settings are detailed
in the supplementary material; residual correlations are computed on the
complete official test sets using a common evaluator.
Classical CNN comparisons use fixed fc7 features with fitted readouts
(\cref{sec:supp-cnns}). For the method of Yamins et al.,
VGG16 Ridge supplies EEG, Alljoined and NSD metrics, and MEG residual scores.
EEG's main generator is a three-model ensemble; analyses use matched single
models. Adaptation adds FT and Coordinate RM to frozen Anchor--Refiner sources
on sensors, and native spatial FT and RM on NSD. Information/module studies
share sources and supports (\cref{tab:supp-model-index}); removing FT requires refitting RM.

\section{Results}
\label{sec:results}

\subsection{Visual-to-brain generation}

\Cref{tab:within} reports raw Pearson correlation, residual Pearson correlation,
and MSE across four benchmarks.
On EEG2, MEG, and Alljoined, SENSA reaches residual
correlation of $.2140$, $.2047$, and $.0836$, respectively, versus $.0623$,
$.0963$, and $.0306$ for the strongest recent generator
on each dataset. It also attains the highest raw
correlation and lowest MSE among the listed recent generators, including raw $r=.8400$ and
MSE $.0507$ on NSD, where SynBrain's released predictions keep the higher
residual correlation ($.3564$ versus $.3480$).
Simple pooled-CLIP Ridge reaches residual $.1117$ on EEG and $.0462$ on
Alljoined (\cref{tab:supp-local-within}); matched single-model SENSA reaches
$.2004$ and $.0836$. On Alljoined, mean-response and CLIP Ridge references
have lower MSE ($.0835$ and $.0834$) than SENSA ($.0856$).
SENSA's sensor advantage persists after removing shared activity, supporting
image-dependent prediction. Same-feature controls
(\cref{tab:core-mechanisms}) examine response structure.

\begin{figure}[!tbp]
    \centering
    \includegraphics[width=\columnwidth]{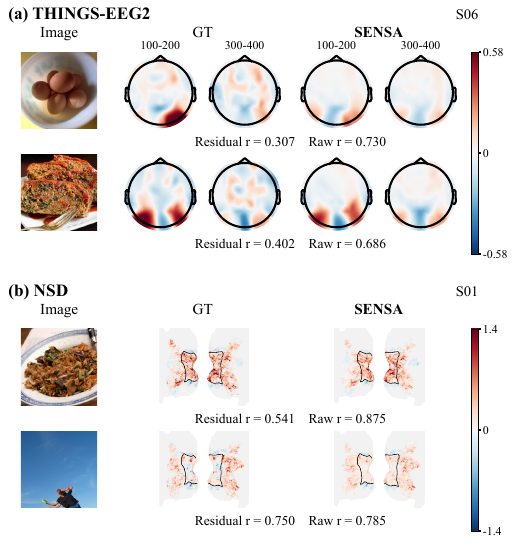}
    \caption{\textbf{Selected native-space reconstructions.}
    (a) THINGS-EEG2 scalp averages at 100--200 and 300--400\,ms.
    (b) NSD visual cortex with three-repeat-averaged targets.
    Two high-residual examples with distinct predictions per dataset;
    measured/predicted responses share scales.
    Scores use complete responses. Selection and model identities:
    \cref{sec:supp-reconstruction-cases}.}
    \label{fig:four-dataset-cases}
\end{figure}

\paragraph{Stimulus-dependent response structure.}
\Cref{fig:four-dataset-cases} shows posterior EEG polarity contrasts and
distinct NSD cortical distributions. Shared measured/predicted scales
expose recovered patterns and amplitude differences. These selected cases
complement complete-test quantification. MEG and Alljoined:
\cref{fig:supp-native-cases}.

\subsection{Stimulus-specific prediction}
\label{sec:image-analysis}

\begin{figure}[!tbp]
    \centering
    \includegraphics[width=\columnwidth]{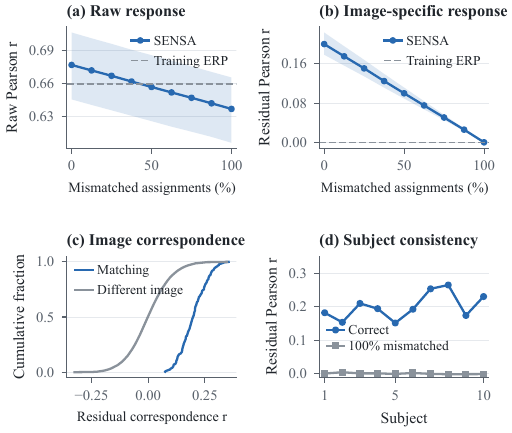}
    \caption{\textbf{Stimulus identity beyond shared responses.}
    One fixed model, 10 EEG2 subjects, all 200 official test images.
    (a-b) Mismatching curves average 128 seeded permutations per nonzero
    fraction; bands are 95\% subject-bootstrap intervals.
    Dashed: training ERP. (c) Cohort-mean matching/different-image
    residual distributions. (d) Correct versus mismatched correspondence
    by subject.
    }
    \label{fig:stimulus-identity}
\end{figure}

\paragraph{Image-identity perturbations.}
With fixed predictions and completely mismatched images
(\cref{fig:stimulus-identity}), raw Pearson remains $.6369$, versus $.6771$
for correct assignments and $.6599$ for the training ERP.
Residual correlation instead changes from $.2004$ to $.0004$.
With predictions unchanged, this contrast isolates image identity:
high raw correlation survives mismatching, while residual predictivity
depends on the correct stimulus. Correct assignments exceed mismatching
in all ten subjects.

\begin{figure}[!tbp]
    \centering
    \includegraphics[width=\columnwidth]{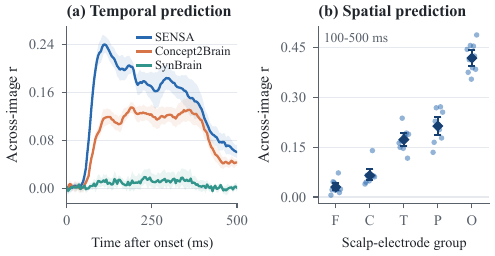}
    \caption{\textbf{When and where image-dependent responses are predictable.}
    All 200 EEG test images; Concept2Brain is trained on the complete
    official training set.
    (a) Across-image correlation averaged over electrodes and subjects;
    bands: 95\% subject-bootstrap intervals.
    (b) 100--500\,ms scalp groups: frontal, central, temporal, parietal,
    occipital. Dots: subjects; diamonds: means and bootstrap intervals.
    Full electrode map: \cref{fig:all-electrode-map}.}
    \label{fig:spatiotemporal}
\end{figure}

\paragraph{Spatiotemporal prediction patterns.}
\Cref{fig:spatiotemporal} resolves predictivity over electrodes and time.
Over 100--500\,ms, mean across-image correlation is $.1563$ for SENSA,
$.1050$ for Concept2Brain, and $.0083$ for SynBrain.
SENSA peaks at occipital electrodes with early and sustained predictivity.
Here each score compares images at a fixed electrode and time, testing
stimulus-dependent variation beyond similarity of the overall waveform.
These localized image-dependent patterns complement the identity control.
Image-wise comparisons are in
\cref{fig:residual-examples}.

\subsection{Calibration to unseen subjects}

\Cref{tab:adaptation} compares methods at equal support budgets.
SENSA reaches residual $.1091$, $.0825$, and $.0438$
on EEG, MEG, and Alljoined, respectively, exceeding the corresponding baseline scores.

\begin{table*}[t]
    \centering
    \tableformat
    \setlength{\tabcolsep}{3pt}
    \caption{\textbf{Unseen-subject adaptation.}
    Residual Pearson with zero or 400 paired target-training
    images. Reported results cover complete subject cohorts.
    SENSA uses frozen Anchor--Refiner, FT and Coordinate RM on sensors;
    external methods use LOSO extensions.
    NSD zero-shot anatomically transfers source predictions;
    SENSA's $K400$ calibrates native spatial bases and RM.
    Full metrics: \cref{tab:supp-external-calibration,tab:supp-final-information}.}
    \label{tab:adaptation}
    \begin{tabular*}{\textwidth}{@{\extracolsep{\fill}}l*{8}{c}@{}}
        \toprule
        & \multicolumn{2}{c}{THINGS-EEG2~\cite{gifford2022things}}
        & \multicolumn{2}{c}{THINGS-MEG~\cite{hebart2023things}}
        & \multicolumn{2}{c}{Alljoined-1.6M~\cite{xu2025alljoined}}
        & \multicolumn{2}{c}{NSD~\cite{allen2022nsd}} \\
        \cmidrule(lr){2-3}\cmidrule(lr){4-5}\cmidrule(lr){6-7}\cmidrule(lr){8-9}
        Method
        & Zero-shot & Paired $K400$
        & Zero-shot & Paired $K400$
        & Zero-shot & Paired $K400$
        & Zero-shot & Paired $K400$ \\
        \midrule
        SynBrain~\cite{mai2025synbrain}
            & .0109 & .0120
            & .0438 & .0249
            & -.0059 & .0020
            & .0459 & .2623 \\
        Concept2Brain~\cite{concept2brain2026}
            & .0184 & .0450
            & .0528 & .0631
            & .0037 & .0206
            & .0298 & .1587 \\
        \method (ours)
            & .\textbf{0374} & \textbf{.1091} & .\textbf{0588} & \textbf{.0825}
            & .\textbf{0126} & \textbf{.0438} & \textbf{.0460} & \textbf{.2995} \\
        \bottomrule
    \end{tabular*}

\end{table*}

\begin{table}[!b]
    \centering
    \tableformat
    \setlength{\tabcolsep}{3pt}
    \caption{\textbf{Target information for calibration.}
    Residual Pearson on complete subject cohorts.
    Support conditions use the same $K400$ samples;
    shuffled pairs refit FT and RM with incorrect correspondence.
    Full metrics: \cref{tab:supp-final-information}.}
    \label{tab:target-information}
    \begin{tabular*}{\columnwidth}{@{\extracolsep{\fill}}lrrrr@{}}
        \toprule
        Support & EEG2 & MEG & Alljoined & NSD \\
        \midrule
        No support & .0374 & .0588 & .0126 & .0460 \\
        Response-only & .0865 & .0592 & .0361 & .1272 \\
        Shuffled & .0261 & .0247 & .0184 & .0005 \\
        Correct & \textbf{.1091} & \textbf{.0825} & \textbf{.0438} & \textbf{.2995} \\
        \bottomrule
    \end{tabular*}
\end{table}

\begin{table}[!t]
    \centering
    \tableformat
    \setlength{\tabcolsep}{3pt}
    \caption{\textbf{Structured response generation.}
    Same-feature readouts: EEG2 (10 subjects), MEG (4), Alljoined (20).
    The two banks match training budgets but use different regularization
    penalties; the full bank has more parameters.
    Ridge matches features, data and priors, with training-only
    GCV~\cite{golub1979gcv};
    it is not compute- or parameter-matched.
    Complete metrics: \cref{sec:supp-review-controls}.}
    \label{tab:core-mechanisms}
    \begin{tabular*}{\columnwidth}{@{\extracolsep{\fill}}llrrr@{}}
        \toprule
        Dataset & Metric & \shortstack{Full\\waveform}
        & \shortstack{Feature\\Ridge} & Separable \\
        \midrule
        \multirow{2}{*}{EEG2}
            & Resid.\ $\uparrow$ & .1631 & .1847 & \textbf{.2004} \\
            & MSE $\downarrow$ & .0215 & .0205 & \textbf{.0203} \\
        \midrule
        \multirow{2}{*}{MEG}
            & Resid.\ $\uparrow$ & .1726 & .2011 & \textbf{.2047} \\
            & MSE $\downarrow$ & .5981 & .5447 & \textbf{.5444} \\
        \midrule
        \multirow{2}{*}{Alljoined}
            & Resid.\ $\uparrow$ & .0518 & .0723 & \textbf{.0836} \\
            & MSE $\downarrow$ & .0873 & .0857 & \textbf{.0856} \\
        \bottomrule
    \end{tabular*}
\end{table}

\begin{table}[!b]
    \centering
    \tableformat
    \caption{\textbf{Matched subject-calibration modules.}
    Residual Pearson at $K400$.
    Conditions share source weights, supports, and support means.
    NSD uses anatomically mapped source bases and spatial-only FT;
    RM is refitted for each FT setting.
    Full metrics: \cref{tab:supp-final-factorial,tab:supp-nsd-factorial}.}
    \label{tab:calibration-modules}
    \begin{tabular*}{\columnwidth}{@{\extracolsep{\fill}}ccrrrr@{}}
        \toprule
        FT & RM & EEG2 & MEG & Alljoined & NSD \\
        \midrule
        $\times$ & $\times$ & .0865 & .0592 & .0361 & .1272 \\
        $\checkmark$ & $\times$ & .0979 & .0806 & .0410 & \textbf{.2995} \\
        $\times$ & $\checkmark$ & .1032 & .0821 & .0420 & .1371 \\
        $\checkmark$ & $\checkmark$ & \textbf{.1091} & \textbf{.0825} & \textbf{.0438} & \textbf{.2995} \\
        \bottomrule
    \end{tabular*}
\end{table}

\paragraph{Calibration under different support conditions.}
With matched $K400$ supports, correct pairing improves EEG residual
correlation from $.0865$ with response-only calibration to $.1091$.
Correct pairs outperform shuffled supports in 10/10 EEG, 4/4 MEG,
19/20 Alljoined, and all four NSD subjects
(\cref{tab:target-information}), showing that calibration uses
image--response correspondence beyond target statistics.

\subsection{Generation and calibration mechanisms}
\label{sec:module-ablations}

\paragraph{Separating shared activity preserves image variation.}
Without the independent prior, residual correlation changes from
$.2004$, $.2047$, and $.0836$ to $.1520$, $.1522$, and $-.0040$
on EEG, MEG, and Alljoined, respectively (\cref{tab:supp-within-modules}).
Alljoined retains the visual-training budget but omits prior pretraining.

\paragraph{Separable bases improve prediction.}
Separable bases improve over full waveforms in 10/10 EEG, 4/4 MEG,
and 19/20 Alljoined subjects (\cref{tab:core-mechanisms}).
With fixed visual descriptors, these comparisons support separable response
structure under the tested training budgets. Ridge is lower on EEG/Alljoined
and comparable on MEG, separating structural benefits from feature availability.
Retrained visual-branch controls and frozen temporal sensitivity appear in
\cref{tab:supp-within-modules,fig:supp-complete-diagnostics}.
Three-seed controls retain structural gains in every EEG/MEG subject under
common waveform regularization (\cref{tab:supp-common-regularization}),
separating the advantage from penalty choice.

\begin{table}[!t]
    \centering
    \tableformat
    \setlength{\tabcolsep}{4pt}
    \caption{\textbf{Calibration across generation strategies.}
    Residual Pearson; EEG LOSO (10 subjects, seed 42, $K400$, all 200 test images).
    Direct: \cref{tab:adaptation}'s SENSA; other heads: 5,000 updates.
    Shared support means; RM refitted per FT setting.
    Full metrics: \cref{tab:supp-generator-ft-rm}.}
    \label{tab:generator-calibration}
    \begin{tabular*}{\columnwidth}{@{\extracolsep{\fill}}lrrrr@{}}
        \toprule
        Generator & Neither & FT only & RM only & FT+RM \\
        \midrule
        Direct & .0865 & .0979 & .1032 & \textbf{.1091} \\
        cVAE & .0812 & .0943 & .0985 & \textbf{.1051} \\
        Residual diffusion & .0173 & .0226 & .0219 & \textbf{.0245} \\
        Flow matching & .0185 & .0250 & .0239 & \textbf{.0270} \\
        cGAN & .0489 & .0668 & .0619 & \textbf{.0713} \\
        \bottomrule
    \end{tabular*}
\end{table}

\paragraph{Structural alignment and image-conditioned correction.}
Complete EEG calibration exceeds either module in all ten subjects at $K400$
(\cref{tab:calibration-modules}), reaching $.1093\pm.0003$ across three
source seeds (mean$\pm$SD). RM improves FT across three support draws
(\cref{sec:supp-source-seeds,fig:support-gains}).
FT aligns shared response structure; RM corrects its remaining image-dependent
error, with dataset-dependent contributions.
Alljoined shows individual variation and a raw/MSE trade-off; MEG's
$.0825$ is close to RM-only's $.0821$.
On NSD, FT raises $.1272$ to $.2995$ in all four subjects;
RM alone reaches $.1371$, with zero correction selected after FT.

\paragraph{Calibration transfers across generation strategies.}
FT+RM improves SENSA and four alternative generators' no-module references
in all ten subjects (\cref{tab:generator-calibration}), supporting transfer
across generation objectives.
FT+RM retains gains after 16-sample diffusion averaging
($.0487$ to $.0693$; \cref{tab:supp-diffusion-averaging}).

\begin{figure}[!tbp]
    \centering
    \includegraphics[width=\columnwidth]{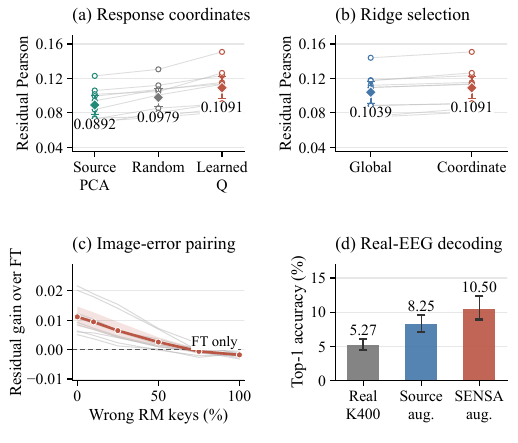}
    \caption{\textbf{Source-grounded memory and downstream utility.}
    EEG: 10 subjects, seed 42, $K400$, all 200 test images.
    (a) Coordinates. (b) Global versus coordinate-wise Ridge.
    (c) RM-key mismatch, FT fixed; three permutations/subject.
    (d) Real-EEG decoding with/without augmentation;
    three decoder seeds/subject.
    Diamonds/bars/bold: means; circles/lines: subjects;
    whiskers/band: 95\% subject-bootstrap intervals.
    Protocols: \cref{sec:supp-response-coordinate-controls,sec:supp-synthetic-augmentation}.}
    \label{fig:calibration-information}
\end{figure}

\paragraph{Source coordinates and coordinate-wise selection matter.}
With FT fixed, learned coordinates ($.1091$) exceed source-response PCA
($.0892$) and random coordinates ($.0979$) in all ten EEG subjects,
using 64-dimensional spans and Coordinate PRESS (\cref{fig:calibration-information}a).
Within the learned span, coordinate-wise Ridge selection improves global
selection ($.1039$ to $.1091$) with fixed keys and supports
(\cref{fig:calibration-information}b).
Same-span rotations average $.1036$ versus $.1091$ for learned axes
(\cref{tab:supp-same-span-rotations}).
Span and rotation controls support learned response coordinates;
the within-span comparison supports per-direction regularization.

\paragraph{Memory requires image--error correspondence.}
With FT fixed, 50\% and complete RM-key mismatch reduce its increment from
$.0112$ to $.0026$ and $-.0018$, respectively
(\cref{fig:calibration-information}c).
The correction therefore depends on image--error correspondence. Further controls:
\cref{fig:supp-source-transfer,tab:supp-frozen-readouts}.

\paragraph{Decoding real EEG with generated responses.}
Ranking all 200 templates against real EEG by support-centered Pearson
yields Top-1 of $16.60\%$ with source plus support mean,
$19.70\%$ with FT, and $25.45\%$ with RM, versus $2.95\%$ for target-only
encoding Ridge. At $K400$, RM improves FT in 9/10 subjects by $5.75$ points
(95\% subject-bootstrap interval: $3.10$--$8.15$).
This template test connects encoding quality to image identification
directly from measured responses, without training a separate decoder
(\cref{sec:supp-real-query-retrieval}).

\paragraph{Augmenting scarce decoder training data.}
Adding 16,140 synthetic pairs to $K400$ measured pairs raises real-EEG
Top-1 from $5.27\%$ to $10.50\%$, versus $8.25\%$ with source-generated pairs
(\cref{fig:calibration-information}d).
Tests use 80-repeat averages and three decoder initializations.
Relative to source-generated augmentation, personalization adds $2.25$
points (95\% subject-bootstrap interval $[1.30,3.17]$), showing the value
of adapting synthetic responses beyond increasing their number.
Using 2,964 additional images reaches $10.75\%$.
Both SENSA augmentation settings improve all ten subjects over real-only
training in the full 200-way test
(\cref{sec:supp-synthetic-augmentation}).

Increasing paired support improves individualized prediction across all four
datasets without source retraining
(\cref{sec:supp-support-budget}).

\begin{samepage}
\section{Conclusion}
\label{sec:conclusion}

SENSA separates shared activity from image-dependent generation through
structured response bases. Subject-Adaptive Factor Transport and
Source-Grounded Residual Memory calibrate their expression and remaining
errors for a new subject.
Residual evaluation, identity perturbations, and matched support comparisons
connect stimulus-specific prediction to individualized transfer.
Calibration preserves shared generator weights while personalizing predictions
from limited paired recordings.
The generated responses also improve real-EEG decoding.

\end{samepage}

\section*{LLM Usage Statement}

OpenAI ChatGPT and Codex assisted with code development and with drafting
and revising the initial manuscript. The authors take full responsibility
for the final manuscript and its scientific claims.

{
    \small
    \bibliographystyle{ieeenat_fullname}
    \bibliography{main}
}

\clearpage
\setcounter{section}{0}
\renewcommand{\thesection}{\Alph{section}}
\renewcommand{\theHsection}{supp.\arabic{section}}
\setcounter{figure}{0}
\renewcommand{\thefigure}{S\arabic{figure}}
\renewcommand{\theHfigure}{supp.\arabic{figure}}
\setcounter{table}{0}
\renewcommand{\thetable}{S\arabic{table}}
\renewcommand{\theHtable}{supp.\arabic{table}}

\maketitlesupplementary

\section{SENSA implementation and calibration}
\label{sec:supp-method-details}

\subsection{Model identity and visual representation}

SENSA's structured continuous generator is the generation backbone in
\cref{tab:within}. The complete sensor-time architecture in
\cref{fig:architecture} augments it with a Response Refiner,
Subject-Adaptive Factor Transport, and Source-Grounded Residual Memory
using coordinate-wise PRESS-selected Ridge.
\Cref{tab:adaptation} reports this sensor pipeline and its matched
target-information conditions. Module comparisons use the same frozen
Anchor--Refiner sources.

\begin{table*}[t]
    \centering
    \tableformat
    \caption{Configuration index. Module effects are compared within one
    source; results across rows do not define module increments.}
    \label{tab:supp-model-index}
    \begin{tabular*}{\textwidth}{@{\extracolsep{\fill}}llccc@{}}
    \toprule
    Evaluation & Shared source & Refiner & Calibration & RM \\
    \midrule
    Within-subject sensors (\cref{tab:within})
      & Structured backbone & No & None & No \\
    LOSO sensors (\cref{tab:adaptation})
      & Fixed LOSO Anchor--Refiner & Yes & FT & Coordinate PRESS \\
    Sensor module study (\cref{tab:calibration-modules})
      & Same complete LOSO sources & Yes & FT on/off & Refitted on/off \\
    LOSO fMRI (\cref{tab:adaptation,tab:calibration-modules})
      & Mapped source heads & No & Spatial FT & Coordinate PRESS \\
    NSD readout control
      & Source-supervised visual latent & No & Native voxel Ridge & No \\
    \bottomrule
    \end{tabular*}
\end{table*}

The four sensor descriptor groups are spatial, region, blur, and low-level
features. Spatial features concatenate $4\times4$ patch grids from frozen
CLIP ViT-L/14 hidden layers 8, 12, and 16, with training-fitted 32-dimensional
patch projections. This yields $3\times16\times32=1{,}536$ dimensions.
Region features pool the intermediate patch representation using foreground
and background masks, each split into center and border regions. Three layers
and four regions yield 384 dimensions, supplemented by four mask-geometry
statistics. Blur features use Gaussian scales $\sigma\in\{0,2,4,8\}$ and the
same three layers, yielding 384 dimensions. Low-level color, contrast,
and Gabor grids contribute 160, 64, and 192 dimensions.
These are complementary descriptor constructions, not four separately trained
vision encoders. Low-level descriptors bypass CLIP.
Region pooling uses fixed supplied soft foreground masks and their
complements, partitioned by center/border patch weights. A fixed
training-image audit checked the supplied masks against BiRefNet outputs;
mask construction does not use brain responses.

The 64-dimensional linear group projectors and learned scalar group gains
produce the fused code in \cref{eq:visual-code}. Feature standardization and
PCA statistics use training images. Source-model code statistics standardize
the refiner context and memory keys; target test images do not fit them.
In within-subject generation, the response factors and independent prior
are fitted using training responses. For LOSO, the held subject is excluded
from the source fit. The source factor amplitudes absorb the learned source
coefficient gate, preserving the anchor decoder.

\subsection{Generation and transport objectives}

For sensor responses, the structured backbone uses
\begin{equation}
    \mathcal L_{\mathrm{gen}}
    =\mathcal L_{\mathrm{MSE}}+
      \lambda_c\mathcal L_{\mathrm{corr}}+
      \lambda_t\mathcal R_{\mathrm{time}},
\end{equation}
where $\mathcal L_{\mathrm{corr}}$ is one minus mean sample-wise Pearson
correlation. Temporal regularization penalizes first and second differences
of normalized temporal factors. The prior and visual branches have separate
training stages; native budgets and objective controls retain their
dataset-specific settings.

The direct Response Refiner concatenates the anchor with its current waveform
input, maps the $2C$ channels to 64 hidden channels, and uses three temporal
blocks. Each block combines depthwise local convolution (kernel 7) with
a coarser average-pooled path (pooling factor 4, kernel 5), then upsamples
and mixes the features. Source-standardized visual context supplies feature
modulation. A final channel projection, initialized to zero, outputs the
correction. The direct branch evaluates this field once at the anchor.
Its source objective is MSE against the normalized source-average residual
response. The response scale is restored after prediction.
No diffusion sampling or test-time refiner-weight update is used.

In the EEG transport implementation, 64 factors are partitioned into four
groups of 16. This grouping indexes response factors, not the four visual
inputs. At 250\,Hz, each 250-sample basis is divided into 25 slots of ten
samples (40\,ms). Channel and factor parameters are shared across slots;
temporal gains act on the ten within-slot positions.
For channel rank $d=8$,
\begin{equation}
    M_s=I+\frac{0.1}{\sqrt d}L_sR_s^\top,
    \qquad g(v)=\exp(\log(2)\tanh v).
\end{equation}
$L_s$ initializes to zero, and the initial columns of $R_s$ are normalized.
All log-gain parameters initialize to zero. Shifts are
$2\tanh(\ell_{s,g})$ samples, bounded by 8\,ms in either direction.
Bilinear interpolation with border padding acts independently within each
slot. This is not a fitted global ERP onset shift.

Let $\mathcal C_s$ subtract the support mean and let
$Y_s$ stack support responses. Write
$F_s=\operatorname{stack}_i b_s(x_i)$ and
$\sigma_s^2=(KCT)^{-1}\|\mathcal C_sY_s\|_F^2$.
The final sensor FT minimizes
\begin{equation}
\begin{split}
    \mathcal L_{\mathrm{FT}}
    ={}&\operatorname{MSE}\bigl(\mathcal C_sF_s/\sigma_s,
                                \mathcal C_sY_s/\sigma_s\bigr)\\
      &+\lambda_{\mathrm{id}}\mathcal R_{\mathrm{id}}.
\end{split}
\end{equation}
Here $\lambda_{\mathrm{id}}=.001$ for EEG/Alljoined and $0$ for MEG;
$\sigma_s$ is clamped below at $10^{-5}$ in computation.
The EEG/Alljoined regularizer averages squared mixing-left, channel-gain,
temporal-gain, shift, and factor-gain parameters. After fitting, store
$\overline\Delta_s$ from the support. The query output retains $b(x)$ and adds
$\Delta_s(x)-\overline\Delta_s$; it does not pass the refined waveform through
the basis transport a second time. The matched EEG study uses 300 adapter
updates. MEG uses 200 Adam updates, with learning rate $.02$ for scalp and
time factors and $.0002$ for amplitude; its parameter-family controls retain
these rates. Alljoined retains its documented native implementation.

\subsection{Response coordinates and memory}

For sensor RM, flatten the source response bases as rows
of $A\in\mathbb R^{64\times CT}$, normalize each row, and orthonormalize:
\begin{equation}
    Q=(AA^\top)^{-1/2}A.
\end{equation}
The source bank has full row rank; no target response is used to select or
learn this span. Thus $QQ^\top=I$ up to numerical precision. The rows of
$Q$ are fixed response directions; image keys identify which correction
to retrieve for a query.

Support responses are averaged across available repeats. Let $\sigma_0$
denote the fixed source response scale. With all waveforms vectorized,
\begin{equation}
    E=\frac{1}{\sigma_0}\mathcal C_s
      \bigl[Y_s-\mathbf 1p_s^\top-B_s\bigr],
\end{equation}
where $B_s$ stacks the vectorized FT outputs. Subtracting $p_s$ is
equivalent to omitting it before the outer centering.
Set $V=EQ^\top$ and
$U=V(QQ^\top+0.05I)^{-1}$.
The latter is a stabilized coordinate projection, not an additional
gradient-trained encoder.

The frozen key function is $\kappa(x)=\tanh(\bar z(x)/4)$.
Subtract its support mean to form $K_s$, and fit
\begin{equation}
    W_s(\lambda)=(K_s^\top K_s+\lambda I)^{-1}K_s^\top U.
\end{equation}
This is a $64\times64$ target state. The key map, response bank, generator,
and fitted FT remain fixed. The solve uses all 400 supports without a
held-out fitting partition.

\subsection{Conditional PRESS selection}

For each candidate $\lambda$, let $F=K_sW_s(\lambda)$ and include the
unregularized intercept in the leverage:
\begin{equation}
    H=\frac{\mathbf 1\mathbf 1^\top}{K}
      +K_s(K_s^\top K_s+\lambda I)^{-1}K_s^\top,\qquad h_i=H_{ii}.
\end{equation}
For $K>1$, the implementation computes the omitted-fit, intercept-centered
prediction and target as
\begin{equation}
    q_i=\frac{F_i-h_iU_i}{1-h_i}+\frac{U_i}{K-1},
    \qquad t_i=\frac{K}{K-1}V_i.
\end{equation}
The extra centering terms express each observation relative to the intercept
of its omitted fit. Coordinate selection uses \cref{eq:memory-selection},
and the correction is decoded by \cref{eq:memory-output}.
Multiplication by $\sigma_0$ restores the decoded error to measurement units.
Its mean over support keys is zero, leaving the shared response mean
outside the memory correction.

PRESS here is conditional on frozen keys, response directions, and the
already fitted upstream response. It removes each row's self-influence on
this last linear mapping, not its influence on FT or on earlier fitting.
Support key means, selected $W_s$, and $Q$ are stored for image-only query
inference. At test time no response, gradient update, or new memory fit
is required.

\paragraph{Order and controls.}
Fit support response statistics, then FT, then RM against the remaining
error. Removing FT requires recomputing the errors and refitting RM.
Shuffled controls retain the same support images and response groups but
refit the appropriate calibration stages with incorrect correspondence.
The global PRESS and bounded-gain controls keep the same keys,
response coordinates, and upstream predictions; their outcomes are in
\cref{tab:supp-factorial-complete}.

\paragraph{Adopted Coordinate PRESS RM.}
The fit illustrated in \cref{fig:architecture}c uses centered
keys, targets $U$, fixed $Q$, and fitted upstream response. For each
$\lambda\in\{\exp(-5+0.5l):l=0,\ldots,20\}$, solve
$W(\lambda)=(K_s^\top K_s+\lambda I)^{-1}K_s^\top U$ and compute the
corresponding omitted-fit predictions $q(\lambda)$ with the same intercept
centering as above. Each output coordinate selects the candidate minimizing
$\sum_i(q_{ij}(\lambda)-t_{ij})^2$. The additional infinite-penalty candidate
sets its column to zero. Assemble the selected columns into $W_s$ and decode
$\delta_s(x)=\sigma_0\kappa_c(x)W_sQ$, without post-fit gains.
This fit uses all support images and no test responses or held-out
validation partition. Bounded-gain and coordinate-selected RM results retain
separate labels throughout the reported comparisons.

\paragraph{Bounded-gain reference.}
The bounded-gain control fixes $\lambda=1$, using the same omitted-fit
centering to fit
\begin{equation}
    \gamma_{s,j}=\operatorname{clip}_{[0,1]}
      \frac{\sum_iq_{ij}t_{ij}}{\max(\sum_iq_{ij}^2,10^{-12})}.
    \label{eq:memory-gain}
\end{equation}
It decodes $\sigma_0\kappa_cW_s(1)\operatorname{diag}(\gamma_s)Q$.
The bounded gate averages the available repeat views before computing
directional cross-products; it does not fit repeat-specific gains.
Its final-source factorial and same-upstream algorithm controls appear in
\cref{tab:supp-final-factorial,tab:supp-factorial-complete,tab:supp-coordinate-controls},
with bounded-gain and Coordinate PRESS variants labeled separately.

\subsection{Native fMRI implementation}
\label{sec:supp-nsd-implementation}

\paragraph{Visual projection and native spaces.}
NSD uses a native-voxel implementation rather than sensor-time bases.
Mean pooling 256 frozen OpenCLIP ViT-bigG/14 patch tokens gives
$f^{\mathrm{NSD}}(x)\in\mathbb R^{1664}$.
Training-fitted feature means and standard deviations give
$\tilde f=(f^{\mathrm{NSD}}-\mu_f)\oslash d_f$.
The learned projection is
\begin{equation}
    \phi_\theta(x)=L_2\!\left(
       \operatorname{LN}(\operatorname{GELU}(L_1\tilde f(x)))\right)
       \in\mathbb R^{64},
\end{equation}
where $L_1:1664\to256$ and $L_2:256\to64$ are affine layers.
Each training subject $r$ has a separate bias-free head
$A_r\in\mathbb R^{64\times V_r}$; source training keeps these voxel
spaces separate. Subjects 1/2/5/7 retain
15,724/14,278/13,039/12,682 voxels, respectively.
Training and test targets average the three available repetitions.

\paragraph{Source supervision and target calibration.}
For held subject $s$, source subjects are
$\mathcal J_s=\{1,2,5,7\}\setminus\{s\}$.
Feature statistics use their pooled training features. For each source,
let $p_r^{\mathrm{tr}}$ and $\sigma_r^{\mathrm{tr}}$ be its training mean
and residual RMS, and
$\tilde y_r=(y_r-p_r^{\mathrm{tr}})/\sigma_r^{\mathrm{tr}}$.
Feature standard deviations and response RMS are clamped below at $10^{-5}$.
The projection and source heads jointly minimize
\begin{equation}
\begin{split}
    \mathcal L_{\mathrm{NSD}}
    =\frac{1}{|\mathcal J_s|}\sum_{r\in\mathcal J_s}
    \bigl[&\operatorname{MSE}(\phi_\theta A_r,\tilde y_r)\\
          &+.03(1-\bar r(\phi_\theta A_r,\tilde y_r))\bigr],
\end{split}
\end{equation}
where $\bar r$ averages sample-wise Pearson over the minibatch.
Each source contributes all 9,000 training images. AdamW uses learning
rate $3\times10^{-4}$, weight decay $10^{-4}$, batch size 64, gradient
clipping at 5, seed $42+s$, and a fixed final epoch 15.
The held subject contributes no source-training responses.

Freeze $\phi_\theta$ and its feature statistics before fitting the target.
For $K400$ supports, let $\Phi_s$ stack their 64-dimensional projections,
$Z_s=\mathcal C_s\Phi_s$, and $Y_s$ stack native voxel responses.
The unanchored native-readout control fits
\begin{equation}
    H_s=(Z_s^\top Z_s+I_{64})^{-1}Z_s^\top\mathcal C_sY_s.
\end{equation}
This float64 Ridge solve uses fixed $\alpha=1$, support means
$p_s$ and $\bar\phi_s$, and no target validation split.
Its mean-only control predicts $p_s$; its shuffled control permutes supports
before refitting. This readout is retained in the native parameter and
budget controls; the matched spatial FT/RM implementation follows below.

\paragraph{Anatomical zero-shot transfer.}
To predict in an unseen native voxel space without paired target responses,
all three methods in \cref{tab:adaptation} use the same fixed anatomical
transfer. Let $q_r(j)$ be source voxel $j$'s MNI coordinate and $q_s(v)$
the target coordinate, obtained from the official
\texttt{MNI-to-func1pt8} pullback maps with one-based indices converted to
zero-based coordinates. Define
\begin{equation}
\begin{split}
    j_r(v)&=\arg\min_{j\in\mathcal V_r}\|q_r(j)-q_s(v)\|_2,\\
    \hat y_s^{\,0}(x)_v
      &=\frac{1}{|\mathcal J_s|}
        \sum_{r\in\mathcal J_s}\hat y_r(x)_{j_r(v)}.
\end{split}
\end{equation}
Here $\mathcal V_r$ is the measured source mask. SENSA's source prediction
is $p_r^{\mathrm{tr}}+\sigma_r^{\mathrm{tr}}\phi_\theta(x)A_r$;
Concept2Brain decodes under each source's one-hot condition and restores
its saved source scale; SynBrain uses its source mapper at $t=0$ and
decodes at each source's native voxel count. Every target voxel is
retained, with nearest-neighbor extrapolation outside source coverage.
There is no target-dependent fitting, smoothing, or source weighting.
Target masks and registration are available; no target responses, means,
or scales enter prediction. Dataset preprocessing is retained, so zero-shot
denotes zero paired calibration. All twelve method--fold predictions are
frozen before target scoring; the target-training mean is used only by
the residual evaluator.

\paragraph{Source-anchored spatial FT and RM.}
To personalize the source response structure in the native voxel space,
let $P_{r\to s}\in\{0,1\}^{V_r\times V_s}$ select the nearest source
voxel $j_r(v)$ defined above. Restoring source response scales gives
\begin{equation}
\begin{split}
    B_s^0&=\frac{1}{|\mathcal J_s|}
           \sum_{r\in\mathcal J_s}\sigma_r^{\mathrm{tr}}A_rP_{r\to s},\\
    D_s&=(Z_s^\top Z_s+I_{64})^{-1}
          Z_s^\top(\mathcal C_sY_s-Z_sB_s^0).
\end{split}
\end{equation}
The native spatial FT minimizes
$\|\mathcal C_sY_s-Z_s(B_s^0+D_s)\|_F^2+\|D_s\|_F^2$.
The unit penalty is fixed from the native-readout protocol.
Unlike sensor-time FT, this update has no temporal or latency parameters.
With FT off, $D_s=0$; all supported conditions retain $p_s$ and the same
support-centered image prediction.
Response-only retains the unordered support image set for centering,
without using image--response correspondence.

Row normalization and polar orthonormalization of $B_s^0$ give
$Q_s\in\mathbb R^{64\times V_s}$, fixed across FT settings.
RM reuses the coordinate projection and conditional PRESS fitting above,
with $\sigma_0=|\mathcal J_s|^{-1}\sum_r\sigma_r^{\mathrm{tr}}$.
Keys are $\tanh((\phi_\theta(x)-\mu_\phi)\oslash d_\phi/4)$;
$\mu_\phi,d_\phi$ use the pooled 27,000 source-training image projections.
RM is refitted against each FT condition's remaining support error.
Shuffled controls permute the same responses before both fits.
The four held subjects, all native voxels, and all 1,000 official test
images are evaluated in \cref{tab:supp-nsd-factorial}.
Source checkpoints, maps, support identities, and the PRESS grid are fixed;
all four subjects' predictions are saved before target-test scoring.

\paragraph{Within-subject training and final readout.}
The NSD entry in \cref{tab:within} instead trains an independent projection
and head on all 9,000 images of the evaluated subject, without other
subjects' responses. It retains the architecture, optimizer, batch size,
clipping, seed, and 15-epoch budget above, using correlation weight $.3$
and learning rates $3\times10^{-4}$, $3\times10^{-5}$, and $3\times10^{-6}$
for epochs 1--3, 4--8, and 9--15.
Its final direction-balanced readout differs from the fixed-$\alpha$
LOSO calibration. For training residual $v_i=y_i-p^{\mathrm{tr}}$,
fit Ridge to $d_i=v_i/\max(\sqrt{\|v_i\|_2^2/V_s},10^{-8})$.
Ridge uses an unpenalized intercept $\bar d$; conditional training PRESS
averaged over voxels selects a single global penalty
$\alpha\in\{10^{-4},10^{-3},\ldots,10^6\}$.
With centered latent matrix $Z$
and centered direction targets $D_c$, the resulting prediction is
\begin{equation}
\begin{split}
    W_\alpha&=(Z^\top Z+\alpha I_{64})^{-1}Z^\top D_c,\\
    \hat y(x)&=p^{\mathrm{tr}}+
       \gamma\bigl[\bar d+(\phi_\theta(x)-\bar\phi)W_\alpha\bigr].
\end{split}
\end{equation}
The nonnegative scalar
$\gamma=\max(0,\sum_i\langle\hat d_{-i},v_i\rangle/
\sum_i\|\hat d_{-i}\|_2^2)$ restores response amplitude,
where $\hat d_{-i}$ is the analytical omitted-fit direction prediction.
PRESS and this gain are conditional on the frozen full-training encoder,
prior, and direction normalization. All readout choices use training
responses; saved predictions are evaluated on all 1,000 official test
images. The matched ablations in \cref{tab:supp-nsd-within-modules}
retain this final recipe.

\subsection{Selection of reconstruction examples}
\label{sec:supp-reconstruction-cases}

\begin{figure}[!t]
    \centering
    \includegraphics[width=\columnwidth]{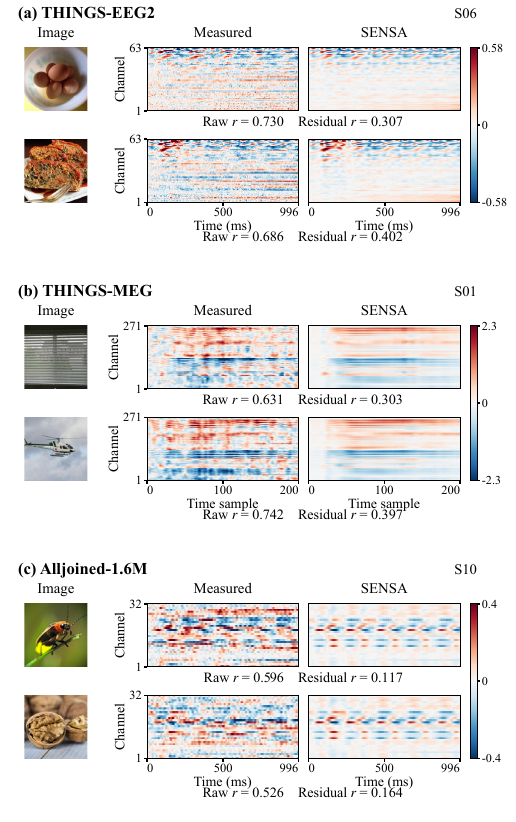}
    \caption{\textbf{Channel--time reconstructions across sensor datasets.}
    (a) THINGS-EEG2, 63 channels; (b) THINGS-MEG, 271 channels;
    (c) Alljoined-1.6M, 32 channels.
    Each dataset shows two existing examples with the stimulus, measured
    response, and SENSA prediction. All native channels and time samples
    are retained. EEG and Alljoined show time in milliseconds; MEG retains
    sample indices. Measured and predicted raw responses share one fixed
    color scale per dataset, without model-specific normalization.
    Raw and residual correlations use the complete response arrays.
    EEG uses a single seed-42 model. Selection and scaling are specified above.}
    \label{fig:supp-native-cases}
    \label{fig:supp-reconstruction-cases}
\end{figure}

\Cref{fig:four-dataset-cases,fig:supp-native-cases} illustrate selected successful reconstructions;
aggregate results use the complete official tests.
The main figure retains EEG subject 06 and NSD subject 01, using the
single seed-42 EEG model and the final 15-epoch direction-PRESS NSD
readout in \cref{tab:within}. EEG uses a single model rather than the
main-table ensemble.
Within each subject, candidates belong to the top 20\% of residual
correlations and exceed residual thresholds of $.2952$ for EEG and
$.4845$ for NSD. Raw correlation must be at least $.65$ and $.75$,
respectively. EEG additionally requires correlation of at least $.65$
between the concatenated measured and predicted scalp-window values.
Among eligible pairs, we minimize their predicted-response correlation;
for EEG, the objective is the larger of full-response and concatenated
scalp-window correlations. Ties favor higher minimum residual correlation,
then higher mean residual correlation and earlier test indices.
The selected residual correlations are $.3067/.4016$ for EEG and
$.5414/.7501$ for NSD. These are illustrative high-fidelity, diverse cases.
MEG and Alljoined retain the lower-median subject and images at ascending
residual-correlation ranks 150 and 180 of 200.
All native sensors are retained in original order.
Measured and predicted responses share one scale within each dataset.
MEG uses time-sample indices because its source tensors omit a time vector.

The main figure retains EEG and NSD; \cref{fig:supp-native-cases} unifies
EEG, MEG, and Alljoined as complete channel--time maps, with two examples
per dataset and matched image, measured-response, and prediction columns.
Scalp views elsewhere average the EEG and Alljoined raw responses over two fixed windows,
100--200 and 300--400\,ms, and interpolate the complete native electrode
set on the MNE \texttt{standard\_1020} scalp template. These are sensor-space
maps, not cortical source estimates. Sensor color limits are fixed at the
99th absolute percentile of two measured reference responses per dataset;
NSD uses a training-derived scale.
Measured and predicted cases share these fixed limits. No model-specific
amplitude normalization is applied. Raw and residual scores are calculated
over the complete sensor--time response or native voxel vector, rather than
only the displayed windows.

\begin{figure*}[p]
    \centering
    \includegraphics[width=\textwidth,height=.90\textheight,keepaspectratio]{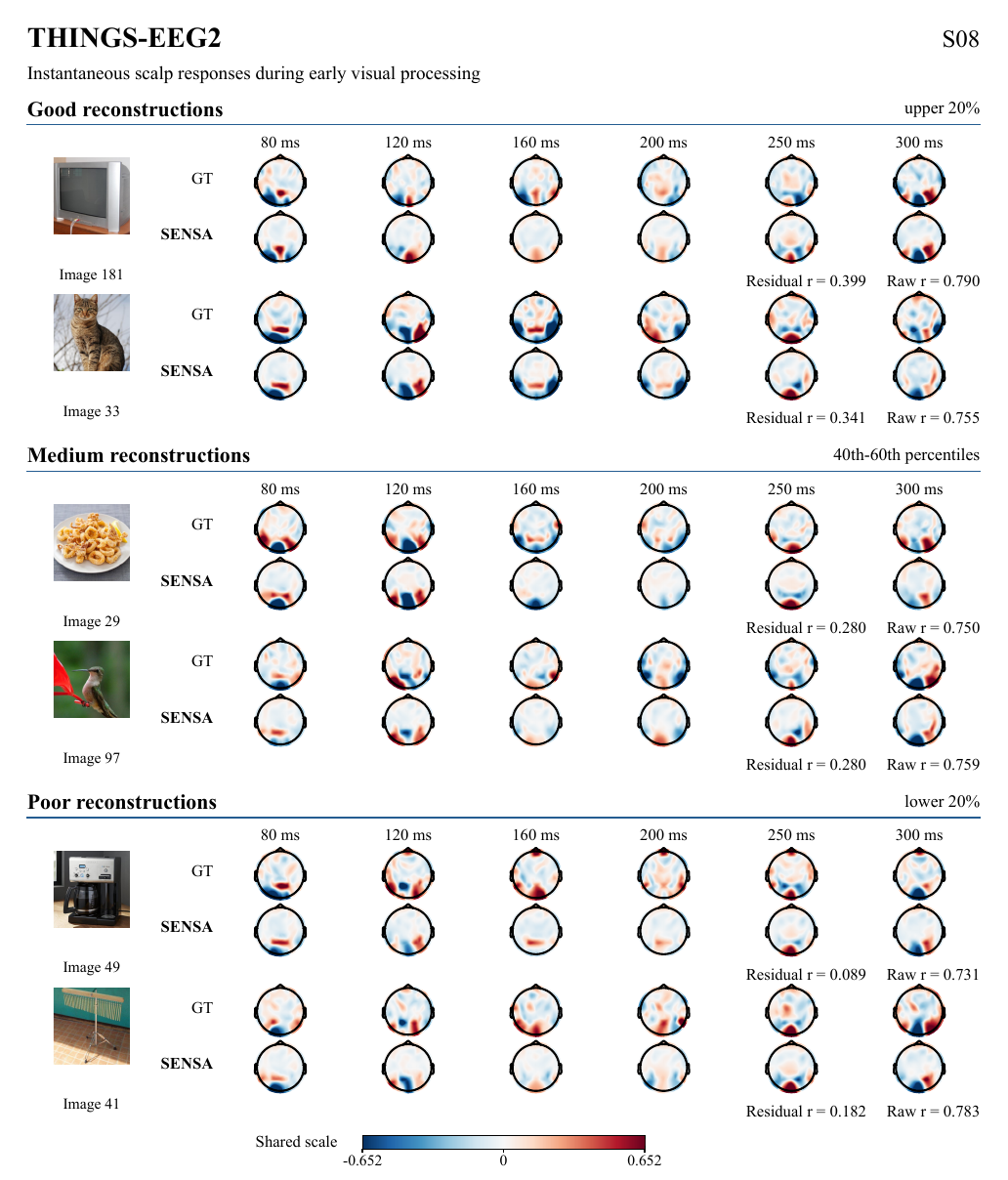}
    \caption{\textbf{THINGS-EEG2 reconstruction atlas.}
    Six images from S08, grouped by complete-response residual correlation.
    Each image shows measured (GT) and generated (SENSA) instantaneous
    scalp responses at six early visual-response times.
    All 63 electrodes use the same MNE scalp template and color scale.
    Raw and residual scores use the complete sensor--time response.}
    \label{fig:supp-atlas-eeg}
\end{figure*}

\begin{figure*}[p]
    \centering
    \includegraphics[width=\textwidth,height=.90\textheight,keepaspectratio]{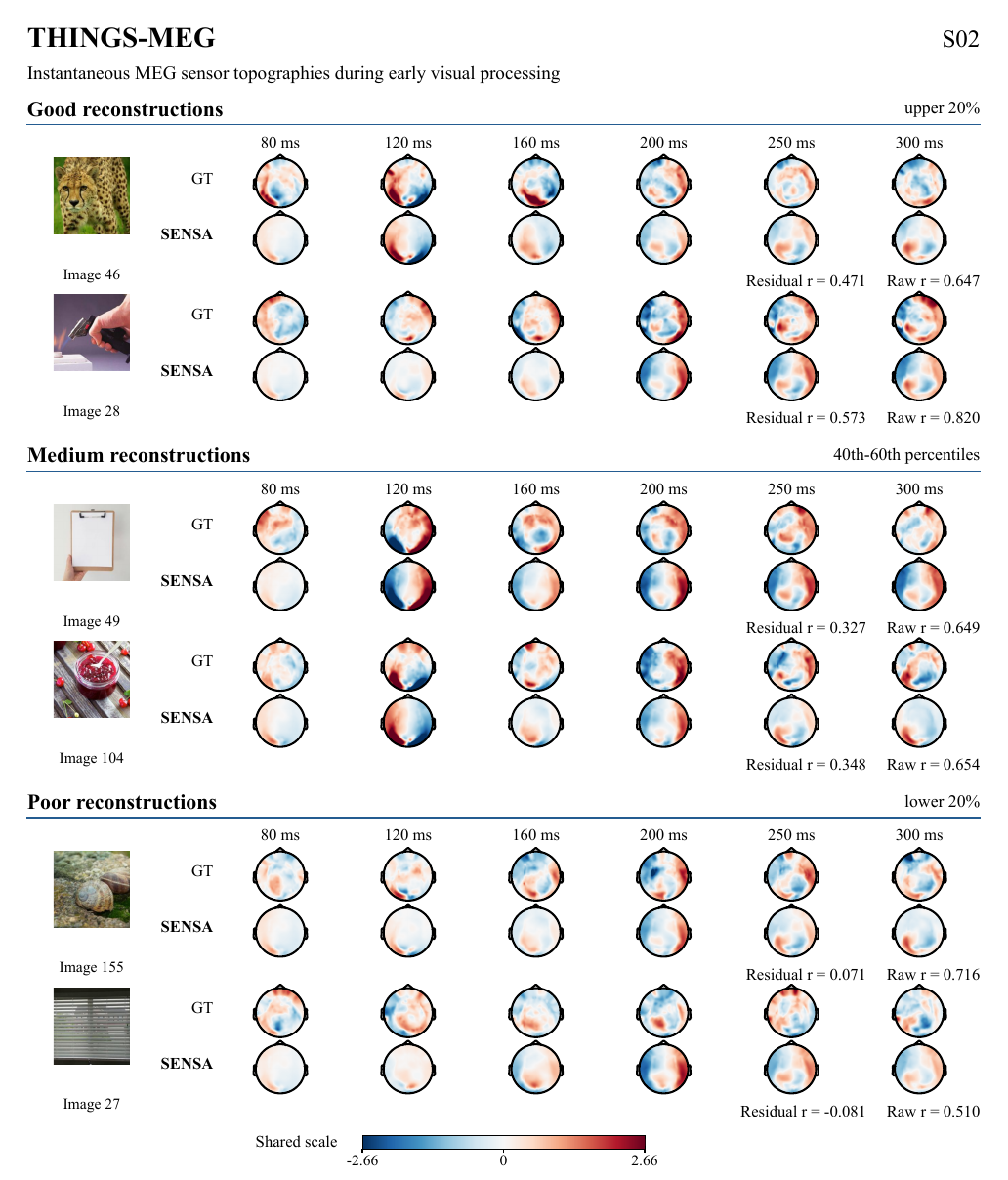}
    \caption{\textbf{THINGS-MEG reconstruction atlas.}
    Six images from S02, grouped by complete-response residual correlation.
    Measured and generated sensor topographies use all 271 native
    sensors and their original recording geometry, with a schematic
    head outline. The common scale is fixed across all cases.
    Raw and residual scores use the complete sensor--time response.}
    \label{fig:supp-atlas-meg}
\end{figure*}

\begin{figure*}[p]
    \centering
    \includegraphics[width=\textwidth,height=.90\textheight,keepaspectratio]{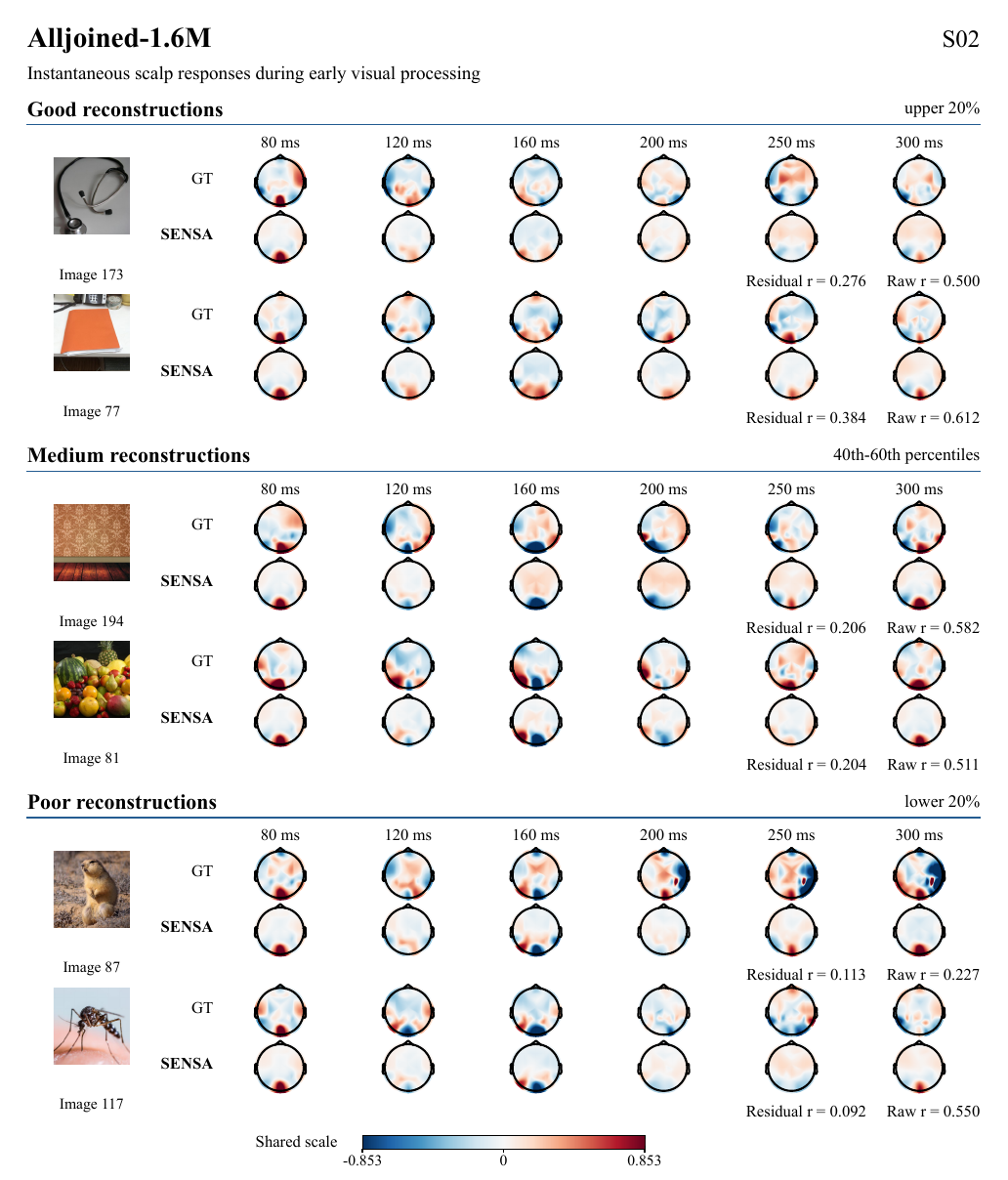}
    \caption{\textbf{Alljoined-1.6M reconstruction atlas.}
    Six images from S02, grouped by complete-response residual correlation.
    Measured and generated scalp responses retain all 32 native
    electrodes, with the same template, times, and fixed scale across
    cases. Raw and residual scores use the complete sensor--time response.}
    \label{fig:supp-atlas-alljoined}
\end{figure*}

\begin{figure*}[p]
    \centering
    \includegraphics[width=\textwidth,height=.90\textheight,keepaspectratio]{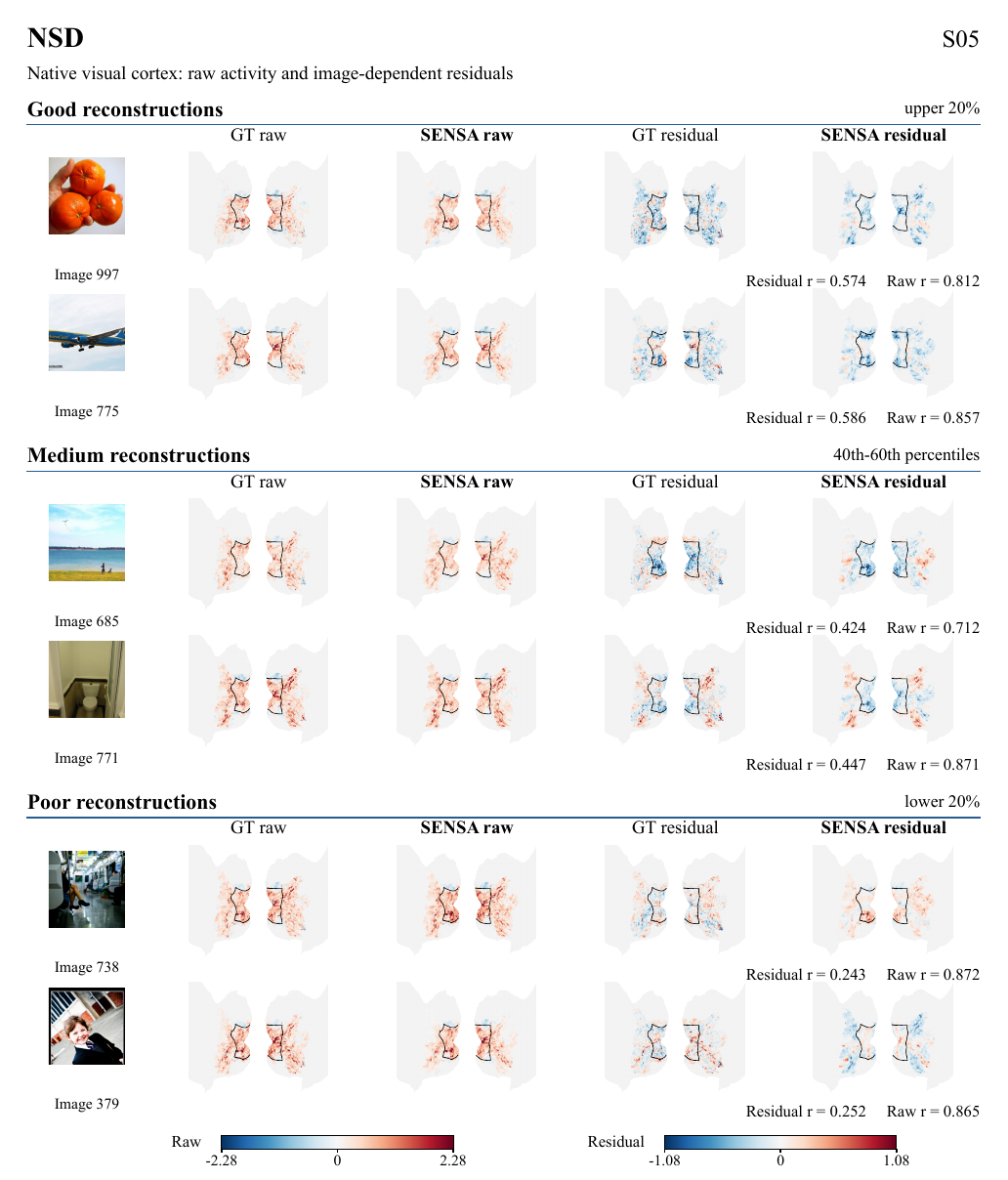}
    \caption{\textbf{NSD reconstruction atlas.}
    Six images from S05, grouped by native-voxel residual correlation.
    Each case shows bilateral visual-cortex maps of measured and generated
    raw activity and training-mean-subtracted residuals.
    Black contours mark early visual areas. Each response type has one
    fixed scale shared by all measured and generated examples;
    measurements average three repetitions, and scores use the complete
    native voxel vector.}
    \label{fig:supp-atlas-nsd}
\end{figure*}

\paragraph{Extended reconstruction atlas.}
\Cref{fig:supp-atlas-eeg,fig:supp-atlas-meg,fig:supp-atlas-alljoined,fig:supp-atlas-nsd}
provide six additional examples per dataset, using the subject with the
highest mean residual correlation over the complete official test:
EEG S08, MEG S02, Alljoined S02, and NSD S05.
The frozen prediction sources are the same as above.
Within each subject, good, medium, and poor examples occupy residual-score
ranks in the upper 20\%, 40th--60th percentiles, and lower 20\%,
respectively. Two examples per band are selected jointly to reduce
similarity among the six predicted responses, considering both raw and
training-mean-subtracted patterns and, for sensor data, the displayed
early-response samples. These are selected illustrations of three
reconstruction-quality bands.
Scores use complete responses. Each atlas fixes a common color scale at
the 99th percentile of absolute measured values over the full test:
0--350\,ms for sensor data and all native voxels for NSD.
NSD additionally fixes its residual scale from all measured test residuals
using the same percentile rule. No case-specific or prediction-specific
amplitude normalization is used.
All sensor atlases show 80, 120, 160, 200, 250, and 300\,ms.
Measured and predicted responses are linearly interpolated on their native
time axes at these exact times; scores use the original complete responses.
MEG topographies use the 271 sensor names, locations, and device-to-head
transform from S02's original preprocessed recording, retaining the
released preprocessing's channel order. A schematic head outline encloses
the projected sensor array. These are sensor-space maps.
NSD uses subject 05's official flat surfaces and nearest native
volume-to-surface transforms. Its residual panels subtract the training
mean from both measured and predicted responses.

\section{Evaluation protocols and result provenance}
\label{sec:supp-protocol}

\paragraph{Official partitions and aggregation.}
Sensor benchmarks use the complete official training and test images
specified in \cref{sec:experiments}. EEG targets are MVNN-normalized
$63\times250$ responses: four training repeats and 80 test repeats are
averaged per image. NSD retains each subject's native voxel mask and
averages three responses per test image. Sample-wise correlations flatten
each response, average over test images, then equally over subjects.
Residual scores subtract the target-training mean from predictions and
measurements. This is an evaluator statistic, not extra zero-shot input.
MSE retains the native evaluation scale of each dataset.

\paragraph{Fixed preprocessing and calibration budgets.}
THINGS-EEG2 uses the fixed UBP MVNN preprocessing: session-specific
covariance matrices are estimated from the complete target-training
recordings and applied to both training and test responses.
The $K400$ budget counts paired examples used for generator calibration;
these preprocessing matrices remain fixed across support budgets.

\paragraph{Fitting and checkpoint policy.}
Final fits use the complete official training sets.
Fits and support-based regularization choices precede test scoring;
test responses do not fit
normalizers, priors, visual projections, or model parameters.
The sensor structure study uses final fixed-budget weights: five visual
epochs for EEG, eleven for MEG, and one or two visual epochs for Alljoined.
Alljoined uses two for subjects 1/2/3/5/6/12/13/14/15/19/20 and one for the
remaining subjects, with 60 prior epochs, AdamW learning rate $.001$,
weight decay $.0001$ and batch size 128. EEG uses AdamW learning rate
$.0003$, weight decay $.0001$ and batch size 32.
These budgets are fixed from training-side experiments; full-data fits
use no internal validation split.
The final within-subject NSD model uses 15 epochs, correlation weight $.3$,
and learning rates $3\times10^{-4}$, $3\times10^{-5}$, and $3\times10^{-6}$
for epochs 1--3, 4--8, and 9--15. A direction-balanced native Ridge readout
uses training-only conditional PRESS over $10^{-4},\ldots,10^6$ and a
nonnegative training-LOO gain. The final-matched NSD ablations retain this
recipe (\cref{tab:supp-nsd-within-modules}).
Within-subject Concept2Brain on the sensor datasets first selects the
CVAE and mapper budgets using a seed-42 training-only split of 1,489/165
concepts (14,890/1,650 images for EEG and Alljoined; 17,868/1,980 for MEG),
then retrains both stages from scratch on the complete official training set.
The locked CVAE/mapper budgets are 1,300/19 epochs for EEG,
1,299/5 for MEG, and 1,300/10 for Alljoined.
The NSD native-voxel adaptation uses separate
subject models with a 50-epoch CVAE and ten-epoch mapper budget, restoring
the lowest training-loss checkpoint.
The sensor LOSO extension fixes both
source-mapper training and paired target calibration to ten epochs before
scoring; the CVAE retains its training-loss plateau rule and 1,300-epoch cap.
The NSD LOSO extension fixes source CVAE training to 50 epochs,
source-mapper training to ten, and paired target calibration to ten.
The EEG temporal, paired-image, and efficiency figures use the same
final full-training Concept2Brain refit as \cref{tab:within}.

\paragraph{Within-subject baseline evaluation.}
Completed NSD entries use the same averaged target and evaluator; the
repeat-wise alternative is retained separately. Concept2Brain NSD uses
the complete four-subject native-voxel evaluation with the fixed
short budgets described above.
CLIP-Diffusion on NSD uses a native-voxel adaptation of the released
DiT components with image-only CLIP conditioning and velocity prediction.
It trains for 50 fixed epochs on all 9,000 training images per subject
and uses the final checkpoint. Its reported prediction averages eight
independently initialized DDIM-50 samples, with this sampling rule fixed
before evaluating all 1,000 test images.
The EEG, Alljoined and NSD G{\"u}{\c{c}}l{\"u} entries and the MEG residual scores use a fixed-fc7 CNN-S
adaptation, rather than an exact replication of the original layer-selection
procedure (\cref{sec:supp-cnns}).
The EEG, Alljoined and NSD entries and the MEG residual scores in the Yamins row use fixed VGG16 fc7 features
and the same Ridge readout protocol, adapting the object-trained-CNN
encoding approach rather than reproducing the original HMO architecture.

\paragraph{Model identities and target information.}
The main EEG generator is a three-model ensemble. Image analyses and
structure comparisons use their matched single-model references.
\Cref{tab:adaptation} uses one frozen Anchor--Refiner source per sensor fold.
Response-only estimates the support mean without paired parameter fitting;
correct and shuffled pairs fit FT and Coordinate PRESS RM with the same
400 training images, repeats, initialization, and budget.
In NSD, response-only retains anatomically mapped source bases with
the support mean; paired support fits native spatial FT and RM;
zero-shot anatomically transfers source predictions
(\cref{tab:supp-nsd-zero-shot}). Complete target-information, module, and memory
comparisons appear in
\cref{tab:supp-final-information,tab:supp-final-factorial,tab:supp-coordinate-controls};
each retains its matched source and correction definition.

\section{Same-feature and target-only regression controls}
\label{sec:supp-review-controls}

\paragraph{Same visual information, unrestricted native readout.}
This control replaces the response representation with an unconstrained
Ridge map from exactly the four cached descriptor groups used by the sensor
backbone. The 2,724 input coordinates are standardized on the complete
official training images; each group is divided by the square root of its
dimension before concatenation. No output PCA or learned response basis is
used. The output prior matches the structured model: training ERP
for EEG/MEG and the fixed learned prior for Alljoined.
We solve native multi-output Ridge in float64. Regularization minimizes
conditional training-only generalized cross-validation~\cite{golub1979gcv} over
$10^{-4},10^{-3},\ldots,10^4$ times the mean diagonal of the training Gram
matrix. The fitted descriptor transforms and prior stay fixed in this
calculation. There is no internal validation split or test-based selection.
The comparison matches data, descriptors and priors; parameter counts,
optimization procedures and compute differ.

\begin{table*}[t]
    \centering
    \tableformat
    \caption{Same-feature within-subject comparison on complete official
    test sets. Every SENSA entry uses a single seed-42 model.
    Ridge is a deterministic fit to the same descriptors and prior.
    TC denotes across-test-image-centered sample-wise Pearson,
    distinct from training-ERP residual correlation.
    The main EEG ensemble is evaluated separately in \cref{tab:within}.}
    \label{tab:supp-feature-ridge}
    \begin{tabular*}{\textwidth}{@{\extracolsep{\fill}}llrrrr@{}}
    \toprule
    Dataset & Readout & Raw & Residual & TC & MSE \\
    \midrule
    EEG2 & Same-feature Ridge & .6732 & .1847 & .1930 & .0205 \\
         & Separable, seed 42 & .6771 & .2004 & .2093 & .0203 \\
    \midrule
    MEG  & Same-feature Ridge & .5610 & .2011 & .2021 & .5447 \\
         & Separable, seed 42 & .5621 & .2047 & .2046 & .5444 \\
    \midrule
    Alljoined & Same-feature Ridge & .3819 & .0723 & .0746 & .0857 \\
         & Separable, seed 42 & .3838 & .0836 & .0854 & .0856 \\
    \bottomrule
    \end{tabular*}
\end{table*}

The seed-42 separable model exceeds Ridge in 9/10 EEG, 2/4 MEG and
17/20 Alljoined subjects. Paired cohort residual differences are
$.0157$, $.0036$, and $.0113$; 95\% subject-bootstrap intervals
(10,000 draws, seed 20261104, processed EEG/MEG/Alljoined in order) are
$[.0087,.0228]$, $[-.0040,.0126]$, and $[.0070,.0158]$.
Thus the comparison supports a structural benefit for EEG and Alljoined,
and comparable MEG prediction.

\paragraph{Within-subject EEG initialization stability.}
Additional within-subject EEG fits reach residual correlation $.1996$
with seed 43 and $.2022$ with seed 44, compared with $.2004$ for seed 42.
Each score comes from a single EEG model. Cross-subject source-training
stability is evaluated separately in \cref{sec:supp-source-seeds}.

\paragraph{Population supervision at K400.}
For all ten EEG subjects, we reuse exactly the 400 support images and
support response mean from the complete Anchor--Refiner factorial study
in \cref{tab:calibration-modules}. Target-only Ridge receives the same four
cached image descriptors before brain-supervised projection, but no source
brain responses, source-trained visual-code projector or response basis.
Fixed image-only descriptor projections are retained.
Its feature centering and scaling use only these 400 support images.
We fit centered native responses with an unregularized intercept in
float64 using a dual $400\times400$ system. Conditional analytic PRESS MSE
selects regularization on the fixed relative grid
$10^{-4},10^{-3},\ldots,10^2$, scaled by mean support-Gram diagonal.
The grid is fixed before test scoring; all ten subjects select its largest
value. Conditional PRESS holds preprocessing fixed across omitted fits.
Source comparisons use the same frozen predictions as
\cref{tab:calibration-modules}.

\begin{table}[t]
    \centering
    \tableformat
    \setlength{\tabcolsep}{3pt}
    \caption{Same-support EEG $K400$ controls, ten subjects and all 200
    official test images. Source methods use the complete factorial
    source used in \cref{tab:adaptation}.}
    \label{tab:supp-target-ridge}
    \begin{tabular*}{\columnwidth}{@{\extracolsep{\fill}}lrrrr@{}}
    \toprule
    Method & Raw & Resid. & TC & MSE \\
    \midrule
    Support mean only & .6398 & -.0029 & .0000 & .0222 \\
    Target-only Ridge & .6419 & .0272 & .0731 & .0221 \\
    Source + support mean & .6476 & .0865 & .1429 & .0218 \\
    Source + FT & .6496 & .0979 & .1558 & .0217 \\
    FT + bounded RM & .6505 & .1070 & .1628 & .0217 \\
    FT + Coord.\ RM & .6508 & .1091 & .1649 & .0217 \\
    \bottomrule
    \end{tabular*}
\end{table}

Source FT+Coordinate RM exceeds target-only Ridge in all ten subjects,
with mean residual gain $.0819$ and 95\% subject-bootstrap interval
$[.0752,.0893]$ (10,000 draws, seed 20261104).
The unadapted source with support mean already exceeds it by $.0593$
$[.0551,.0623]$, separating source representation value from subsequent
calibration gains. The constant support-mean control here differs from
response-only calibration in \cref{tab:adaptation}, which retains
image prediction and estimates native statistics.

\begin{table}[!t]
    \centering
    \tableformat
    \caption{\textbf{Source-transfer controls for every held EEG subject.}
    Residual Pearson on the same 200 official images.
    All fits share the 400 support identities.
    Source mean retains the frozen shared generator and adds the support
    response mean; FT+RM uses the adopted Coordinate PRESS memory.
    These subjects also form \cref{fig:supp-source-transfer}.}
    \label{tab:supp-target-ridge-subjects}
    \begin{tabular*}{\columnwidth}{@{\extracolsep{\fill}}lrrrr@{}}
        \toprule
        Subject & Target Ridge & Source mean & FT & FT+RM \\
        \midrule
        01 & .0055 & .0695 & .0760 & .0853 \\
        02 & .0099 & .0690 & .0750 & .0810 \\
        03 & .0375 & .1008 & .1120 & .1233 \\
        04 & .0102 & .0725 & .0854 & .0905 \\
        05 & .0255 & .0679 & .0804 & .0912 \\
        06 & .0393 & .0992 & .1070 & .1154 \\
        07 & .0546 & .1192 & .1307 & .1508 \\
        08 & .0229 & .0833 & .1047 & .1264 \\
        09 & .0423 & .1016 & .1073 & .1137 \\
        10 & .0238 & .0818 & .1009 & .1131 \\
        \midrule
        Mean & .0272 & .0865 & .0979 & .1091 \\
        \bottomrule
    \end{tabular*}
\end{table}

\begin{table}[t]
    \centering
    \tableformat
    \caption{Per-subject residual correlation for the same-feature
    within-subject control. Each pair is Ridge and seed-42 separable;
    -- denotes no subject in that cohort.}
    \label{tab:supp-feature-ridge-subjects}
    \begin{tabular*}{\columnwidth}{@{\extracolsep{\fill}}rrrrrrr@{}}
    \toprule
    & \multicolumn{2}{c}{EEG2} & \multicolumn{2}{c}{MEG}
    & \multicolumn{2}{c}{Alljoined} \\
    Subject & Ridge & Sep. & Ridge & Sep. & Ridge & Sep. \\
    \midrule
     1 & .1638 & .1818 & .1752 & .1740 & .0870 & .1023 \\
     2 & .1378 & .1533 & .2930 & .2863 & .1718 & .2004 \\
     3 & .1952 & .2096 & .1802 & .1975 & .1091 & .1353 \\
     4 & .1603 & .1939 & .1558 & .1611 & .0883 & .0940 \\
     5 & .1539 & .1513 & -- & -- & .0653 & .0902 \\
     6 & .1916 & .1923 & -- & -- & .0823 & .1044 \\
     7 & .2197 & .2532 & -- & -- & .0297 & .0384 \\
     8 & .2530 & .2645 & -- & -- & .0851 & .0821 \\
     9 & .1631 & .1736 & -- & -- & -.0318 & -.0217 \\
    10 & .2082 & .2301 & -- & -- & .0800 & .0825 \\
    11 & -- & -- & -- & -- & .0624 & .0675 \\
    12 & -- & -- & -- & -- & .0872 & .0955 \\
    13 & -- & -- & -- & -- & .0599 & .0708 \\
    14 & -- & -- & -- & -- & .1297 & .1566 \\
    15 & -- & -- & -- & -- & .0412 & .0532 \\
    16 & -- & -- & -- & -- & .0260 & .0314 \\
    17 & -- & -- & -- & -- & .0432 & .0605 \\
    18 & -- & -- & -- & -- & .0931 & .0878 \\
    19 & -- & -- & -- & -- & .0271 & .0249 \\
    20 & -- & -- & -- & -- & .1088 & .1160 \\
    \bottomrule
    \end{tabular*}
\end{table}

\begin{figure}[!t]
    \centering
    \includegraphics[trim=0 0 124.2bp 0,clip,width=.65\columnwidth]{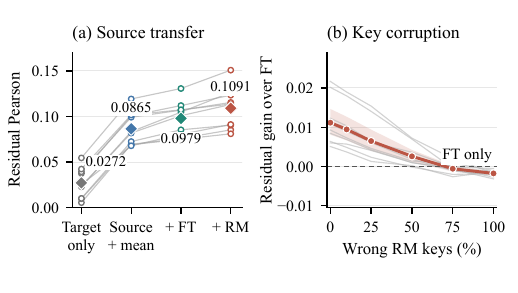}
    \caption{\textbf{Source supervision supports target calibration.}
    Ten EEG subjects, source seed 42, matched $K400$ supports and all 200
    official test images. Lines: subjects; diamonds: cohort means.
    Target-only Ridge uses the same four descriptors without source brain
    supervision. Source + mean retains the frozen shared predictor and
    estimates the target support mean; FT and RM add paired calibration.}
    \label{fig:supp-source-transfer}
\end{figure}

\paragraph{Partial errors in memory associations.}
To isolate memory behavior, keep the correctly fitted FT, support
mean, source checkpoint and 400 support responses fixed, and corrupt only
the image keys associated with remaining support errors.
This differs from refitting the complete adapter with incorrect pairs in
\cref{tab:adaptation,tab:supp-final-information}.
We predefine fractions 0, .1, .25, .5, .75 and 1, with three permutations
per subject and all ten EEG subjects. A random support ordering, seeded
$20261104+1000s+d$ for subject $s$ and draw $d\in\{0,1,2\}$, selects the
affected images; a cyclic shift on that subset creates exactly the stated
number of mismatches.
Both memory algorithms use the same fitting settings across corruption
levels, including Coordinate PRESS's null candidate.

\begin{table*}[t]
    \centering
    \tableformat
    \caption{Memory-key association stress test with correctly fitted FT
    frozen. All ten EEG subjects and all 200 official test images;
    mean of three predetermined corruption draws.
    Frozen FT without memory has Raw $.6496$, Residual $.0979$, TC $.1558$
    and MSE $.0217$. This is not end-to-end pairing corruption.}
    \label{tab:supp-memory-corruption}
    \begin{tabular*}{\textwidth}{@{\extracolsep{\fill}}l*{8}{r}@{}}
    \toprule
    & \multicolumn{4}{c}{Bounded RM}
    & \multicolumn{4}{c}{Coordinate PRESS RM (adopted)} \\
    \cmidrule(lr){2-5}\cmidrule(lr){6-9}
    Wrong keys & Raw & Residual & TC & MSE & Raw & Residual & TC & MSE \\
    \midrule
    0\%   & .6505 & .1070 & .1628 & .0217 & .6508 & .1091 & .1649 & .0217 \\
    10\%  & .6502 & .1051 & .1605 & .0217 & .6506 & .1074 & .1636 & .0217 \\
    25\%  & .6498 & .1025 & .1581 & .0217 & .6502 & .1044 & .1606 & .0217 \\
    50\%  & .6494 & .0993 & .1549 & .0217 & .6495 & .1005 & .1563 & .0217 \\
    75\%  & .6489 & .0974 & .1522 & .0217 & .6488 & .0973 & .1520 & .0217 \\
    100\% & .6486 & .0966 & .1505 & .0218 & .6484 & .0962 & .1495 & .0218 \\
    \bottomrule
    \end{tabular*}
\end{table*}

Coordinate PRESS retains higher residual correlation at 0--50\% wrong
keys. At 75--100\%, the bounded-gain control is slightly higher, but both
fall below the frozen no-memory FT reference. The adopted memory improves
the clean and moderately corrupted associations; severe key mismatching
removes that advantage in this isolated stress test.

\section{Simple within-subject controls}

\label{sec:supp-complete-within}

\Cref{tab:supp-local-within} reports mean-response and CLIP-feature controls
on complete subject cohorts. The training mean has zero residual correlation
by construction, and Ridge on CLIP features is the strongest of these controls
(residual $.1117$ on EEG, $.0462$ on Alljoined, and $.2804$ on NSD).

\paragraph{Generative baseline implementations.}
Sensor baselines use released model components adapted to native response
shapes and image-only conditioning. EEG CLIP-Diffusion uses 50 epochs,
the final checkpoint, no captions, and 63-to-64 channel padding to obtain
1,008 tokens.

\begin{table*}[!t]
\centering
\tableformat
\setlength{\tabcolsep}{2pt}
\caption{Within-subject mean-response and CLIP-feature controls. Each row evaluates the complete subject cohort on the official test set. Ridge tuning uses training data only; NSD targets average three response repeats.}
\label{tab:supp-local-within}
\begin{tabular*}{\textwidth}{@{\extracolsep{\fill}}llrrrr@{}}
\toprule
Dataset & Method & Subjects & Raw $\uparrow$ & Resid.\ $\uparrow$ & MSE $\downarrow$ \\
\midrule
THINGS-EEG2 & Training mean & 10 & .6599 & .0000 & .0212 \\
THINGS-EEG2 & Pooled CLIP Ridge & 10 & .6621 & .1117 & .0211 \\
THINGS-EEG2 & Pooled CLIP MLP & 10 & .4176 & .0218 & .0527 \\
Alljoined-1.6M & Training mean & 20 & .3685 & .0000 & .0835 \\
Alljoined-1.6M & CLIP Ridge & 20 & .3697 & .0462 & .0834 \\
Alljoined-1.6M & CLIP MLP & 20 & .1573 & .0078 & .1554 \\
NSD (three-repeat mean) & Training mean & 4 & .8163 & .0000 & .0587 \\
NSD (three-repeat mean) & CLIP Ridge & 4 & .8148 & .2804 & .0583 \\
\bottomrule
\end{tabular*}
\end{table*}

\section{NSD comparisons under repeat-wise scoring}
\label{sec:supp-nsd-release}

The main comparison in \cref{tab:within} averages the three measured responses
to each NSD test image before scoring. Here we retain the repeat-wise convention
used for SynBrain's released-output evaluation: each predicted response is
scored against the separate measured repeats, and the scores are aggregated.
Both conventions use the same four subjects (1, 2, 5, and 7), native voxel
masks, and 1,000 test images. The difference is the aggregation of measured
responses, not a change in predictions.

\subsection{Within-subject comparison}

\Cref{tab:nsd-repeat-within} evaluates the within-subject predictions from
\cref{tab:within} under repeat-wise scoring. SENSA uses all 9,000 official
training images per subject. SynBrain uses the authors' released predictions;
its values below are recomputed under the common evaluator rather than copied
at the precision of the published table.

\begin{table}[!htbp]
    \centering
    \tableformat
    \setlength{\tabcolsep}{3.5pt}
    \caption{NSD within-subject comparison under repeat-wise scoring.
    Raw and Resid.\ denote raw and training-response-residual Pearson.
    All values are four-subject macro averages.}
    \label{tab:nsd-repeat-within}
    \begin{tabular*}{\columnwidth}{@{\extracolsep{\fill}}lrrr@{}}
        \toprule
        Method & Raw $\uparrow$ & Resid.\ $\uparrow$ & MSE $\downarrow$ \\
        \midrule
        G{\"u}{\c{c}}l{\"u} CNN-S
            & .6118 & .1238 & .1688 \\
        VGG16 Ridge
            & .6089 & .1214 & .1706 \\
        SynBrain~\cite{mai2025synbrain}
            & .6866 & \textbf{.2406} & .1386 \\
        \method
            & \textbf{.7002} & .2353 & \textbf{.1270} \\
        \bottomrule
    \end{tabular*}
\end{table}

\section{Complete unseen-subject calibration}

\label{sec:supp-complete-calibration}

\subsection{External-method calibration}
\label{sec:supp-external-calibration}

The Concept2Brain extension covers all ten EEG, four MEG,
twenty Alljoined, and four NSD folds.
The source image mapper and paired target calibration each use ten fixed
epochs; target calibration jointly updates the mapper, decoder, and
subject-conditioning weights. The CVAE is fitted only on source subjects.
On EEG it reaches raw/residual correlation $.4612/.0184$ zero-shot and
$.5927/.0450$ after calibration, with MSE $.0299$ and $.0260$.
On MEG it reaches raw/residual correlation $.3422/.0528$ zero-shot and
$.4729/.0631$ after calibration, with MSE $.7258$ and $.6542$, respectively.
On Alljoined it reaches raw/residual correlation $.2192/.0037$ zero-shot
and $.3107/.0206$ after calibration, with MSE $.0937$ and $.0907$, respectively.
On NSD, the native-voxel paired $K400$ extension reaches raw/residual
correlation $.8190/.1587$ and MSE $.0568$ on the three-repeat-averaged
targets. Its source CVAE uses 50 fixed epochs, followed by ten mapper
epochs and ten target-calibration epochs.

The SynBrain LOSO extension in \cref{tab:supp-external-calibration}
trains its population model on the other subjects' complete official
training sets and calibrates with the same 400 target-image identities as
SENSA. It retains SynBrain's generator and its fixed source/target fitting
budgets; equal support does not imply equal training compute.
All ten EEG, four MEG, twenty Alljoined, and four NSD subjects are included,
with no internal validation.
NSD retains subject-specific voxel masks with different voxel counts and
spatial correspondences. All three zero-shot extensions use the same fixed
MNI nearest-neighbor transfer of source predictions into the held subject's
native space, with no target-response fitting
(\cref{sec:supp-nsd-implementation,tab:supp-nsd-zero-shot}).
Paired calibration instead fits a target-native prediction path.
NSD targets average three repeats; the repeat-wise columns rescore the same
saved predictions.

On EEG, SynBrain's residual correlation changes from $.0109$ to $.0120$,
while raw correlation falls from $.3342$ to $.3230$ and MSE rises from
$.0351$ to $.0804$; SENSA reaches $.0374$ and $.1091$ in the same cells.
On MEG, SynBrain's raw correlation rises from $.2451$ to $.2732$, while
residual correlation falls from $.0438$ to $.0249$ and MSE rises from
$1.3127$ to $2.6067$. On Alljoined, SynBrain attains residual correlation
$-.0059$ zero-shot and $.0020$ after calibration, compared with SENSA's
$.0126$ and $.0438$. On NSD, the paired extension attains residual
correlation $.2623$, compared with SENSA's $.2995$ under the averaged
target protocol. These are cross-subject extensions, separate from
published within-subject scores.

\begin{table*}[!t]
    \centering
    \tableformat
    \setlength{\tabcolsep}{3pt}
    \caption{\textbf{Completed LOSO calibration.}
    Every fold uses the official test set and the locked $K400$ support
    identities. Macro rows weight subjects equally; EEG and Alljoined report
    complete-cohort macro averages. NSD zero-shot uses fixed anatomical
    transfer (\cref{tab:supp-nsd-zero-shot}).
    Concept2Brain reports paired $K400$ with three-repeat-averaged targets;
    the final SynBrain block additionally rescores the same predictions
    against individual repeats.}
    \label{tab:supp-external-calibration}
    \begin{tabular*}{\textwidth}{@{\extracolsep{\fill}}ll*{6}{r}@{}}
        \toprule
        & & \multicolumn{3}{c}{Zero-shot}
        & \multicolumn{3}{c}{Paired $K400$} \\
        \cmidrule(lr){3-5}\cmidrule(lr){6-8}
        Dataset & Subject & Raw & Resid. & MSE & Raw & Resid. & MSE \\
        \midrule
        \multicolumn{8}{l}{\textit{Concept2Brain}} \\
        THINGS-EEG2 & 01 & .4843 & .0321 & .0297 & .6111 & .0199 & .0245 \\
         & 02 & .5078 & -.0592 & .0398 & .6541 & .0227 & .0317 \\
         & 03 & .4027 & -.0269 & .0266 & .6499 & .0444 & .0181 \\
         & 04 & .5216 & .0315 & .0349 & .3819 & .0214 & .0540 \\
         & 05 & .4522 & .0153 & .0279 & .5542 & .0458 & .0244 \\
         & 06 & .4112 & .0118 & .0259 & .5693 & .0503 & .0209 \\
         & 07 & .4942 & .0067 & .0306 & .6450 & .0595 & .0245 \\
         & 08 & .5228 & .0697 & .0278 & .7118 & .0614 & .0194 \\
         & 09 & .3248 & .0431 & .0245 & .5110 & .0725 & .0178 \\
         & 10 & .4906 & .0597 & .0311 & .6385 & .0521 & .0248 \\
         & Macro & .4612 & .0184 & .0299 & .5927 & .0450 & .0260 \\
        \midrule
        THINGS-MEG & 01 & .3715 & .0233 & .4421 & .5019 & .0357 & .3672 \\
         & 02 & .4126 & .0724 & .7832 & .5614 & .1384 & .6878 \\
         & 03 & .2861 & .0722 & 1.1845 & .4413 & .0272 & 1.1555 \\
         & 04 & .2984 & .0431 & .4933 & .3868 & .0512 & .4062 \\
         & Macro & .3422 & .0528 & .7258 & .4729 & .0631 & .6542 \\
        \midrule
        Alljoined-1.6M & Macro & .2192 & .0037 & .0937 & .3107 & .0206 & .0907 \\
        \midrule
        NSD & 01 & .2397 & .0381 & .1343 & .8330 & .1573 & .0341 \\
         & 02 & .1908 & .0295 & .1810 & .7897 & .1626 & .0579 \\
         & 05 & .2322 & .0179 & .2857 & .8451 & .1784 & .0802 \\
         & 07 & .2498 & .0339 & .1665 & .8081 & .1364 & .0551 \\
         & Macro & .2281 & .0298 & .1919 & .8190 & .1587 & .0568 \\
        \midrule
        \multicolumn{8}{l}{\textit{SynBrain}} \\
        THINGS-EEG2 & Macro & .3342 & .0109 & .0351 & .3230 & .0120 & .0804 \\
        \midrule
        THINGS-MEG & 01 & .2759 & .0354 & 1.1816 & .2599 & .0249 & 1.8987 \\
         & 02 & .2757 & .0545 & 1.3312 & .3960 & .0409 & 2.8739 \\
         & 03 & .2238 & .0554 & 1.6165 & .2277 & .0234 & 3.9571 \\
         & 04 & .2050 & .0302 & 1.1218 & .2090 & .0102 & 1.6971 \\
         & Macro & .2451 & .0438 & 1.3127 & .2732 & .0249 & 2.6067 \\
        \midrule
        Alljoined-1.6M & Macro & .1621 & -.0059 & .0956 & .0865 & .0020 & .1949 \\
        \midrule
        & & \multicolumn{3}{c}{Paired $K400$, three-repeat mean}
        & \multicolumn{3}{c}{Paired $K400$, repeat-wise} \\
        \cmidrule(lr){3-5}\cmidrule(lr){6-8}
        Dataset & Subject & Raw & Resid. & MSE & Raw & Resid. & MSE \\
        \midrule
        NSD & 01 & .8099 & .2565 & .0403 & .6856 & .1736 & .0838 \\
         & 02 & .7663 & .2467 & .0672 & .6291 & .1672 & .1442 \\
         & 05 & .8314 & .3232 & .0906 & .7097 & .2206 & .1959 \\
         & 07 & .7810 & .2229 & .0654 & .6348 & .1487 & .1449 \\
         & Macro & .7972 & .2623 & .0659 & .6648 & .1775 & .1422 \\
        \bottomrule
    \end{tabular*}
\end{table*}

\paragraph{Anatomical zero-shot transfer on NSD.}
\Cref{tab:supp-nsd-zero-shot} evaluates the frozen source models without
paired target calibration. Residual Pearson is $.0459$, $.0298$, and
$.0460$ for SynBrain, Concept2Brain, and SENSA, respectively.
SENSA and SynBrain have similar residual scores; paired $K400$ fitting
raises them to $.2995$ and $.2623$ in \cref{tab:adaptation}.
These anatomical extensions retain all native voxels and all 1,000
official test images per subject.

\begin{table*}[!t]
    \centering
    \tableformat
    \setlength{\tabcolsep}{4pt}
    \caption{\textbf{NSD zero-shot with a common anatomical transfer.}
    Source-only checkpoints, zero paired target images, and fixed MNI
    nearest-neighbor mapping averaged over three source subjects.
    Dataset-provided masks and registration are available; target responses,
    means, and scales do not enter prediction. Three test repeats are
    averaged. TC centers predictions and measurements separately across
    test images before Pearson scoring.
    These are anatomical LOSO extensions of the evaluated implementations.}
    \label{tab:supp-nsd-zero-shot}
    \begin{tabular*}{\textwidth}{@{\extracolsep{\fill}}llrrrr@{}}
        \toprule
        Method & Subject & Raw $\uparrow$ & Resid.\ $\uparrow$
        & TC $\uparrow$ & MSE $\downarrow$ \\
        \midrule
        SynBrain & 01 & .1371 & .0631 & .1023 & .1435 \\
         & 02 & .1265 & .0444 & .0900 & .1681 \\
         & 05 & .1207 & .0334 & .1179 & .2981 \\
         & 07 & .1384 & .0428 & .0865 & .1682 \\
         & Macro & .1307 & .0459 & .0992 & .1945 \\
        \midrule
        Concept2Brain & 01 & .2397 & .0381 & .0975 & .1343 \\
         & 02 & .1908 & .0295 & .0942 & .1810 \\
         & 05 & .2322 & .0179 & .1061 & .2857 \\
         & 07 & .2498 & .0339 & .0845 & .1665 \\
         & Macro & .2281 & .0298 & .0956 & .1919 \\
        \midrule
        \method & 01 & .2486 & .0567 & .1286 & .1364 \\
         & 02 & .2026 & .0460 & .1237 & .1801 \\
         & 05 & .2413 & .0313 & .1444 & .2862 \\
         & 07 & .2585 & .0499 & .1168 & .1690 \\
         & Macro & .2377 & .0460 & .1284 & .1929 \\
        \bottomrule
    \end{tabular*}
\end{table*}

\subsection{Target information for SENSA}
\label{sec:supp-final-information}

\Cref{tab:supp-final-information} compares no support, response-only,
shuffled-pair, and correct-pair calibration with all three metrics.
Sensor conditions share the frozen Anchor--Refiner source and first
predefined $K400$ support draw. Response-only adds the support mean
without fitting FT or RM; its predictions are invariant to support
pair shuffling. Correct pairs fit FT and then Coordinate PRESS RM.
The shuffled condition refits both stages with incorrect correspondence.
All sensor conditions cover complete cohorts (10/4/20 subjects).
NSD retains anatomically mapped source bases, spatial FT, and Coordinate RM
(\cref{tab:supp-nsd-factorial}).

\begin{table*}[!t]
\centering
\tableformat
\setlength{\tabcolsep}{2pt}
\caption{Target information for SENSA on the sources in
\cref{tab:adaptation}. Full-cohort macro averages, complete official test sets,
and $K400$ supports.
NSD uses three-repeat-averaged targets; no support uses anatomical transfer
(\cref{tab:supp-nsd-zero-shot}). Response-only adds the support mean to
support-centered source predictions; paired conditions fit spatial FT and RM.}
\label{tab:supp-final-information}
\begin{tabular*}{\textwidth}{@{\extracolsep{\fill}}l*{12}{r}@{}}
\toprule
& \multicolumn{3}{c}{THINGS-EEG2} & \multicolumn{3}{c}{THINGS-MEG}
& \multicolumn{3}{c}{Alljoined-1.6M} & \multicolumn{3}{c}{NSD} \\
\cmidrule(lr){2-4}\cmidrule(lr){5-7}\cmidrule(lr){8-10}\cmidrule(lr){11-13}
Condition & Raw & Resid. & MSE & Raw & Resid. & MSE
& Raw & Resid. & MSE & Raw & Resid. & MSE \\
\midrule
No support & .4959 & .0374 & .0284 & .3553 & .0588 & .7521
& .2397 & .0126 & .0929 & .2377 & .0460 & .1929 \\
Response-only & .6476 & .0865 & .0218 & .4921 & .0592 & .6348
& .3415 & .0361 & .0878 & .8089 & .1272 & .0605 \\
Shuffled & .6397 & .0261 & .0222 & .4843 & .0247 & .6496
& .3338 & .0184 & .0882 & .8046 & .0005 & .0629 \\
Correct & .6508 & .1091 & .0217 & .4976 & .0825 & .6350
& .3395 & .0438 & .0879 & .8308 & .2995 & .0532 \\
\bottomrule
\end{tabular*}
\end{table*}

\section{Complete module-level ablations}
\label{sec:supp-module-ablations}

The completed generator and native-calibration matrices use the frozen
single-seed protocol of the reported ablation runs. Every sensor condition
covers all ten EEG, four MEG, or twenty Alljoined subjects; NSD covers
subjects 1, 2, 5, and 7. Models use complete official training and test
partitions without an internal validation split. Each study below identifies
its frozen source and its relation to the matched EEG calibration study in
\cref{sec:supp-eeg-factorial}.

\subsection{Within-subject generator and loss controls}

\Cref{tab:supp-within-modules} provides all three metrics, including the two
loss controls omitted from the compact main table. Removing a feature group
zeros that branch during both training and inference while preserving the
normalization divisor of the remaining branches. The spatial, region,
multiscale-blur, and low-level groups correspond to feature-cache groups,
not four independent semantic encoders.

\begin{table*}[t]
    \centering
    \tableformat
    \setlength{\tabcolsep}{2.2pt}
    \caption{\textbf{Complete within-subject sensor-response ablations.}
    Raw and Resid.\ denote raw and training-response-residual Pearson.
    All variants use their matched dataset-specific training budgets.
    Values are subject macro averages; all correlations are maximized and
    MSE is minimized.}
    \label{tab:supp-within-modules}
    \begin{tabular*}{\textwidth}{@{\extracolsep{\fill}}l*{9}{r}@{}}
        \toprule
        & \multicolumn{3}{c}{THINGS-EEG2}
        & \multicolumn{3}{c}{THINGS-MEG}
        & \multicolumn{3}{c}{Alljoined-1.6M} \\
        \cmidrule(lr){2-4}\cmidrule(lr){5-7}\cmidrule(lr){8-10}
        Variant & Raw & Resid. & MSE
                & Raw & Resid. & MSE
                & Raw & Resid. & MSE \\
        \midrule
        Matched generator
            & .6771 & .2004 & .0203 & .5621 & .2047 & .5444 & .3838 & .0836 & .0856 \\
        w/o spatial grid
            & .6767 & .1966 & .0204 & .5615 & .1971 & .5451 & .3843 & .0828 & .0856 \\
        w/o region descriptors
            & .6771 & .2002 & .0203 & .5618 & .2035 & .5447 & .3840 & .0838 & .0856 \\
        w/o multiscale blur
            & .6770 & .2001 & .0203 & .5618 & .2025 & .5444 & .3835 & .0820 & .0856 \\
        w/o low-level descriptors
            & .6765 & .1971 & .0204 & .5617 & .2020 & .5447 & .3836 & .0828 & .0856 \\
        w/o separate prior
            & .6698 & .1520 & .0207 & .5398 & .1522 & .5701 & .1210 & -.0040 & .0952 \\
        Full waveform bank
            & .6552 & .1631 & .0215 & .5285 & .1726 & .5981 & .3514 & .0518 & .0873 \\
        w/o correlation loss
            & .6770 & .2022 & .0203 & .5618 & .2019 & .5451 & .3831 & .0809 & .0856 \\
        w/o temporal penalty
            & .6770 & .2000 & .0203 & .5621 & .2047 & .5444 & .3839 & .0834 & .0856 \\
        \bottomrule
    \end{tabular*}
\end{table*}

The full waveform bank increases trainable capacity and applies temporal
regularization to complete waveforms rather than separate temporal factors;
this is a matched-budget structural comparison, not an equal-parameter
experiment. Alljoined reuses the same frozen learned prior across visual
variants. Its no-prior condition fits raw responses under the same fixed
visual-training budget, but does not match the additional prior-pretraining
compute. Removing the separate prior causes the largest residual drop on
all three datasets (to $.1520$, $.1522$, and $-.0040$). Individual feature
and loss removals have smaller effects; EEG without correlation loss
($.2022$) and Alljoined without region descriptors ($.0838$) score slightly
higher than the matched generator.

NSD uses pooled 1,664-dimensional visual features and a native voxel head;
the four sensor feature-group removals are inapplicable to that model.
\Cref{tab:supp-nsd-within-modules} reports final-matched controls.
Linear projection removes GELU and LayerNorm while
retaining the two learned linear layers. Under the final training and readout
recipe, it reaches $.3592$ versus $.3480$ for the nonlinear reference,
improving all four subjects. This favors a compact linear projection for
the native voxel setting.
Removing the correlation objective gives $.3446$; using native
$\alpha=1$ Ridge or the learned training head gives $.3407$ or $.3246$.
Each retrained condition refits its own readout on all 9,000 training images.

\begin{table}[!htbp]
    \centering
    \tableformat
    \setlength{\tabcolsep}{2.5pt}
    \caption{\textbf{Native NSD within-subject ablations.}
    All four subjects use 9,000 official training images and 1,000 test
    images; targets are averaged over three repeats.
    Final-matched 15-epoch recipe; the reference is \cref{tab:within}.
    Linear and no-correlation conditions are retrained; readout controls
    reuse the reference checkpoint.}
    \label{tab:supp-nsd-within-modules}
    \begin{tabular*}{\columnwidth}{@{\extracolsep{\fill}}lrrr@{}}
        \toprule
        Variant & Raw & Resid. & MSE \\
        \midrule
        Final reference & .8400 & .3480 & .0507 \\
        Linear projection & .8414 & .3592 & .0500 \\
        w/o correlation loss & .8397 & .3446 & .0508 \\
        Native Ridge ($\alpha=1$) & .8388 & .3407 & .0512 \\
        Learned head, no refit & .8371 & .3246 & .0512 \\
        \bottomrule
    \end{tabular*}
\end{table}

\subsection{Native calibration across measurement geometries}
\label{sec:supp-native-calibration}

\Cref{tab:supp-native-calibration-modules} removes one adapted parameter
family at a time on the final frozen sensor Anchor--Refiner sources.
Each family is frozen at initialization before fitting FT, and Coordinate
RM is then refitted for that FT condition. The complete sensor reference
rows reproduce the final correct-pair scores in \cref{tab:adaptation}.
Every variant uses the same 400 support images, support mean, initialization,
and dataset-specific fitting budget. NSD's unanchored native-readout
controls are retained separately from the spatial FT/RM factorial in
\cref{tab:supp-nsd-factorial}.

\begin{table*}[t]
    \centering
    \tableformat
    \setlength{\tabcolsep}{3pt}
    \caption{\textbf{Calibration parameter families on the final frozen sources.}
    All held subjects (ten EEG, twenty Alljoined, four MEG, four NSD),
    identical $K=400$ support, and frozen source models.
    (a,b) Each FT intervention is followed by its own Coordinate RM fit.
    (a) EEG and Alljoined Factor Transport. (b) MEG scalp, time, and amplitude
    factors. (c) Unanchored NSD native-readout controls, scored against
    repetition-averaged test responses.}
    \label{tab:supp-native-calibration-modules}
    \begin{tabular*}{\textwidth}{@{\extracolsep{\fill}}l*{6}{r}@{}}
        \toprule
        \multicolumn{7}{l}{\textit{(a) Native Factor Transport}} \\
        & \multicolumn{3}{c}{THINGS-EEG2}
        & \multicolumn{3}{c}{Alljoined-1.6M} \\
        \cmidrule(lr){2-4}\cmidrule(lr){5-7}
        Variant & Raw & Resid. & MSE & Raw & Resid. & MSE \\
        \midrule
        FT + RM reference & .6508 & .1091 & .0217 & .3395 & .0438 & .0879 \\
        w/o channel mixing & .6506 & .1053 & .0217 & .3394 & .0426 & .0879 \\
        w/o channel gain & .6506 & .1075 & .0217 & .3395 & .0433 & .0879 \\
        w/o temporal gain & .6508 & .1088 & .0217 & .3395 & .0435 & .0879 \\
        w/o latency shift & .6507 & .1086 & .0217 & .3395 & .0437 & .0879 \\
        w/o factor gain & .6508 & .1085 & .0217 & .3395 & .0433 & .0879 \\
        w/o support centering & .6508 & .1091 & .0217 & .3395 & .0438 & .0879 \\
        \bottomrule
    \end{tabular*}

    \vspace{3pt}
    \begin{minipage}[t]{.485\textwidth}
    \vspace{0pt}
    \begin{tabular*}{\linewidth}{@{\extracolsep{\fill}}lrrr@{}}
        \toprule
        \multicolumn{4}{l}{\textit{(b) Native MEG calibration}} \\
        Variant & Raw & Resid. & MSE \\
        \midrule
        FT + RM reference & .4976 & .0825 & .6350 \\
        w/o scalp update & .5006 & .0770 & .6286 \\
        w/o time update & .4979 & .0794 & .6314 \\
        w/o amplitude update & .4985 & .0819 & .6337 \\
        w/o support centering & .4988 & .0818 & .6357 \\
        \bottomrule
    \end{tabular*}
    \end{minipage}\hfill
    \begin{minipage}[t]{.485\textwidth}
    \vspace{0pt}
    \begin{tabular*}{\linewidth}{@{\extracolsep{\fill}}lrrr@{}}
        \toprule
        \multicolumn{4}{l}{\textit{(c) Unanchored NSD readout}} \\
        Variant & Raw & Resid. & MSE \\
        \midrule
        Native reference & .8308 & .2995 & .0532 \\
        w/o source projection & .7078 & .1414 & .0990 \\
        w/o support prior & .2021 & .0998 & .2175 \\
        Support mean only & .8158 & -.0029 & .0589 \\
        \bottomrule
    \end{tabular*}
    \end{minipage}
\end{table*}

On EEG and Alljoined, removing channel mixing produces the largest residual
change, from $.1091$ to $.1053$ and from $.0438$ to $.0426$, respectively,
even after refitting RM. The separate transport families therefore contribute
to the final image-dependent correction.

MEG adapts native scalp, time, and amplitude factors rather than the
channel-mixing parameterization used for EEG and Alljoined
(\cref{tab:supp-native-calibration-modules}b). Freezing scalp or time updates
gives residual correlations $.0770$ and $.0794$, compared with $.0825$ for
the reference. Freezing amplitude updates instead gives $.0819$, with raw
correlation $.4985$ and MSE $.6337$; the preferred parameter allocation
depends on the measurement geometry.

For the unanchored NSD readout, panel (c)
separates the source-learned projection from the support mean. Removing the
source projection lowers residual correlation from $.2995$ to $.1414$,
removing the support prior lowers it to $.0998$, and the support mean alone
gives $-.0029$.

\subsection{Residual-memory controls}
\label{sec:supp-memory-controls}

The EEG memory comparisons share the locked complete source and appear with
correct and shuffled support in \cref{tab:supp-factorial-complete}.
Adopted Coordinate PRESS RM reaches residual $.1091$, compared with $.1070$
for the bounded-gain control. Ridge chooses from a fixed 21-value
regularization grid and an exact
zero-correction option. PRESS is conditional on frozen features, response
bases, and upstream predictions; it is not leave-one-out validation of the
whole adapter.

Complete sensor factorials are in \cref{tab:supp-final-factorial};
\cref{tab:supp-coordinate-controls} adds source-coordinate and key controls.
All comparisons retain their matched final sources and support identities.
NSD's matched spatial FT/RM factorial is in \cref{tab:supp-nsd-factorial};
PRESS selects a null correction after spatial FT in all four subjects.
Unanchored native-readout controls appear separately in
\cref{tab:supp-native-calibration-modules}c.

\subsection{Matched generator-head replacements}

\label{sec:supp-generator-families}

This strict LOSO EEG experiment uses ten held subjects, full official
training and test sets, and paired $K=400$ calibration. Frozen visual/anchor
inputs and source subjects are shared. Each head receives 5,000 fixed
training updates; each paired and fully shuffled native adapter receives
300 steps. The comparison matches update counts across architectures
with different FLOP costs. The deterministic architecture control uses
the cVAE inference network without its latent-variable objective.

\begin{table*}[!t]
\centering
\tableformat
\setlength{\tabcolsep}{3pt}
\caption{\textbf{Matched EEG generator-head replacements: complete metrics.}
Ten-subject macro averages on all 200 official test images.
Every supported condition uses the same $K400$ identities.
Each head retains its own correctly fitted or fully shuffled native adapter;
no Residual Memory is added.
Inference parameters are 63,167 for Direct and 70,431 for the deterministic
control and cVAE, excluding the training-only cVAE posterior (93,088).
The deterministic control allocates that posterior without using it, so
it and cVAE both allocate 163,519 parameters.}
\label{tab:supp-head-replacement}
\begin{tabular*}{\textwidth}{@{\extracolsep{\fill}}l*{6}{r}@{}}
\toprule
& \multicolumn{3}{c}{Correct pairs} & \multicolumn{3}{c}{Shuffled pairs} \\
\cmidrule(lr){2-4}\cmidrule(lr){5-7}
Head / inference & Raw & Resid. & MSE & Raw & Resid. & MSE \\
\midrule
Direct & .6487 & .1009 & .0218 & .6448 & .0616 & .0219 \\
Deterministic control & .6486 & .1007 & .0218 & .6448 & .0612 & .0219 \\
cVAE prior mean & .6475 & .0959 & .0218 & .6434 & .0558 & .0220 \\
cVAE MC16 average & .6466 & .0971 & .0219 & .6428 & .0564 & .0220 \\
cVAE single sample & .6163 & .0791 & .0237 & .6125 & .0387 & .0238 \\
cVAE zero latent & .6456 & .0933 & .0219 & .6417 & .0514 & .0221 \\
\midrule
& \multicolumn{3}{c}{Response-only support} & \multicolumn{3}{c}{No support (pure LOSO)} \\
\cmidrule(lr){2-4}\cmidrule(lr){5-7}
Head / inference & Raw & Resid. & MSE & Raw & Resid. & MSE \\
\midrule
Direct & .6468 & .0877 & .0218 & .4951 & .0385 & .0285 \\
Deterministic control & .6468 & .0874 & .0218 & .4950 & .0384 & .0285 \\
cVAE prior mean & .6456 & .0822 & .0219 & .4933 & .0360 & .0285 \\
cVAE MC16 average & .6449 & .0821 & .0219 & .4926 & .0366 & .0286 \\
cVAE single sample & .6148 & .0600 & .0237 & .4586 & .0327 & .0303 \\
cVAE zero latent & .6438 & .0775 & .0220 & .4910 & .0347 & .0286 \\
\bottomrule
\end{tabular*}
\end{table*}

In \cref{tab:supp-head-replacement}, Direct has the highest residual correlation in all four support conditions. It also has the highest raw correlation and the lowest MSE in each condition; its MSE margins over the deterministic control and the cVAE prior mean are below $.0001$. It achieves higher paired residual correlation than the cVAE prior mean and MC16 average in 9/10 subjects, than a single sample in 10/10, and than the deterministic architecture control in 8/10. The deterministic control shares the cVAE inference network yet stays within $.0004$ residual of Direct in every condition, so the cVAE deficit in this run comes from latent-variable training rather than from the network. These single-seed LOSO comparisons evaluate generation of repeat-averaged EEG responses.

\subsection{Complementary calibration across generation strategies}
\label{sec:supp-generator-ft-rm}

We test FT and Coordinate PRESS RM with SENSA and four alternative generation
strategies: a conditional VAE, residual diffusion, conditional flow matching,
and a conditional GAN.
The ten strict LOSO folds share frozen visual features, native
response factors, source-only scaling, and the same $K=400$ target-training
images and support mean. SENSA reuses the main-table source checkpoint and
the four calibration conditions in \cref{tab:calibration-modules}, retaining
its 1,200-update Refiner training. Alternative heads are trained on all
16,540 official images from the nine source subjects for 5,000 updates.
All use batch size 64 and AdamW
(learning rate $3\times10^{-4}$, weight decay $10^{-4}$).
Final-update weights are used for every head. FT is fitted for 300 steps
separately for each head, and RM is refitted for each FT setting.
The no-module condition retains the same support mean; it is not pure LOSO.

\begin{table*}[!t]
\centering
\tableformat
\setlength{\tabcolsep}{3pt}
\caption{\textbf{FT and RM across generation strategies.}
Ten-subject macro averages on the complete 200-image THINGS-EEG2 test set,
with identical paired $K=400$ support.
Direct denotes SENSA with the main-table results; alternative generators replace its
response head under the training budgets specified above.
Residual correlation subtracts the target-training ERP;
centered correlation removes the test-image mean separately from each
response set.}
\label{tab:supp-generator-ft-rm}
\begin{tabular*}{\textwidth}{@{\extracolsep{\fill}}lccrrrr@{}}
\toprule
Generator & FT & RM & Raw $\uparrow$ & Resid.\ $\uparrow$
& Centered $\uparrow$ & MSE $\downarrow$ \\
\midrule
Direct & $\times$ & $\times$ & .6476 & .0865 & .1429 & .02180 \\
        & $\checkmark$ & $\times$ & .6496 & .0979 & .1558 & .02171 \\
        & $\times$ & $\checkmark$ & .6500 & .1032 & .1590 & .02169 \\
        & $\checkmark$ & $\checkmark$ & .6508 & \textbf{.1091} & .1649 & .02166 \\
\midrule
Conditional VAE & $\times$ & $\times$ & .6451 & .0812 & .1374 & .02194 \\
                & $\checkmark$ & $\times$ & .6470 & .0943 & .1511 & .02186 \\
                & $\times$ & $\checkmark$ & .6474 & .0985 & .1542 & .02184 \\
                & $\checkmark$ & $\checkmark$ & .6482 & \textbf{.1051} & .1605 & .02181 \\
\midrule
Residual diffusion & $\times$ & $\times$ & .3575 & .0173 & .0185 & .06078 \\
                   & $\checkmark$ & $\times$ & .3604 & .0226 & .0241 & .06072 \\
                   & $\times$ & $\checkmark$ & .3600 & .0219 & .0233 & .06067 \\
                   & $\checkmark$ & $\checkmark$ & .3614 & \textbf{.0245} & .0260 & .06065 \\
\midrule
Conditional flow matching & $\times$ & $\times$ & .3856 & .0185 & .0205 & .05197 \\
                          & $\checkmark$ & $\times$ & .3888 & .0250 & .0274 & .05192 \\
                          & $\times$ & $\checkmark$ & .3883 & .0239 & .0261 & .05187 \\
                          & $\checkmark$ & $\checkmark$ & .3898 & \textbf{.0270} & .0294 & .05184 \\
\midrule
Conditional GAN & $\times$ & $\times$ & .6016 & .0489 & .0938 & .02456 \\
                & $\checkmark$ & $\times$ & .6078 & .0668 & .1219 & .02424 \\
                & $\times$ & $\checkmark$ & .6049 & .0619 & .1121 & .02440 \\
                & $\checkmark$ & $\checkmark$ & .6099 & \textbf{.0713} & .1302 & .02412 \\
\bottomrule
\end{tabular*}
\end{table*}

FT and RM are complementary for all five strategies
(\cref{tab:supp-generator-ft-rm}). Adding RM after FT increases residual
correlation by $.0112$, $.0108$, and $.0019$ for SENSA, VAE, and diffusion,
respectively; adding FT after RM increases it by $.0058$, $.0065$, and
$.0026$. Each of these incremental comparisons improves all ten subjects.
The combined paired-minus-fully-shuffled residual margins are $.0830$,
$.0806$, and $.0199$, using adapters and memories refitted along the entire
shuffled chain.

For flow matching and the conditional GAN, FT and RM together raise
residual correlation by $.0085$ and $.0225$ over their respective
no-module conditions. FT alone, RM alone, and their combination each
improve all ten subjects relative to the corresponding no-module head.
Adding RM after FT improves 10/10 flow subjects and 9/10 GAN subjects;
adding FT after RM improves all ten for both heads.
Their combined paired-minus-fully-shuffled residual margins are $.0226$
and $.0628$, respectively.

The VAE uses a 16-dimensional latent and its image-conditioned prior mean
at inference. Its KL weight ramps to $.001$ over the first 500 updates.
Diffusion predicts velocity on a cosine variance-preserving corruption of
the target-minus-anchor response, with 1,000 training noise levels and
20 deterministic DDIM steps at inference. It uses one saved Gaussian
initialization per image, without sample averaging. The alternative heads'
common update budget fixes training exposure rather than FLOPs or convergence.
Flow matching learns a linear path from Gaussian noise with standard
deviation $.15$ to the target-minus-anchor response and uses 20 Euler
steps at inference. The conditional GAN combines waveform MSE with an
image-conditioned projection discriminator trained using a hinge loss.
Its adversarial weight ramps from zero to $.01$ between updates 500 and
1,000, with one discriminator update per generator update.
Both heads use one saved noise realization per image without averaging.
The result supports calibration complementarity for SENSA and the four
alternative generation strategies.

\subsection{Subject consistency of the structure comparison}

\Cref{tab:supp-structure-subjects} gives the per-subject differences behind
\cref{tab:core-mechanisms}. The separable bank is higher in 10/10 EEG, 4/4
MEG, and 19/20 Alljoined subjects; the exception, Alljoined subject 09, is
below zero with both banks. Each subject contributes one summary over all
official test images from the matched single-seed structure study, so the
counts measure subject consistency, not independent-seed replication.

\begin{table*}[!t]
\centering
\tableformat
\setlength{\tabcolsep}{3pt}
\caption{All per-subject residual correlations for full waveform and separable banks. Delta is separable minus full, computed before rounding.}
\label{tab:supp-structure-subjects}
\begin{minipage}[t]{.485\textwidth}
\vspace{0pt}
\begin{tabular*}{\linewidth}{@{\extracolsep{\fill}}lrrr@{}}
\toprule
Subject & Full & Separable & Delta \\
\midrule
\multicolumn{4}{@{}l}{\textit{THINGS-EEG2}} \\
S01 & .1541 & .1818 & .0277 \\
S02 & .1103 & .1533 & .0431 \\
S03 & .1756 & .2096 & .0340 \\
S04 & .1430 & .1939 & .0509 \\
S05 & .1231 & .1513 & .0282 \\
S06 & .1469 & .1923 & .0454 \\
S07 & .2079 & .2532 & .0453 \\
S08 & .2313 & .2645 & .0332 \\
S09 & .1482 & .1736 & .0254 \\
S10 & .1904 & .2301 & .0397 \\
\midrule
\multicolumn{4}{@{}l}{\textit{THINGS-MEG}} \\
S01 & .1542 & .1740 & .0198 \\
S02 & .2672 & .2863 & .0190 \\
S03 & .1424 & .1975 & .0550 \\
S04 & .1266 & .1611 & .0345 \\
\bottomrule
\end{tabular*}
\end{minipage}\hfill
\begin{minipage}[t]{.485\textwidth}
\vspace{0pt}
\begin{tabular*}{\linewidth}{@{\extracolsep{\fill}}lrrr@{}}
\toprule
Subject & Full & Separable & Delta \\
\midrule
\multicolumn{4}{@{}l}{\textit{Alljoined-1.6M}} \\
S01 & .0561 & .1023 & .0462 \\
S02 & .1445 & .2004 & .0560 \\
S03 & .0797 & .1353 & .0556 \\
S04 & .0482 & .0940 & .0458 \\
S05 & .0640 & .0902 & .0261 \\
S06 & .0724 & .1044 & .0320 \\
S07 & .0179 & .0384 & .0204 \\
S08 & .0620 & .0821 & .0201 \\
S09 & -.0096 & -.0217 & -.0121 \\
S10 & .0528 & .0825 & .0296 \\
S11 & .0371 & .0675 & .0304 \\
S12 & .0587 & .0955 & .0368 \\
S13 & .0297 & .0708 & .0411 \\
S14 & .1124 & .1566 & .0443 \\
S15 & .0225 & .0532 & .0307 \\
S16 & .0203 & .0314 & .0111 \\
S17 & .0281 & .0605 & .0324 \\
S18 & .0498 & .0878 & .0380 \\
S19 & .0193 & .0249 & .0057 \\
S20 & .0697 & .1160 & .0463 \\
\bottomrule
\end{tabular*}
\end{minipage}
\end{table*}

\section{Fixed-feature CNN encoding controls}
\label{sec:supp-cnns}

The Alljoined and NSD G{\"u}{\c{c}}l{\"u} entries in \cref{tab:within}
use an adaptation of the CNN-S encoding approach~\cite{guclu2015deep}.
We freeze the original ImageNet CNN-S backbone and preselect its 4,096
post-ReLU fc7 features. Feature standardization and the response-mean
intercept use the complete official training set of each subject.
The Ridge readout fixes $\alpha=1$ before testing, without layer selection
or an internal validation split. This specifies a fixed-feature adaptation;
it does not reproduce the original voxel-wise layer-selection procedure.
Alljoined includes all twenty subjects, 16,540 training images and 200
test images each. NSD includes subjects 1/2/5/7, 9,000 training images
and 1,000 three-repeat-averaged test targets each.

VGG16 Ridge uses the same subjects, complete official splits,
full-training-set standardization, training-mean intercept, and fixed
$\alpha=1$ readout, applied to the 4,096 post-ReLU fc7 features of the
frozen ImageNet-pretrained VGG16. Since pretrained weights for the original
HMO model of Yamins et al.\ were unavailable for our evaluation, we use
VGG16 as a substitute within the same object-trained-CNN encoding paradigm.

\Cref{tab:supp-cnns-subjects} lists residual correlation for every subject,
including negative scores, under correct and permuted test correspondence.
The permutation control shuffles saved predictions with seed 42; it does
not refit the readout with shuffled training pairs. Permutation removes the
NSD residual signal of both readouts (every permuted subject lies within
$.009$ of zero). The Alljoined signal is weak: correct correspondence
exceeds the permutation in 13/20 subjects for CNN-S and 15/20 for VGG16,
and every Alljoined score lies within $\pm.026$.
Macro raw correlation and MSE appear in
\cref{tab:within}; the same NSD predictions under repeat-wise
scoring appear in \cref{tab:nsd-repeat-within}.

\begin{table}[!t]
    \centering
    \tableformat
    \setlength{\tabcolsep}{3pt}
    \caption{\textbf{Fixed-feature CNN controls by subject.}
    Residual Pearson for CNN-S and VGG16 Ridge (fixed fc7, $\alpha=1$).
    Correct and permuted test correspondence use identical saved
    predictions. NSD targets average three repeats.}
    \label{tab:supp-cnns-subjects}
    \begin{tabular*}{\columnwidth}{@{\extracolsep{\fill}}lrrrr@{}}
        \toprule
        & \multicolumn{2}{c}{CNN-S} & \multicolumn{2}{c}{VGG16 Ridge} \\
        \cmidrule(lr){2-3}\cmidrule(lr){4-5}
        Subject & Correct & Permuted & Correct & Permuted \\
        \midrule
        \multicolumn{5}{l}{\textit{Alljoined-1.6M}} \\
        01 & .0142 & -.0084 & .0048 & -.0124 \\
        02 & .0235 & .0011 & .0159 & -.0035 \\
        03 & .0008 & .0068 & .0008 & -.0135 \\
        04 & .0195 & .0021 & .0130 & -.0025 \\
        05 & -.0053 & .0252 & .0029 & .0105 \\
        06 & .0186 & -.0065 & .0195 & -.0001 \\
        07 & .0075 & -.0097 & -.0071 & -.0028 \\
        08 & .0015 & .0021 & .0111 & -.0015 \\
        09 & -.0151 & .0140 & -.0063 & -.0088 \\
        10 & -.0034 & -.0042 & .0015 & -.0008 \\
        11 & .0013 & .0007 & -.0073 & .0010 \\
        12 & .0080 & .0057 & .0039 & .0017 \\
        13 & .0142 & -.0027 & -.0085 & .0044 \\
        14 & .0056 & .0005 & .0239 & .0044 \\
        15 & -.0013 & .0020 & -.0077 & .0076 \\
        16 & .0033 & .0063 & .0149 & -.0020 \\
        17 & .0035 & .0057 & .0041 & -.0044 \\
        18 & .0183 & .0004 & .0187 & .0002 \\
        19 & .0036 & .0008 & .0043 & -.0041 \\
        20 & .0120 & .0039 & .0103 & .0043 \\
        \midrule
        \multicolumn{5}{l}{\textit{NSD}} \\
        01 & .1884 & .0045 & .1795 & .0077 \\
        02 & .1784 & .0021 & .1701 & .0082 \\
        05 & .2204 & -.0020 & .2219 & -.0028 \\
        07 & .1434 & -.0022 & .1432 & -.0053 \\
        \bottomrule
    \end{tabular*}
\end{table}

\section{Matched sensor calibration and memory controls}
\label{sec:supp-eeg-factorial}

\paragraph{Matched sources and supports.}
The FT$\times$RM experiment uses the frozen Anchor--Refiner population
checkpoints, seed 42, all 10 EEG, 4 MEG, and 20 Alljoined LOSO folds, and 400 distinct support
images from each held subject's complete official training set.
Each fold excludes the held subject from source training and evaluates
all 200 official test images. All conditions retain the same
support-response mean; Off/Off is therefore a support-matched reference.
RM is fitted to the remaining error of its own FT condition, including
a fresh fit when FT is off. Shuffled conditions refit the entire
calibration path with incorrect support correspondence and the same
budget. Neither source-generator retraining nor test-based checkpoint
selection is introduced.

\paragraph{Coordinate PRESS memory.}
\Cref{tab:supp-final-factorial} reports the complete sensor sources using
Coordinate PRESS RM. At the first predefined support draw, full-model
residual correlation is $.1091$, $.0825$, and $.0438$ on EEG, MEG, and
Alljoined. EEG improves over FT alone by $.0112$ and over the bounded-gain
control by $.0021$, both in all ten subjects.
The FT-off fits reach $.1032$, $.0821$, and $.0420$.
The complete model exceeds either single module in all ten EEG subjects.
Alljoined improves over FT-only in 11/20 subjects and over RM-only in 12/20.
Its residual gain accompanies a raw-correlation decrease
($.3425\to.3395$) and MSE increase ($.0877\to.0879$).
On MEG, FT reaches $.0806$ and Coordinate RM raises it to $.0825$;
RM-only reaches $.0821$, with higher raw correlation and lower MSE.
\Cref{tab:supp-factorial-complete} includes correct and shuffled
Coordinate RM fits with FT both off and on.
It also retains the bounded-gain factorial and same-upstream memory
controls; individual final-model scores appear in
\cref{tab:supp-coordinate-subjects}.
These share the EEG source in \cref{tab:adaptation}.

\begin{figure}[!tbp]
    \centering
    \includegraphics[width=\columnwidth]{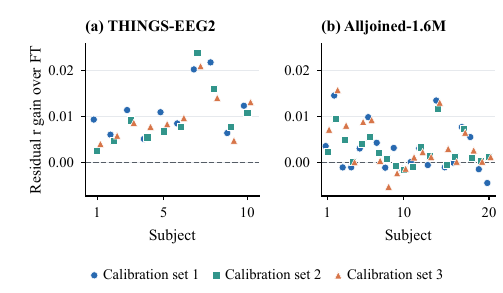}
    \caption{\textbf{Residual correction across support samples.}
    FT+Coordinate RM minus FT in (a) EEG and (b) Alljoined.
    Three predefined $K400$ draws; all subjects and 200 official test images.
    Each matched pair shares its source, FT, and support mean.
    The draws vary supports for one frozen source seed.}
    \label{fig:support-gains}
\end{figure}

\paragraph{Support stability and memory structure.}
Three predefined $K=400$ support draws reuse the same source checkpoints.
For each draw, adopted FT+RM exceeds FT alone in all ten EEG subjects,
with macro gains $.0112$, $.0094$, and $.0097$.
Alljoined gains $.0028$, $.0027$, and $.0036$, with positive changes in
11/20, 16/20, and 17/20 subjects, respectively.
MEG gains $.0019$, $.0019$, and $.0012$, with positive changes in
2/4, 4/4, and 3/4 subjects. The EEG and Alljoined gains are shown in
\cref{fig:support-gains}.
These are support resamples of one source seed, not independent training
seeds.

\begin{table*}[!t]
    \centering
    \tableformat
    \setlength{\tabcolsep}{3pt}
    \caption{\textbf{Matched final-source calibration across sensor datasets.}
    First predefined draw, all official test images, $K400$ supports.
    Coordinate PRESS is the adopted RM; each FT-off memory is refitted.
    Bounded-gain results are retained as separately labeled controls.
    Rows share the source; all supported rows retain the same support mean.
    Full-model scores supply the complete row of \cref{tab:adaptation}.}
    \label{tab:supp-final-factorial}
    \begin{tabular*}{\textwidth}{@{\extracolsep{\fill}}cc*{9}{r}@{}}
        \toprule
        & & \multicolumn{3}{c}{THINGS-EEG2}
          & \multicolumn{3}{c}{THINGS-MEG}
          & \multicolumn{3}{c}{Alljoined-1.6M} \\
        \cmidrule(lr){3-5}\cmidrule(lr){6-8}\cmidrule(lr){9-11}
        FT & RM & Raw & Resid. & MSE
                & Raw & Resid. & MSE & Raw & Resid. & MSE \\
        \midrule
        -- & No support & .4959 & .0374 & .0284 & .3553 & .0588 & .7521 & .2397 & .0126 & .0929 \\
        Off & None & .6476 & .0865 & .0218 & .4921 & .0592 & .6348 & .3415 & .0361 & .0878 \\
        On & None & .6496 & .0979 & .0217 & .4981 & .0806 & .6344 & .3425 & .0410 & .0877 \\
        Off & Coordinate & .6500 & .1032 & .0217 & .5038 & .0821 & .6253 & .3388 & .0420 & .0880 \\
        On & Coordinate & .6508 & .1091 & .0217 & .4976 & .0825 & .6350 & .3395 & .0438 & .0879 \\
        \midrule
        Off & Bounded & .6494 & .1001 & .0217 & .4980 & .0744 & .6300 & .3387 & .0399 & .0880 \\
        On & Bounded & .6505 & .1070 & .0217 & .4977 & .0825 & .6355 & .3394 & .0426 & .0879 \\
        \bottomrule
    \end{tabular*}
\end{table*}

\paragraph{Native spatial calibration on NSD.}
\Cref{tab:supp-nsd-factorial} adds the NSD counterpart of
\cref{tab:calibration-modules}. The frozen source heads are first mapped
into the target anatomy; native spatial FT updates these response bases,
and RM retrieves errors in their fixed source-derived coordinates
(\cref{sec:supp-nsd-implementation}).
With identical $K400$ supports and support means, spatial FT raises residual
correlation from $.1272$ to $.2995$ in all four subjects.
RM alone reaches $.1371$, improving all four FT-off references.
After FT, all 64 directions select the zero-correction candidate in every
subject; full and FT-only predictions are identical. Thus NSD supports the
spatial calibration component, while the additional RM increment is
established on the sensor datasets.
The unanchored native-readout control reaches $.2995$ at the displayed
precision and is kept separate from the four matched cells.

\begin{table*}[!t]
    \centering
    \tableformat
    \setlength{\tabcolsep}{4pt}
    \caption{\textbf{NSD source-anchored spatial FT/RM factorial.}
    Same LOSO source checkpoints, anatomical maps, and $K400$ support
    identities; three-repeat-averaged targets, all native voxels and
    all 1,000 official test images per subject. (a) Four-subject macro
    averages; TC is test-image-centered Pearson. (b) Per-subject residual
    Pearson. All supported rows retain the same support mean.
    RM is freshly fitted for each FT setting; full-chain shuffling refits
    both stages. The unanchored native readout is a separate control.}
    \label{tab:supp-nsd-factorial}
    \begin{tabular*}{\textwidth}{@{\extracolsep{\fill}}lrrrr@{}}
        \toprule
        \multicolumn{5}{l}{\textit{(a) Complete native-space metrics}} \\
        Condition & Raw $\uparrow$ & Resid.\ $\uparrow$ & TC $\uparrow$ & MSE $\downarrow$ \\
        \midrule
        Neither (response-only) & .8089 & .1272 & .1284 & .0605 \\
        Spatial FT only & .8308 & .2995 & .3000 & .0532 \\
        RM only & .8195 & .1371 & .1434 & .0571 \\
        Spatial FT + RM & .8308 & .2995 & .3000 & .0532 \\
        \midrule
        No support (anatomical transfer) & .2377 & .0460 & .1284 & .1929 \\
        Shuffled spatial FT + RM & .8046 & .0005 & .0015 & .0629 \\
        Unanchored native readout & .8308 & .2995 & .3000 & .0532 \\
        \bottomrule
    \end{tabular*}

    \vspace{3pt}
    \begin{tabular*}{\textwidth}{@{\extracolsep{\fill}}lrrrrr@{}}
        \toprule
        \multicolumn{6}{l}{\textit{(b) Matched per-subject residual correlations}} \\
        Subject & Neither & FT only & RM only & FT + RM & Shuffled \\
        \midrule
        01 & .1280 & .2928 & .1325 & .2928 & .0371 \\
        02 & .1228 & .2865 & .1343 & .2865 & -.0419 \\
        05 & .1420 & .3610 & .1620 & .3610 & -.0071 \\
        07 & .1159 & .2577 & .1195 & .2577 & .0140 \\
        \bottomrule
    \end{tabular*}
\end{table*}

\begin{table*}[!t]
    \centering
    \tableformat
    \setlength{\tabcolsep}{3pt}
    \caption{\textbf{Same-source EEG factorial and memory controls.}
    Ten-subject macro averages; every condition uses the same $K=400$
    support identities and support mean. Shuffled columns refit the
    calibration with incorrect support pairs, including a fresh RM fit
    without FT. Coordinate RM is the adopted memory; bounded-gain variants
    are separate controls. Axis-permuted gains permute
    the memory-direction gains; Coordinate Ridge retains its zero-correction
    option. Raw/MSE are retained alongside the residual metric.}
    \label{tab:supp-factorial-complete}
    \begin{tabular*}{\textwidth}{@{\extracolsep{\fill}}l*{6}{r}@{}}
        \toprule
        & \multicolumn{3}{c}{Correct support pairs}
        & \multicolumn{3}{c}{Shuffled support pairs} \\
        \cmidrule(lr){2-4}\cmidrule(lr){5-7}
        Condition & Raw & Resid. & MSE & Raw & Resid. & MSE \\
        \midrule
        FT off / RM off & .6476 & .0865 & .0218 & .6476 & .0865 & .0218 \\
        FT on / RM off & .6496 & .0979 & .0217 & .6457 & .0597 & .0219 \\
        FT off / Coordinate RM & .6500 & .1032 & .0217 & .6407 & .0304 & .0221 \\
        FT on / Coordinate RM & .6508 & .1091 & .0217 & .6397 & .0261 & .0222 \\
        \midrule
        FT off / bounded RM & .6494 & .1001 & .0217 & .6414 & .0451 & .0221 \\
        FT on / bounded RM & .6505 & .1070 & .0217 & .6418 & .0410 & .0221 \\
        \midrule
        FT + in-sample memory gains & .6183 & .0876 & .0235 & .5992 & .0009 & .0244 \\
        FT + one global memory gain & .6502 & .1011 & .0217 & .6445 & .0529 & .0219 \\
        FT + axis-permuted memory gains & .6478 & .1002 & .0218 & .6413 & .0465 & .0221 \\
        FT + Global PRESS Ridge & .6505 & .1039 & .0217 & .6411 & .0325 & .0221 \\
        \bottomrule
    \end{tabular*}
\end{table*}

\section{Final Coordinate RM controls and diagnostics}
\label{sec:supp-coordinate-controls}

\paragraph{Matched calibration protocol.}
These controls use the final sources in \cref{tab:adaptation}, seed 42,
the first predefined $K400$ draw, and all 200 official test images per
subject. No source or Refiner is retrained. Removing FT requires a new
RM fit to that condition's errors; fully shuffled pairs refit FT and RM.
The four matched conditions are in \cref{tab:supp-final-factorial}, and
individual subjects are in \cref{tab:supp-coordinate-subjects}.
Source checkpoints, support indices, fitting budgets and the
MEG amplitude rate are fixed across conditions.

\begin{figure}[!tbp]
    \centering
    \includegraphics[width=\columnwidth]{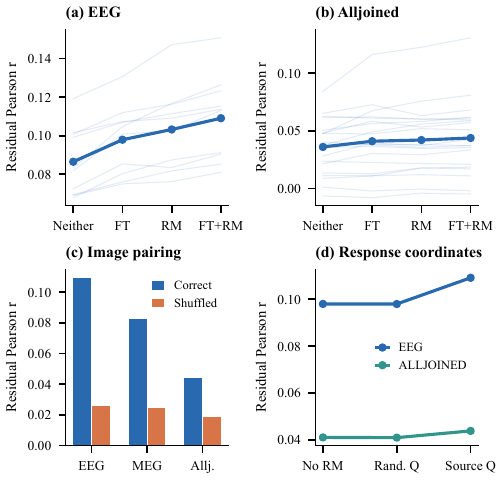}
    \caption{\textbf{Matched calibration and response-coordinate controls.}
    Same sources, $K400$ supports, and official tests.
    (a--b) Matched FT$\times$RM fits: subjects and means.
    (c) Correct versus fully shuffled pairs.
    (d) Correct FT fixed; each RM refits with random or source $Q$.
    MEG's four fits: \cref{tab:calibration-modules}.}
    \label{fig:coordinate-controls}
\end{figure}

\paragraph{Pairing and source coordinates.}
Correct calibration exceeds full-chain shuffled calibration in 10/10 EEG,
4/4 MEG, and 19/20 Alljoined subjects.
The mean residual differences are $.0830$, $.0577$, and $.0254$.
The random-$Q$ control retains the correctly fitted FT, replacing only the
fixed response span with a same-dimensional random orthogonal span.
Keys, supports, candidate penalties, and the null option are unchanged.
The matched subject-level comparisons and source-coordinate controls are
visualized in \cref{fig:coordinate-controls}.
Source coordinates exceed this control by $.0112$ on EEG (10/10 subjects)
and $.0029$ on Alljoined (11/20).
For the separate key-shuffle control, correctly fitted FT and source $Q$
remain fixed; only the pairing between support keys and remaining error
coordinates is shuffled before refitting RM.
MEG and Alljoined attain $.0789$ and $.0388$, respectively.
EEG averages $.0962$ over three fixed key permutations; its
complete coordinate-key corruption curve is in
\cref{tab:supp-memory-corruption}.

\begin{table*}[!t]
    \centering
    \tableformat
    \caption{\textbf{Final-coordinate memory controls: all four metrics.}
    $K400$, full cohorts and official tests.
    RM-only shuffled refits RM without FT using incorrect pairs.
    Random $Q$ and RM-key shuffle retain correctly fitted FT.
    EEG key shuffle averages three fixed permutations; MEG and Alljoined
    use their recorded single permutation.
    Centered denotes sample-wise Pearson after across-test-image centering;
    it is distinct from training-ERP residual Pearson.}
    \label{tab:supp-coordinate-controls}
    \begin{tabular*}{\textwidth}{@{\extracolsep{\fill}}llrrrr@{}}
        \toprule
        Dataset & Condition & Raw & Residual & Centered & MSE \\
        \midrule
        EEG2 & FT+RM & .6508 & .1091 & .1649 & .0217 \\
             & RM only & .6500 & .1032 & .1590 & .0217 \\
             & RM-only shuffled & .6407 & .0304 & .0716 & .0221 \\
             & Random $Q$ & .6496 & .0979 & .1554 & .0217 \\
             & RM-key shuffle & .6484 & .0962 & .1495 & .0218 \\
        \midrule
        MEG & FT+RM & .4976 & .0825 & .1033 & .6350 \\
            & RM only & .5038 & .0821 & .1074 & .6253 \\
            & RM-only shuffled & .4958 & .0370 & .0597 & .6322 \\
            & Random $Q$ & .4981 & .0806 & .1021 & .6344 \\
            & RM-key shuffle & .4949 & .0789 & .0994 & .6381 \\
            & Bounded-gain RM & .4977 & .0825 & .1034 & .6355 \\
            & Global PRESS RM & .4981 & .0806 & .1021 & .6344 \\
        \midrule
        Alljoined-1.6M & FT+RM & .3395 & .0438 & .0687 & .0879 \\
            & RM only & .3388 & .0420 & .0665 & .0880 \\
            & RM-only shuffled & .3335 & .0214 & .0369 & .0883 \\
            & Random $Q$ & .3423 & .0409 & .0725 & .0877 \\
            & RM-key shuffle & .3376 & .0388 & .0617 & .0880 \\
        \bottomrule
    \end{tabular*}
\end{table*}

\begin{table*}[!t]
    \centering
    \tableformat
    \caption{\textbf{Final-coordinate matched controls by subject.}
    Residual Pearson over every official test image.
    Neither retains the support mean; all supported conditions use the
    same 400 images. Full shuffle refits both FT and RM with incorrect
    pairs; random $Q$ retains correct FT and refits RM.}
    \label{tab:supp-coordinate-subjects}
    \begin{tabular*}{\textwidth}{@{\extracolsep{\fill}}lrrrrrr@{}}
        \toprule
        Subject & Neither & FT only & RM only & FT+RM & Full shuffle & Random $Q$ \\
        \midrule
        \multicolumn{7}{l}{\textit{THINGS-EEG2}} \\
        01 & .0695 & .0760 & .0817 & .0853 & .0050 & .0760 \\
        02 & .0690 & .0750 & .0762 & .0810 & .0130 & .0750 \\
        03 & .1008 & .1120 & .1163 & .1233 & .0319 & .1120 \\
        04 & .0725 & .0854 & .0835 & .0905 & .0180 & .0854 \\
        05 & .0679 & .0804 & .0874 & .0912 & .0251 & .0803 \\
        06 & .0992 & .1070 & .1115 & .1154 & .0479 & .1068 \\
        07 & .1192 & .1307 & .1473 & .1508 & .0525 & .1306 \\
        08 & .0833 & .1047 & .1168 & .1264 & .0074 & .1048 \\
        09 & .1016 & .1073 & .1092 & .1137 & .0474 & .1072 \\
        10 & .0818 & .1009 & .1025 & .1131 & .0127 & .1008 \\
        \midrule
        \multicolumn{7}{l}{\textit{THINGS-MEG}} \\
        01 & .0794 & .0707 & .0794 & .0691 & .0359 & .0707 \\
        02 & .1072 & .1608 & .1695 & .1693 & .0342 & .1608 \\
        03 & -.0021 & .0184 & .0111 & .0189 & -.0226 & .0184 \\
        04 & .0524 & .0726 & .0685 & .0726 & .0515 & .0726 \\
        \midrule
        \multicolumn{7}{l}{\textit{Alljoined-1.6M}} \\
        01 & .0476 & .0474 & .0519 & .0509 & .0256 & .0472 \\
        02 & .0841 & .1160 & .1223 & .1305 & .0420 & .1159 \\
        03 & .0480 & .0583 & .0532 & .0572 & .0190 & .0582 \\
        04 & .0613 & .0621 & .0601 & .0610 & .0469 & .0618 \\
        05 & .0283 & .0395 & .0373 & .0425 & .0133 & .0393 \\
        06 & .0376 & .0490 & .0558 & .0588 & .0179 & .0489 \\
        07 & .0135 & .0128 & .0180 & .0171 & .0013 & .0128 \\
        08 & .0404 & .0383 & .0382 & .0371 & .0181 & .0382 \\
        09 & -.0063 & -.0078 & -.0039 & -.0047 & -.0046 & -.0077 \\
        10 & .0626 & .0609 & .0602 & .0594 & .0431 & .0608 \\
        11 & .0013 & -.0020 & -.0004 & -.0019 & -.0213 & -.0021 \\
        12 & .0215 & .0305 & .0293 & .0335 & .0058 & .0305 \\
        13 & .0386 & .0379 & .0357 & .0373 & .0214 & .0377 \\
        14 & .0475 & .0674 & .0757 & .0808 & .0258 & .0674 \\
        15 & .0355 & .0366 & .0347 & .0355 & .0254 & .0364 \\
        16 & .0114 & .0111 & .0121 & .0111 & -.0013 & .0111 \\
        17 & .0089 & .0109 & .0177 & .0186 & .0052 & .0109 \\
        18 & .0505 & .0563 & .0580 & .0618 & .0329 & .0564 \\
        19 & .0237 & .0225 & .0217 & .0211 & .0142 & .0224 \\
        20 & .0652 & .0725 & .0632 & .0681 & .0374 & .0722 \\
        \bottomrule
    \end{tabular*}
\end{table*}

\subsection{Frozen final-model diagnostics}
\label{sec:supp-coordinate-diagnostics}

\begin{figure}[!t]
    \centering
    \includegraphics[width=\columnwidth]{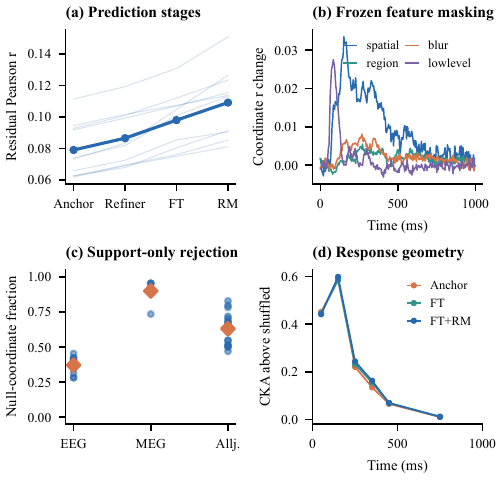}
    \caption{\textbf{Frozen final-model diagnostics.}
    (a) Prediction stages; (b) across-image correlation decrease under
    frozen feature masking.
    (c) Support-selected null fractions: subjects and cohort means.
    (d) Windowed CKA above 32 image-permutation references.
    (a,b,d) All ten EEG subjects; (c) complete sensor cohorts.
    All 200 official test images.}
    \label{fig:supp-complete-diagnostics}
\end{figure}

\Cref{fig:supp-complete-diagnostics} uses final Coordinate RM.
All ten EEG subjects and all official test images contribute to stages,
feature masking, and representation geometry. Stage readouts come from
one fitted pipeline and do not replace independently refitted module
comparisons. Interventions zero one normalized
feature group while freezing the weights, FT, support-key center and RM.
Spatial, region, blur, and low-level removal yield residual $.0712$,
$.1016$, $.0990$, and $.0940$, respectively, against the intact $.1091$.
\Cref{tab:supp-coordinate-masking} retains the other metrics, including
the region-removal control's slightly better raw correlation and MSE.

\begin{table}[!t]
    \centering
    \tableformat
    \setlength{\tabcolsep}{2pt}
    \caption{\textbf{Final-model frozen feature masking.}
    Ten EEG subjects and all official tests.
    Only the specified normalized input group is zeroed; fitting states
    remain frozen. Centered is test-centered sample-wise Pearson.}
    \label{tab:supp-coordinate-masking}
    \begin{tabular*}{\columnwidth}{@{\extracolsep{\fill}}lrrrr@{}}
        \toprule
        Input & Raw & Resid. & Centered & MSE \\
        \midrule
        Intact & .6508 & .1091 & .1649 & .0217 \\
        Spatial zeroed & .6486 & .0712 & .1473 & .0217 \\
        Region zeroed & .6511 & .1016 & .1628 & .0216 \\
        Blur zeroed & .6506 & .0990 & .1603 & .0217 \\
        Low-level zeroed & .6495 & .0940 & .1539 & .0217 \\
        \bottomrule
    \end{tabular*}
\end{table}

\paragraph{Support-only rejection.}
For each subject, the null fraction counts source coordinates whose
support PRESS criterion selects the infinite-penalty candidate out of
64 directions. MEG selects null in $89.84\%$ of coordinates after FT.
The EEG mean null fraction is $37.03\%$.
MEG's retained directions raise residual correlation from $.0806$ to $.0825$;
RM-only reaches $.0821$.
Across EEG subjects, correction removes a mean $0.25\%$ of the remaining
waveform SSE while raising residual correlation from $.0979$ to $.1091$.

\paragraph{Disjoint-repeat consistency.}
The final model is also scored against fixed odd/even halves of the 80
test repeats (forty responses per image per half).
\Cref{tab:supp-coordinate-repeats} reports both halves for the same
frozen predictions. FT+RM remains positive on either half and improves
on the corresponding FT-only prediction.
The measured halves correlate at residual $.1295$; this descriptive
reference does not define a noise ceiling.
Within-subject predictions reach $.1516$ and $.1532$ against the two
halves. They use the full target training set, whereas LOSO calibration
uses $K400$; this reference is not a matched-budget model comparison.

\begin{table}[!t]
    \centering
    \tableformat
    \setlength{\tabcolsep}{2pt}
    \caption{\textbf{EEG consistency across disjoint repeat averages.}
    All ten subjects and 200 test images.
    Each test half averages forty responses.
    (a) Measured-repeat and full-target-training within-subject references,
    plus final LOSO $K400$ FT+Coordinate RM.
    (b) Support-matched calibration conditions; each prediction is scored
    against both halves.}
    \label{tab:supp-coordinate-repeats}
    \begin{tabular*}{\columnwidth}{@{\extracolsep{\fill}}lrrr@{}}
        \toprule
        \multicolumn{4}{l}{\textit{(a) Complete reference metrics}} \\
        Comparison & Raw & Resid. & MSE \\
        \midrule
        Odd vs.\ even measured & .3973 & .1295 & .0658 \\
        Within-subject vs.\ odd & .5689 & .1516 & .0370 \\
        Within-subject vs.\ even & .5704 & .1532 & .0365 \\
        LOSO $K400$ vs.\ odd & .5474 & .0844 & .0383 \\
        LOSO $K400$ vs.\ even & .5477 & .0812 & .0379 \\
        \bottomrule
    \end{tabular*}

    \vspace{3pt}
    \begin{tabular*}{\columnwidth}{@{\extracolsep{\fill}}lrr@{}}
        \toprule
        \multicolumn{3}{l}{\textit{(b) Matched calibration: residual Pearson}} \\
        Prediction & Odd repeats & Even repeats \\
        \midrule
        Response-only & .0672 & .0640 \\
        FT only & .0759 & .0727 \\
        FT+RM & .0844 & .0812 \\
        \bottomrule
    \end{tabular*}
\end{table}

\paragraph{Representation geometry.}
Windowed centered linear CKA uses all 200 images, with
the mean of 32 fixed image-permutation references subtracted.
The displayed curves average subjects, without a test-fitted projection.
Across-image temporal correlations in the masking plot and the
sample-wise flattened residual metric quantify different quantities.
These descriptive diagnostics do not establish neural component causality.

\section{Final-model prediction diagnostics}
\label{sec:supp-frozen-diagnostics}

The stage and visual-input analyses use the final frozen
Anchor--Refiner, FT, and Coordinate RM model from \cref{tab:adaptation},
with all 200 official test images per subject.
They introduce no fitted parameters or selected checkpoints.
Stage comparisons use intermediate outputs from the same fitted model,
not separately retrained models.

\begin{figure}[!t]
    \centering
    \includegraphics[trim=124bp 0 0 0,clip,width=.65\columnwidth]{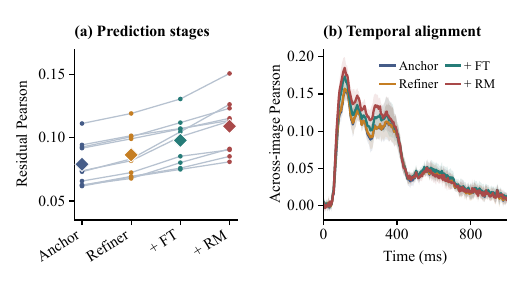}
    \caption{\textbf{Temporal alignment across prediction stages.}
    Ten strict-LOSO EEG folds, the same 400 training supports, and all
    200 official test images. (b) Across-image correlation,
    averaged over electrodes and subjects; bands show 95\% intervals
    from 2,000 subject-bootstrap resamples. Intermediate-stage readouts
    are not independently retrained models. Scalar stage scores appear
    in \cref{fig:supp-complete-diagnostics}a.}
    \label{fig:supp-calibration-stages}
\end{figure}

\paragraph{Where calibration changes prediction.}
The scalar stage summary in \cref{fig:supp-complete-diagnostics}a follows
the source anchor, response refiner, FT, and RM with the same support-derived prior.
Residual correlation increases from $.0790$ at the anchor to $.0865$
after refinement, $.0979$ after FT, and $.1091$ after RM.
FT+RM improves over the frozen refiner on $83.7\%$ of the 2,000
subject--image observations; this pooled fraction is descriptive.
The temporal profiles in \cref{fig:supp-calibration-stages} resample subjects,
retaining all images within each resampled subject.

\begin{figure}[!t]
    \centering
    \includegraphics[trim=124bp 0 0 0,clip,width=.65\columnwidth]{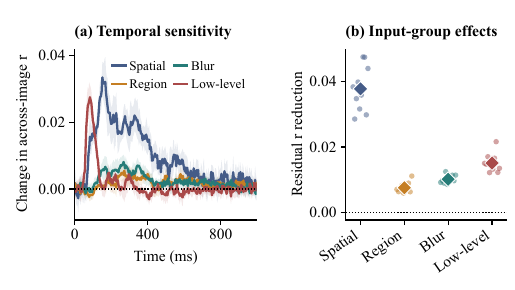}
    \caption{\textbf{Subject-wise visual-input sensitivity.}
    (b) Change in residual Pearson; dots are subjects and diamonds are
    means. All ten subjects and all official test images contribute.
    Weights, branch scaling, calibration states and support centers are
    frozen; these are input interventions rather than retrained ablations.
    Temporal profiles appear in \cref{fig:supp-complete-diagnostics}b.}
    \label{fig:supp-feature-sensitivity}
\end{figure}

\paragraph{Sensitivity to visual structure.}
Zeroing one normalized input-cache group gives the temporal profiles in
\cref{fig:supp-complete-diagnostics}b and subject-wise effects in
\cref{fig:supp-feature-sensitivity}.
Mean residual-correlation reductions are $.0378$ for spatial features,
$.0075$ for region descriptors, $.0101$ for blur features, and $.0151$
for low-level descriptors. Their largest temporal changes occur at
152, 272, 268, and 84\,ms, respectively.
These descriptive peaks characterize reliance of the fitted predictor:
all weights, branch scaling, support centers, FT, and RM remain fixed.
Retrained feature-removal controls are reported separately in
\cref{sec:supp-module-ablations}.

\begin{figure}[!t]
    \centering
    \includegraphics[width=\columnwidth]{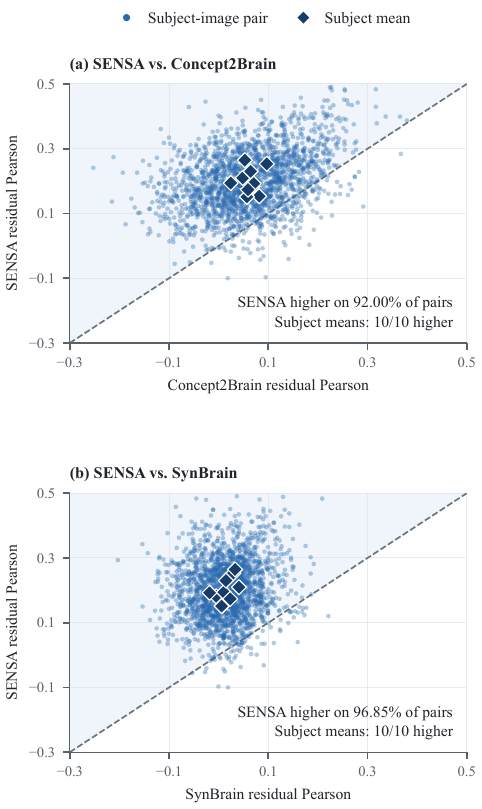}
    \caption{\textbf{Paired image-specific encoding accuracy.}
    Each dot compares residual Pearson for the same subject and test image:
    SENSA on the vertical axis and a baseline on the horizontal axis.
    All ten subjects and 200 official test images contribute to each panel.
    Points above the dashed equality line favor SENSA; diamonds denote the
    ten subject means. Axes use identical metric limits in both panels.
    SENSA uses one fixed seed-42 model, distinct from the main EEG ensemble;
    Concept2Brain uses the final full-training refit.}
    \label{fig:residual-examples}
\end{figure}

\section{Paired image-specific encoding accuracy}

\Cref{fig:residual-examples} compares every subject--image pair across
the complete test set. Each correlation flattens all
$63\times250$ response samples over 0--996\,ms after subtracting the same
subject-specific training ERP from prediction and measurement.
SENSA's mean is $.2004$, compared with $.0623$ for the final
Concept2Brain reference and
$.0143$ for SynBrain. SENSA is higher on $92.00\%$ and $96.85\%$ of
the 2,000 paired observations, respectively, and has the higher subject
mean in all ten subjects for both comparisons. These pooled percentages
describe coverage; they do not treat images as independent subjects.

Correlation is invariant to positive global rescaling, so amplitude is
scored separately by MSE, and no rescaling is applied in this figure.
For the lower-median EEG subject (06), the fixed 25th-,
50th-, and 75th-percentile examples have SENSA-to-measured residual RMS
ratios of $.170$, $.222$, and $.225$ over the full response window:
predicted image-specific deflections are smaller than measured ones.

\begin{figure}[!t]
    \centering
    \includegraphics[width=\columnwidth]{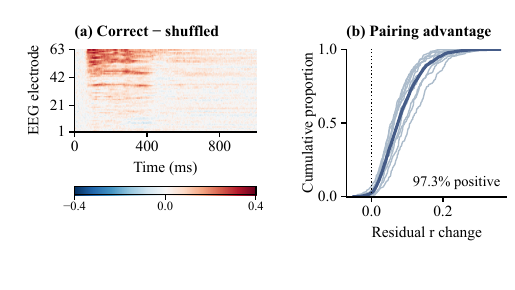}
    \caption{\textbf{Final-model pairing analysis.}
    (a) Subject-averaged correct-minus-shuffled across-image correlation at
    each EEG electrode and time point.
    (b) Changes in sample-wise residual correlation over all 200 official EEG
    test images per subject. Pale curves show individual subjects; the bold
    curve pools all 2,000 subject--image observations descriptively.
    Both conditions use the final Anchor--Refiner, FT, and Coordinate RM
    model with $K=400$; the shuffled condition refits the full calibration
    chain with shuffled support pairs.}
    \label{fig:pairing-detail}
\end{figure}

\section{Detailed pairing analysis}

The final-source information comparison in \cref{tab:supp-final-information}
covers all four datasets. \Cref{fig:pairing-detail} resolves EEG changes
over electrodes, time, and individual images.
Correct pairing improves residual correlation for
$97.30\%$ of EEG subject--image observations. The pooled distribution is
descriptive; bootstrap intervals elsewhere resample subjects.

\begin{figure}[!htbp]
    \centering
    \includegraphics[trim=0 0 0 209bp,clip,width=\columnwidth]{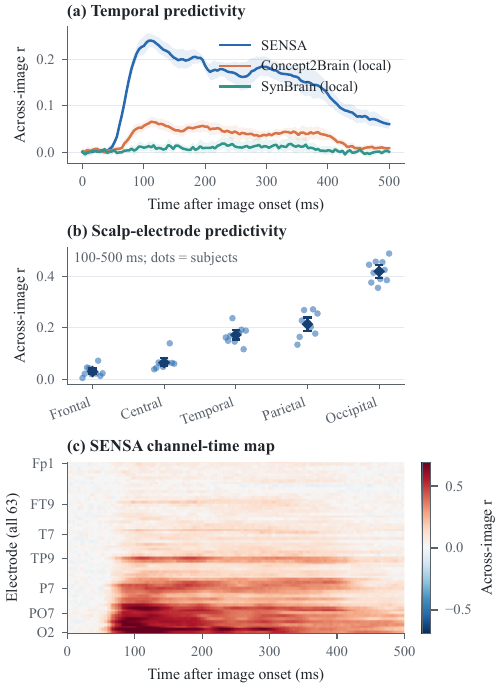}
    \caption{\textbf{Image-specific prediction over all EEG electrodes.}
    Full electrode--time map accompanying \cref{fig:spatiotemporal}.
    Correlation is computed across all 200 official test images at each
    electrode and time, then averaged over ten subjects.
    The color scale retains the original common analysis scale.}
    \label{fig:all-electrode-map}
\end{figure}

\section{Computational efficiency}
\label{sec:supp-efficiency}

\begin{figure}[!t]
    \centering
    \includegraphics[width=\columnwidth]{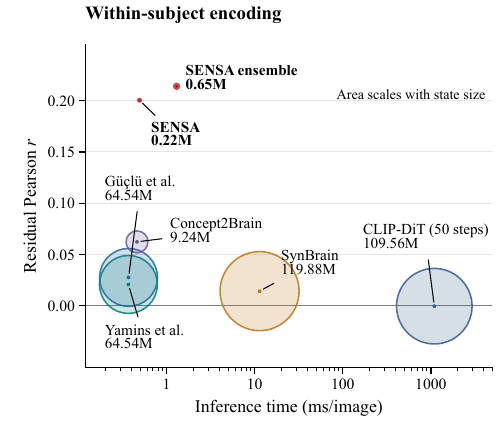}
    \caption{\textbf{Model size, inference time, and reconstruction accuracy.}
    Bubble area scales with response-state coefficients, including prior
    and normalization statistics and excluding frozen visual backbones
    and training-only encoders. Latency is measured on one A800 with
    subject-01 weights, batch size one, cached visual features, and each
    method's native inference precision. Residual correlation averages
    ten subjects on the official 200-image test set.
    SENSA shows single-model and
    three-model ensemble results. Concept2Brain uses its final
    full-training checkpoint.}
    \label{fig:supp-efficiency}
\end{figure}

\Cref{fig:supp-efficiency} compares the deployed response predictors on
THINGS-EEG2. The single-model SENSA checkpoint uses $0.22$M response-state
coefficients and takes $0.49$\,ms per image, with residual correlation
$.2004$. The three-model ensemble uses $0.65$M coefficients and
$1.30$\,ms per image, reaching $.2140$.

\section{Encoding-based decoding of real brain responses}
\label{sec:supp-real-query-retrieval}

\paragraph{Purpose and protocol.}
This experiment tests whether predicted responses can identify the image
that elicited a measured response. Each of the 200 official test images
supplies a generated response template; each measured test response is
ranked against the complete gallery. Queries are real brain recordings.
Unlike image--synthetic-response--image cycle consistency, the query image
identity is withheld during ranking.

We reuse the final LOSO source checkpoints, the first predefined $K400$
support draw, and the final predictions for all ten EEG, four MEG,
and twenty Alljoined subjects. No generator, calibration state, or decoder
is trained for this evaluation. Source training uses the complete official
training partitions. EEG queries average all 80 official test repeats,
and MEG/Alljoined retain their existing official test-average convention.
These are averaged-response retrieval results, not single-trial decoding.

\paragraph{Fixed scoring and controls.}
The primary score is Pearson across all channel--time coordinates after
subtracting the same target-support mean from queries and templates.
The mean uses only the 400 target-training images. Within-vector mean
removal is part of Pearson; no statistic is fitted to test queries or to
the test cohort. Raw Pearson is a fixed secondary score.
The target-only encoding Ridge uses the same four descriptors and K400
support indices, with no source brain supervision. Descriptor scaling and
the Ridge penalty are fitted within the support set; the latter uses
conditional analytic PRESS. This control covers all three sensor cohorts.
The RM-only condition refits memory without FT. Full-chain shuffled
calibration and random-$Q$ use the definitions in
\cref{tab:supp-coordinate-controls}.
Tied candidates are evaluated by uniformly randomized expected ranks,
so constant templates attain exactly $0.50\%$ Top-1 and $2.50\%$ Top-5.

\paragraph{Template-based decoding results.}
\Cref{tab:supp-real-query-retrieval} reports subject-macro averages.
On EEG, the full model exceeds target-only encoding Ridge in all ten
subjects. Its Top-1 gain over FT-only is $5.75$ percentage points
(9/10 subjects; 95\% subject-bootstrap interval $[3.10,8.15]$).
The interval resamples subjects 10,000 times with fixed seed 20261106;
it is descriptive rather than a multiplicity-adjusted significance test.
Alljoined gains $0.50$ points over FT-only, with an interval spanning zero
($[-0.20,1.20]$); MEG changes from $2.38\%$ to $2.50\%$ Top-1.
MEG's raw-score retrieval instead favors RM-only, so this analysis
supports a clear decoding benefit for the final EEG pipeline rather than
a uniform cross-modal gain. It establishes template-based decoding utility;
training a decoder with synthetic responses is a separate experiment.
Scoring is fixed before retrieval evaluation.

\begin{table*}[!t]
    \centering
    \tableformat
    \setlength{\tabcolsep}{4pt}
    \caption{\textbf{Real-query image retrieval from generated response templates.}
    Complete official 200-way galleries, all subjects, final LOSO sources
    and one fixed $K400$ draw. Top-1/Top-5 are percentages; MRR is unitless.
    Panel A uses support-centered Pearson; Panel B uses raw Pearson.
    The unadapted source in Panel B uses no target support.
    No new decoder is fitted.}
    \label{tab:supp-real-query-retrieval}
    \begin{tabular*}{\textwidth}{@{\extracolsep{\fill}}l*{9}{r}@{}}
        \toprule
        & \multicolumn{3}{c}{THINGS-EEG2 (10)}
        & \multicolumn{3}{c}{THINGS-MEG (4)}
        & \multicolumn{3}{c}{Alljoined-1.6M (20)} \\
        \cmidrule(lr){2-4}\cmidrule(lr){5-7}\cmidrule(lr){8-10}
        Condition & Top-1 & Top-5 & MRR
                  & Top-1 & Top-5 & MRR
                  & Top-1 & Top-5 & MRR \\
        \midrule
        \multicolumn{10}{l}{\textbf{A. Support-centered Pearson (primary)}} \\
        Support mean only & 0.50 & 2.50 & .0294
                          & 0.50 & 2.50 & .0294 & 0.50 & 2.50 & .0294 \\
        Target-only encoding Ridge & 2.95 & 10.75 & .0842
                                   & 2.00 & 7.38 & .0618 & 0.90 & 4.23 & .0413 \\
        Source + support mean & 16.60 & 42.05 & .2935
                              & 2.00 & 8.75 & .0651 & 2.65 & 10.68 & .0852 \\
        FT only & 19.70 & 47.20 & .3321
                & 2.38 & 8.00 & .0685 & 2.92 & 11.97 & .0920 \\
        RM only & 24.40 & 53.55 & .3825
                & 2.25 & 8.62 & .0704 & 3.28 & 12.28 & .0939 \\
        FT + RM & 25.45 & 54.40 & .3928
                & 2.50 & 7.63 & .0696 & 3.43 & 12.72 & .0960 \\
        Full-chain shuffled & 4.50 & 16.20 & .1202
                            & 0.88 & 5.50 & .0449 & 1.38 & 5.60 & .0520 \\
        FT + random-$Q$ RM & 19.85 & 47.10 & .3332
                           & 2.38 & 8.00 & .0687 & 2.95 & 11.80 & .0922 \\
        \midrule
        \multicolumn{10}{l}{\textbf{B. Raw Pearson (secondary)}} \\
        Support mean only & 0.50 & 2.50 & .0294
                          & 0.50 & 2.50 & .0294 & 0.50 & 2.50 & .0294 \\
        Unadapted source & 12.80 & 35.10 & .2445
                         & 1.88 & 7.62 & .0639 & 2.83 & 11.40 & .0876 \\
        Target-only encoding Ridge & 2.95 & 10.65 & .0859
                                   & 1.75 & 7.25 & .0627 & 1.10 & 4.38 & .0431 \\
        Source + support mean & 16.90 & 44.55 & .3017
                              & 2.50 & 7.50 & .0676 & 2.75 & 11.53 & .0894 \\
        FT only & 20.25 & 49.65 & .3434
                & 2.25 & 8.25 & .0716 & 3.43 & 12.90 & .0985 \\
        RM only & 26.15 & 56.55 & .4040
                & 2.88 & 9.75 & .0795 & 3.43 & 12.47 & .0952 \\
        FT + RM & 26.10 & 57.65 & .4085
                & 2.00 & 8.38 & .0705 & 3.58 & 12.87 & .0979 \\
        Full-chain shuffled & 4.65 & 15.80 & .1197
                            & 0.88 & 5.13 & .0443 & 1.60 & 5.50 & .0529 \\
        FT + random-$Q$ RM & 20.35 & 49.75 & .3436
                           & 2.25 & 8.25 & .0716 & 3.40 & 12.97 & .0982 \\
        \bottomrule
    \end{tabular*}
\end{table*}

\subsection{Training decoders with generated responses}
\label{sec:supp-synthetic-augmentation}

\paragraph{Two data-scarce applications.}
We train a separate UBP decoder using 400 measured image--EEG pairs from
each new subject and additional synthetic training pairs.
The frozen SENSA source excludes that subject; its FT and Coordinate RM
use the same 400 paired target-training images.
The first application completes a planned stimulus set: SENSA predicts
responses for the remaining 16,140 official training images.
The second expands training content: SENSA predicts responses for 2,964
additional THINGS images absent from the EEG training image set.
The additional-image pool also excludes the archived EEG/MEG/fMRI test
image identities and test concepts; no MEG or fMRI responses are used.
These protocols respectively replace uncollected responses to planned
stimuli and add stimuli without further target EEG acquisition.

All ten subjects use the same predefined $K400$ support draw as the
calibration experiments. Each measured training response averages its four
repetitions. Testing uses the actual EEG averaged over all 80 repetitions
for each of the 200 official test images, ranked against the complete
200-image gallery. Synthetic signals enter decoder training only.

\paragraph{Decoder and matched controls.}
We reuse UBP's released EEG projection architecture, 17 named posterior
channels and 250 time samples, frozen RN50 image features, and its
three-level blur curriculum. A matched augmentation wrapper trains
real-only, source-generated, SENSA-generated and shuffled-pair conditions
from the same initialization and real-minibatch order.
The source-generated condition omits target FT and RM while retaining
the same support mean. Shuffling permutes synthetic image labels while
retaining the generated responses.
All $K400$ conditions use 1,000 updates, AdamW with learning rate
$10^{-4}$ and weight decay $10^{-4}$, real batches of 32, and synthetic
batches of 32 where applicable. The real and synthetic symmetric
contrastive losses are averaged separately and added with equal weight.
Each condition retains its own blur-curriculum state.
All checkpoints are fixed to the final update; no official test EEG is
used for fitting or selection.
All training, additional-image and gallery features share one frozen
RN50 preprocessing implementation.
This is a controlled UBP augmentation experiment rather than a
reproduction of its complete original training recipe.

\paragraph{Decoder initialization and aggregation.}
The real-only, source-generated and SENSA-generated conditions each use
three matched decoder initializations, $440000+s$, $540000+s$ and
$640000+s$ for subject $s$. Source generation and its calibration remain
fixed at source seed 42. Initialization and dropout draws vary together;
real and synthetic minibatch orders and all budgets remain fixed.
For uncertainty, we average initializations within subject before
10,000 subject-bootstrap resamples (seed 20261110).
These runs measure decoder initialization stability.
Shuffled-pair and target-only encoding controls, and the qualitative
examples, use decoder seed $440000+s$.

\paragraph{Synthetic training improves real-EEG retrieval.}
\Cref{tab:supp-ubp-augmentation} reports subject-macro averages.
Across the three decoder initializations, planned-image completion raises
Top-1 from $5.27\%$ to $10.50\%$ and Top-5 from $18.85\%$ to $30.90\%$.
The additional-image pool reaches $10.75\%$ Top-1 and $30.53\%$ Top-5.
Both Top-1 gains hold for all ten subjects after within-subject
initialization averaging. Their descriptive 95\% intervals are
$[3.85,6.77]$ and $[4.47,6.48]$ percentage points.
Relative to source-generated augmentation, individualized SENSA adds
$2.25$ points ($[1.30,3.17]$; 8 improved, 1 tied, 1 lower) and
$1.65$ points ($[.77,2.53]$; 9/10 improved), respectively.
Thus synthetic pairs expand scarce decoder training data, with an
additional EEG benefit from target-specific calibration.

\begin{figure*}[!t]
    \centering
    \includegraphics[width=\textwidth]{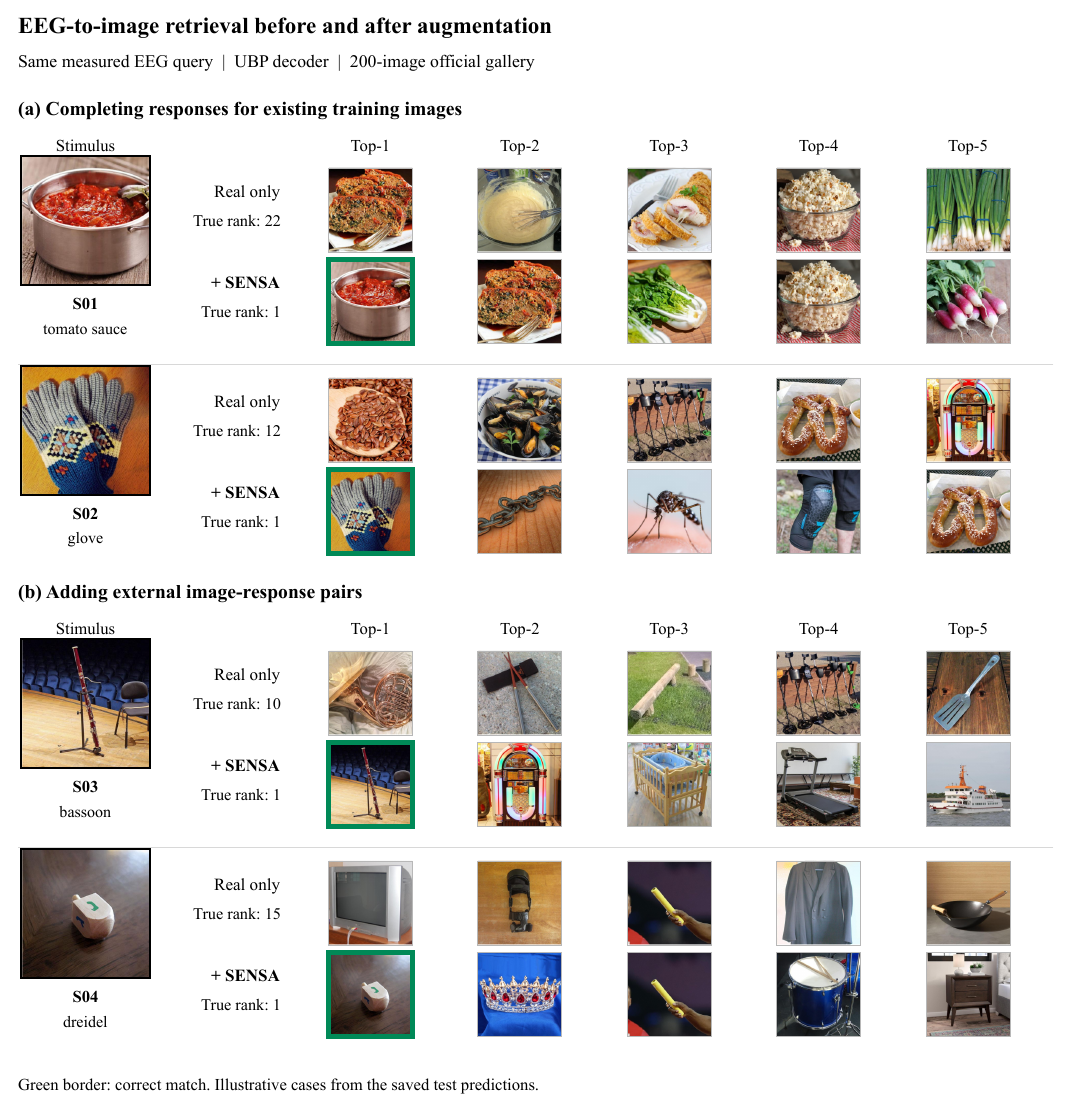}
    \caption{\textbf{Image retrieval before and after synthetic augmentation.}
    Selected THINGS-EEG2 examples from UBP trained on 400 measured pairs
    alone (Real only) or with additional SENSA-generated pairs (+ SENSA).
    Both decoders receive the same measured EEG query and rank all 200
    official test images. The left image shows the stimulus for reference;
    green borders mark correct matches. Panel (a) completes responses for
    existing training images; panel (b) adds external image--response pairs.
    True rank gives the correct image's position in the complete gallery.
    Examples use decoder seed $440000+s$.}
    \label{fig:supp-ubp-retrieval-examples}
\end{figure*}

\paragraph{Single-initialization controls.}
With decoder seed $440000+s$, real-only training reaches
$5.25\%$ Top-1. Source and SENSA augmentation reach $8.75\%$ and $10.40\%$
for planned images, and $9.15\%$ and $10.30\%$ for additional images.
Shuffled pairs instead reach $3.75\%$ and $1.80\%$ Top-1, with Top-5 of
$11.65\%$ and $6.90\%$, supporting image--response correspondence.
For planned-image completion, target-only encoding Ridge augmentation
yields $7.20\%$ Top-1 and $23.85\%$ Top-5.
These controls share initialization and training budgets.
\Cref{fig:supp-ubp-retrieval-examples} shows selected retrieval examples:
for the same measured EEG query, synthetic augmentation moves the
correct stimulus from outside the top five to the first rank.

\paragraph{Extension to Alljoined.}
Panel B of \cref{tab:supp-ubp-augmentation} evaluates all twenty Alljoined
subjects using a linear brain-to-visual decoder and the same predefined
$K400$ support identities. The frozen LOSO generator supplies responses
for the remaining official training images; visual targets are the frozen
multidescriptor features. All decoder fits precede evaluation
on real test responses against the complete 200-image gallery.
SENSA augmentation raises Top-1 from $1.50\%$ to $3.30\%$ and Top-5 from
$7.65\%$ to $12.40\%$. Source-generated and shuffled-label controls retain
the same response budget. The Top-1 gain over real-only training is
$1.80$ points (95\% interval $[1.35,2.30]$; 19 improved, 1 tied).
Relative to source-generated augmentation, the gain is $.125$ points
($[-.375,.675]$). Thus this second EEG benchmark supports augmentation
utility; its increment from individualization is not resolved by the
paired interval.

\begin{table}[!t]
    \centering
    \tableformat
    \setlength{\tabcolsep}{3pt}
    \caption{\textbf{Synthetic responses augment scarce decoder training.}
    Real-response retrieval with $K400$ measured training pairs and the
    complete official 200-way test. Panel A: UBP on ten EEG2 subjects;
    planned images complete the remaining 16,140 training responses,
    while additional images supply 2,964 new responses.
    Panel A averages three matched decoder initializations.
    Panel B: a linear decoder on twenty Alljoined subjects, completing
    the remaining training responses with one decoder fit per subject.
    Full-real uses all measured
    training responses. Values are subject-macro percentages.}
    \label{tab:supp-ubp-augmentation}
    \begin{tabular*}{\columnwidth}{@{\extracolsep{\fill}}lrrrr@{}}
        \toprule
        \multicolumn{5}{l}{\textbf{A. THINGS-EEG2: UBP}} \\
        & \multicolumn{2}{c}{Planned images}
        & \multicolumn{2}{c}{Additional images} \\
        \cmidrule(lr){2-3}\cmidrule(lr){4-5}
        Training data & Top-1 & Top-5 & Top-1 & Top-5 \\
        \midrule
        Real only & 5.27 & 18.85 & 5.27 & 18.85 \\
        + Source synthetic & 8.25 & 25.77 & 9.10 & 25.90 \\
        + SENSA synthetic & \textbf{10.50} & \textbf{30.90}
                          & \textbf{10.75} & \textbf{30.53} \\
        \midrule
        \multicolumn{5}{l}{\textbf{B. Alljoined-1.6M: linear decoder}} \\
        Training data & \multicolumn{2}{c}{Top-1} & \multicolumn{2}{c}{Top-5} \\
        \midrule
        $K400$ real & \multicolumn{2}{c}{1.500} & \multicolumn{2}{c}{7.650} \\
        + Source synthetic & \multicolumn{2}{c}{3.175}
                           & \multicolumn{2}{c}{12.075} \\
        + SENSA synthetic & \multicolumn{2}{c}{\textbf{3.300}}
                          & \multicolumn{2}{c}{\textbf{12.400}} \\
        + Shuffled SENSA pairs & \multicolumn{2}{c}{1.225}
                              & \multicolumn{2}{c}{5.750} \\
        \midrule
        Full-real reference & \multicolumn{2}{c}{7.775}
                            & \multicolumn{2}{c}{24.675} \\
        \bottomrule
    \end{tabular*}
\end{table}

\begin{table*}[!t]
    \centering
    \tableformat
    \caption{\textbf{Decoder initialization stability on THINGS-EEG2.}
    Subject-macro percentages over all ten subjects and the complete
    200-way test. Initialization denotes the base offset; actual decoder
    seeds add subject index $s$. All conditions share initialization,
    minibatch order and training budgets within each run.
    The source generator and calibration are fixed at source seed 42.}
    \label{tab:supp-ubp-initializations}
    \begin{tabular*}{\textwidth}{@{\extracolsep{\fill}}llrrrrrr@{}}
        \toprule
        & & \multicolumn{2}{c}{Real only}
          & \multicolumn{2}{c}{+ Source synthetic}
          & \multicolumn{2}{c}{+ SENSA synthetic} \\
        \cmidrule(lr){3-4}\cmidrule(lr){5-6}\cmidrule(lr){7-8}
        Image pool & Initialization & Top-1 & Top-5 & Top-1 & Top-5 & Top-1 & Top-5 \\
        \midrule
        Planned & 440000 & 5.25 & 18.30 & 8.75 & 25.65 & 10.40 & 30.45 \\
                & 540000 & 5.90 & 19.60 & 8.20 & 25.30 & 10.55 & 31.20 \\
                & 640000 & 4.65 & 18.65 & 7.80 & 26.35 & 10.55 & 31.05 \\
        \midrule
        Additional & 440000 & 5.25 & 18.30 & 9.15 & 25.60 & 10.30 & 30.45 \\
                   & 540000 & 5.90 & 19.60 & 9.25 & 25.85 & 11.00 & 30.65 \\
                   & 640000 & 4.65 & 18.65 & 8.90 & 26.25 & 10.95 & 30.50 \\
        \bottomrule
    \end{tabular*}
\end{table*}

The augmentation gains above concern data-scarce $K400$ training; in a
separate reference with all 16,540 measured pairs, the tested equal-weight
addition of 2,964 synthetic pairs did not improve retrieval.

\section{Stimulus-aware response completion}
\label{sec:supp-response-completion}

A predicted response can also supply an image-conditioned prior for completing
partially observed recordings. We evaluate this use on all ten THINGS-EEG2
subjects with the same frozen LOSO sources and $K400$ calibration pairs.
At inference, the stimulus image and the visible response coordinates are
available. Two fixed masks hide either 16 randomly selected coordinates or
the eight posterior coordinates
PO7/PO3/POz/PO4/PO8/O1/Oz/O2, including all 250 time samples.
The complete calibration recordings fit each readout; hidden test values enter
only the final scoring.

The image prior predicts the complete response. A Ridge readout then estimates
the remaining error in the hidden coordinates from the error observed in
the visible coordinates. Its penalty is selected by PRESS within the same
400 calibration pairs. Native-resolution controls retain all 250 visible time
samples; a temporally pooled variant uses 25 ten-sample bins.
All methods use the same masks and official 200-image test partition.

\Cref{tab:supp-response-completion} reports errors on the hidden coordinates.
Adding the final image prior to native-resolution Ridge raises residual
correlation from $.1274$ to $.2296$ for the random mask and from $.0552$ to
$.3348$ for the posterior mask. Both improvements hold for all ten subjects.
Compared with the unadapted source prior, the final prior also improves
native-resolution completion in all ten subjects for both masks.

This experiment operates in the stored MVNN response representation, which
was spatially whitened before masking. It therefore measures completion of
preprocessed response coordinates, with known stimulus identity, rather than
physical electrode-failure recovery.

\begin{table}[!t]
    \centering
    \tableformat
    \setlength{\tabcolsep}{3pt}
    \caption{\textbf{Completing hidden response coordinates.}
    Subject-macro averages over ten THINGS-EEG2 subjects and all 200 official
    test images, with matched $K400$ calibration.
    Resid.\ is training-ERP-residual Pearson; MSE uses native response units.
    ``Native'' retains all 250 visible time samples; ``pooled'' uses 25 bins.}
    \label{tab:supp-response-completion}
    \begin{tabular}{@{}lrrrr@{}}
        \toprule
        & \multicolumn{2}{c}{Random 16}
        & \multicolumn{2}{c}{Posterior 8} \\
        \cmidrule(lr){2-3}\cmidrule(lr){4-5}
        Readout & Resid.\ $\uparrow$ & MSE $\downarrow$
                & Resid.\ $\uparrow$ & MSE $\downarrow$ \\
        \midrule
        Geometric spline & .1810 & .02835 & .1281 & .05083 \\
        Native Ridge & .1274 & .02120 & .0552 & .02481 \\
        Source prior + native Ridge & .2154 & .02058 & .2842 & .02257 \\
        SENSA prior + native Ridge & \textbf{.2296} & .02044
                                  & .3348 & \textbf{.02153} \\
        SENSA prior + pooled Ridge & .2214 & \textbf{.02020}
                                  & \textbf{.3350} & .02154 \\
        \bottomrule
    \end{tabular}
\end{table}

\newpage
\section{Calibration-data scaling and stimulus-time equivalents}
\label{sec:supp-support-budget}

We vary the number of paired target images to assess how additional
calibration data improves image-specific prediction.
Budgets are $K\in\{10,25,50,100,200,400,800,1200,1600,2000\}$,
with three nested support draws for every held-out subject:
10 THINGS-EEG2, 4 THINGS-MEG, 20 Alljoined-1.6M, and 4 NSD subjects.
The source generator is fixed at seed 42; each budget refits the target
statistics and calibration using the same optimization and regularization
rules. All official test images are evaluated.
Response-only calibration retains the frozen image-dependent prediction
and replaces its mean with the support mean. NSD uses the source-anchored
spatial FT and Coordinate RM from \cref{tab:supp-nsd-factorial}, refitted
at every budget. Its response-only condition uses the unordered support
image set for latent centering without image--response correspondence.
We average the three draws within each subject before computing
subject-macro means and 95\% percentile bootstrap intervals
(10,000 subject resamples).
Thus these curves average support draws, whereas the main comparison
retains its fixed support draw.

\paragraph{Additional support improves individualized prediction.}
\Cref{fig:supp-support-eeg,fig:supp-support-meg,%
fig:supp-support-alljoined,fig:supp-support-nsd} show that the
frozen source representation continues to benefit from target data beyond
$K400$. Increasing support to $K2000$ raises residual correlation from
$.1079$ to $.1570$ on THINGS-EEG2, $.0943$ to $.1312$ on THINGS-MEG,
$.0404$ to $.0677$ on Alljoined, and $.2993$ to $.3390$ on NSD.
Every subject improves under this comparison.
On the sensor benchmarks, the mean contribution of RM over FT-only
also increases from $K400$ to $K2000$: $.0101$ to $.0204$ for EEG,
$.0016$ to $.0033$ for MEG, and $.0030$ to $.0079$ for Alljoined.
The curves therefore connect the calibration budget to both overall
image-specific accuracy and the benefit of residual correction, without
retraining the shared generator.
On NSD, FT-only and complete predictions coincide for $K\geq25$:
PRESS selects zero residual correction in every coordinate. Its scaling
curve therefore measures the benefit of native spatial calibration.

\paragraph{Stimulus-time interpretation.}
The upper axes translate paired-support sizes into stimulus-time equivalents
using the original protocols~\cite{gifford2022things,hebart2023things,
xu2025alljoined,allen2022nsd} and the repetitions used by calibration.
For THINGS-EEG2, four repetitions and a nominal 7.5\,s per
20-image sequence give $T(K)=K/40$ minutes; THINGS-MEG has one training
presentation per image and a mean 1.5\,s onset interval, yielding the
same conversion.
Alljoined uses each selected image's actual retained repetition count
and a nominal 7\,s per 20-image sequence; we average the resulting times
over the same subjects and support draws.
NSD averages three responses per image; 4\,s trial slots and the
scheduled null slots give $T(K)=0.24K$ minutes.
Consequently, $K400$ corresponds to approximately 10, 10, 9.4, and
96 minutes, respectively.
These equivalents describe calibration-data amounts under fixed
preprocessing. The time estimates exclude equipment setup,
between-block or session breaks, preprocessing, model fitting, and
separate catch/test trials. Alljoined additionally excludes rejected
recordings; NSD includes scheduled null slots.

\begin{table}[!htbp]
    \centering
    \tableformat
    \setlength{\tabcolsep}{3.5pt}
    \caption{\textbf{Stimulus-time equivalents of calibration data (minutes).}
    Conversions use the repetitions and protocol allowances described
    above. Alljoined times average actual retained repeats across
    subjects and support draws; NSD includes scheduled null slots.}
    \label{tab:supp-support-time}
    \begin{tabular*}{\columnwidth}{@{\extracolsep{\fill}}lrrrr@{}}
        \toprule
        Dataset & $K400$ & $K1200$ & $K1600$ & $K2000$ \\
        \midrule
        THINGS-EEG2 & 10.0 & 30.0 & 40.0 & 50.0 \\
        THINGS-MEG & 10.0 & 30.0 & 40.0 & 50.0 \\
        Alljoined-1.6M & 9.4 & 28.3 & 37.7 & 47.1 \\
        NSD & 96.0 & 288.0 & 384.0 & 480.0 \\
        \bottomrule
    \end{tabular*}
\end{table}

\clearpage
\begin{figure*}[!t]
    \begin{minipage}[t]{\columnwidth}
    \centering
    \includegraphics[width=\linewidth]{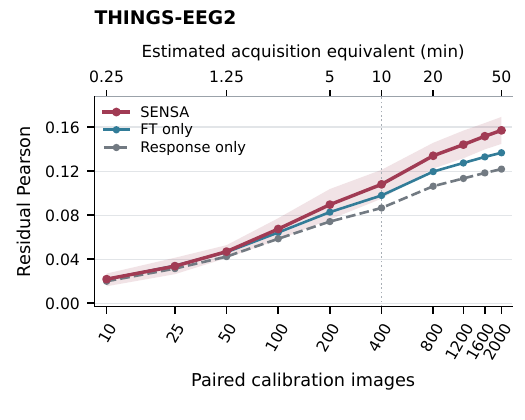}
    \caption{\textbf{THINGS-EEG2 calibration-data scaling.}
    Residual Pearson on all 200 official test images, averaged over
    ten held-out subjects after three support draws per subject.
    The upper axis gives stimulus-time equivalents with four repetitions;
    the dotted line marks $K400$ (approximately 10 minutes).
    Shading shows the 95\% subject-bootstrap interval for SENSA.}
    \label{fig:supp-support-eeg}
    \end{minipage}\hfill
    \begin{minipage}[t]{\columnwidth}
    \centering
    \includegraphics[width=\linewidth]{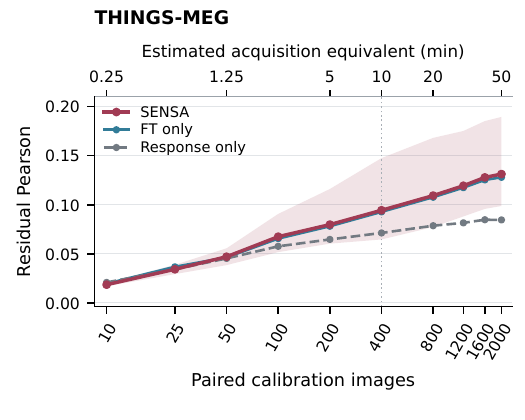}
    \caption{\textbf{THINGS-MEG calibration-data scaling.}
    Residual Pearson on all 200 official test images, averaged over
    all four held-out subjects and three support draws per subject.
    Stimulus-time equivalents use one training presentation per image
    and a mean 1.5\,s onset interval.
    Shading shows the 95\% subject-bootstrap interval for SENSA.}
    \label{fig:supp-support-meg}
    \end{minipage}
\end{figure*}

\begin{figure*}[!t]
    \begin{minipage}[t]{\columnwidth}
    \centering
    \includegraphics[width=\linewidth]{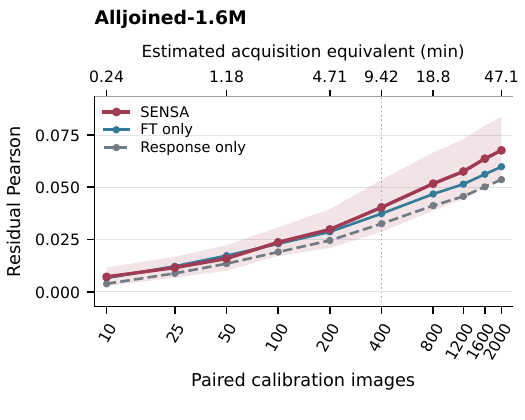}
    \caption{\textbf{Alljoined-1.6M calibration-data scaling.}
    Residual Pearson on all 200 official test images, averaged over
    all 20 held-out subjects and three support draws per subject.
    Stimulus-time equivalents use the actual retained repetitions for
    each selected image; rejected recordings are excluded.
    Shading shows the 95\% subject-bootstrap interval for SENSA.}
    \label{fig:supp-support-alljoined}
    \end{minipage}\hfill
    \begin{minipage}[t]{\columnwidth}
    \centering
    \includegraphics[width=\linewidth]{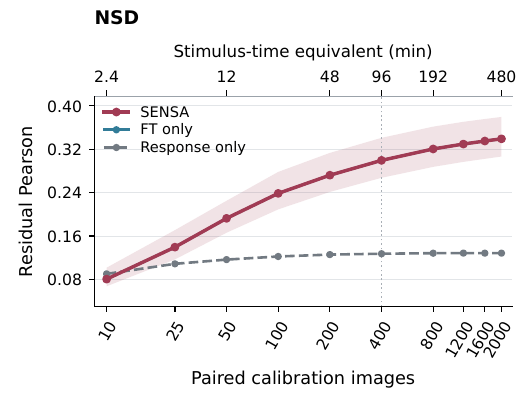}
    \caption{\textbf{NSD calibration-data scaling.}
    Native-voxel residual Pearson on all 1,000 official test images,
    averaged over subjects 1, 2, 5, and 7 after three support draws
    per subject. SENSA uses source-anchored spatial FT and Coordinate RM;
    FT-only and complete curves overlap for $K\geq25$. Response-only
    retains frozen source bases and the support mean.
    Stimulus-time equivalents include three repetitions and scheduled null slots.
    Shading shows the 95\% subject-bootstrap interval for SENSA.}
    \label{fig:supp-support-nsd}
    \end{minipage}
\end{figure*}

\clearpage
\section{Independent source-training stability}
\label{sec:supp-source-seeds}

To test whether individualized prediction depends on a particular source
initialization, we independently retrain both the source generator and
Response Refiner with seeds 42, 43, and 44 on all three
sensor benchmarks. Each seed covers all 10 EEG, 4 MEG, and 20 Alljoined
strict LOSO folds, giving 102 complete fits.
Source training uses the complete official
training partition with the held subject excluded.
Frozen visual features, training budgets, final-checkpoint rules,
the first $K400$ support draw, and target-calibration hyperparameters
remain fixed across seeds. MEG retains the amplitude learning rate
$.0002$. FT-off RM is refitted to its own upstream predictor, and
shuffled controls refit the entire calibration chain.
All conditions evaluate the same 200 official test images per subject.

\Cref{tab:supp-source-seed-full,tab:supp-source-seed-modules,%
tab:supp-source-seed-individual} report complete-model metrics,
matched module comparisons, and individual seed scores.
We first average subjects within a seed, then report the mean and
sample standard deviation of the three seed macro averages.
These repeated fits share the same subjects and test images; the
standard deviations quantify training randomness.
They are distinct from the support-set resampling in
\cref{fig:support-gains}.
These are single-model LOSO repeats; the within-subject EEG ensemble
is evaluated separately in \cref{tab:within}.

The complete pipeline improves residual correlation over response-only
calibration and FT-only in every seed on each sensor benchmark.
Correct pairing also exceeds the full-chain shuffled control in every
seed. On EEG, complete-model residual correlation spans
$.1091$--$.1096$, while Alljoined spans $.0437$--$.0445$.
The matched module table retains both single-module alternatives,
so training stability and module complementarity can be assessed
separately.

\paragraph{Paired module gains across source initializations.}
For subject-level comparisons, we first average each subject's scores over
the three source seeds, then bootstrap subjects 10,000 times
(seed 20261110). This keeps repeated fits of one subject together.
On EEG, FT+RM exceeds FT-only by $.0109$ (95\% interval
$[.0077,.0146]$) and RM-only by $.0059$ ($[.0047,.0072]$),
with gains in all ten subjects for both comparisons.
On Alljoined, the corresponding gains are $.0028$
($[.0010,.0048]$; 14/20 subjects) and $.0017$
($[.0006,.0029]$; 12/20).
MEG's differences are $.0019$ versus FT-only and $-.0006$ versus RM-only;
both intervals span zero. Thus EEG supports consistent module
complementarity, while Alljoined supports an average benefit with
individual variation. MEG's intervals do not establish a gain over either
single-module alternative.

\begin{table}[!htbp]
    \centering
    \tableformat
    \setlength{\tabcolsep}{2pt}
    \caption{\textbf{Complete SENSA across source-training seeds.}
    Mean $\pm$ sample SD across seeds 42--44 for every
    dataset, after subject-macro averaging within seed.
    EEG2, MEG, and Alljoined include 10, 4, and 20 subjects, respectively.
    Fixed $K400$ supports and all official
    test images; MSE retains each dataset's response units.}
    \label{tab:supp-source-seed-full}
    \begin{tabular*}{\columnwidth}{@{\extracolsep{\fill}}lrrr@{}}
        \toprule
        Dataset & Raw $\uparrow$ & Residual $\uparrow$ & MSE $\downarrow$ \\
        \midrule
        EEG2 & $.6509\!\pm\!.0001$ & $.1093\!\pm\!.0003$
             & $.021650\!\pm\!.000005$ \\
        MEG & $.4958\!\pm\!.0023$ & $.0803\!\pm\!.0019$
            & $.635543\!\pm\!.001988$ \\
        Alljoined & $.3400\!\pm\!.0004$ & $.0440\!\pm\!.0004$
                  & $.087891\!\pm\!.000020$ \\
        \bottomrule
    \end{tabular*}
\end{table}

\begin{table}[!htbp]
    \centering
    \tableformat
    \setlength{\tabcolsep}{2pt}
    \caption{\textbf{Calibration mechanisms across source seeds.}
    Residual Pearson, mean $\pm$ sample SD across seeds
    42--44, as in \cref{tab:supp-source-seed-full}.
    Each condition shares its seed's source weights, first $K400$
    supports, and support mean where available. FT-off memory is
    independently refitted; the shuffled row refits both FT and RM.
    Alljoined no-support uses the empirical source-training mean,
    matching \cref{tab:supp-final-information}.}
    \label{tab:supp-source-seed-modules}
    \begin{tabular*}{\columnwidth}{@{\extracolsep{\fill}}lrrr@{}}
        \toprule
        Condition & EEG2 & MEG & Alljoined \\
        \midrule
        No support & $.0375\!\pm\!.0002$ & $.0589\!\pm\!.0003$ & $.0127\!\pm\!.0001$ \\
        Response-only & $.0867\!\pm\!.0004$ & $.0593\!\pm\!.0001$ & $.0362\!\pm\!.0001$ \\
        FT only & $.0984\!\pm\!.0005$ & $.0784\!\pm\!.0020$ & $.0412\!\pm\!.0002$ \\
        RM only & $.1034\!\pm\!.0002$ & $.0809\!\pm\!.0012$ & $.0424\!\pm\!.0004$ \\
        FT+RM & $.1093\!\pm\!.0003$ & $.0803\!\pm\!.0019$ & $.0440\!\pm\!.0004$ \\
        Shuffled FT+RM & $.0265\!\pm\!.0009$ & $.0229\!\pm\!.0037$ & $.0186\!\pm\!.0004$ \\
        \bottomrule
    \end{tabular*}
\end{table}

\begin{table}[!htbp]
    \centering
    \tableformat
    \caption{\textbf{Complete-model residual correlation by seed.}
    Each entry averages the complete subject cohort with the same
    fixed $K400$ supports and source seeds 42--44.}
    \label{tab:supp-source-seed-individual}
    \begin{tabular*}{\columnwidth}{@{\extracolsep{\fill}}lrrr@{}}
        \toprule
        Source seed & EEG2 & MEG & Alljoined \\
        \midrule
        42 & .1091 & .0825 & .0438 \\
        43 & .1096 & .0790 & .0437 \\
        44 & .1093 & .0794 & .0445 \\
        \bottomrule
    \end{tabular*}
\end{table}

\section{Response coordinates and regularization controls}
\label{sec:supp-response-coordinate-controls}

\begin{table*}[!t]
    \centering
    \tableformat
    \caption{\textbf{Frozen-source readout controls at matched $K400$.}
    Full 10 EEG, 4 MEG and 20 Alljoined cohorts, source seeds 42--44,
    all official test images. Values average subject macros over the three
    seeds. Native denotes full channel--time output; $Q$ denotes fixed
    learned response coordinates. Centered is test-centered sample-wise
    Pearson, distinct from training-ERP residual Pearson.}
    \label{tab:supp-frozen-readouts}
    \begin{tabular*}{\textwidth}{@{\extracolsep{\fill}}llrrrr@{}}
        \toprule
        Dataset & Readout & Raw & Residual & Centered & MSE \\
        \midrule
        EEG2 & Native-response Ridge & .6393 & .0054 & .0104 & .022209 \\
             & FT only & .6497 & .0984 & .1560 & .021701 \\
             & FT + native-error Ridge & .6497 & .0984 & .1560 & .021701 \\
             & FT + $Q$-global Ridge & .6507 & .1044 & .1621 & .021656 \\
             & FT + Coordinate RM & .6509 & .1093 & .1651 & .021650 \\
        \midrule
        MEG & Native-response Ridge & .5205 & .0281 & .0439 & .594537 \\
            & FT only & .4957 & .0784 & .0997 & .635664 \\
            & FT + native-error Ridge & .4957 & .0784 & .0997 & .635664 \\
            & FT + $Q$-global Ridge & .4957 & .0784 & .0997 & .635664 \\
            & FT + Coordinate RM & .4958 & .0803 & .1014 & .635543 \\
        \midrule
        Alljoined-1.6M & Native-response Ridge & .3357 & .0053 & .0000 & .087970 \\
            & FT only & .3427 & .0412 & .0745 & .087709 \\
            & FT + native-error Ridge & .3427 & .0412 & .0745 & .087709 \\
            & FT + $Q$-global Ridge & .3424 & .0419 & .0740 & .087725 \\
            & FT + Coordinate RM & .3400 & .0440 & .0704 & .087891 \\
        \bottomrule
    \end{tabular*}
\end{table*}

\subsection{Matched readouts from frozen source features}

To distinguish response-space structure from regularization selection,
we reuse the source generators, Refiner states and paired FT fits from
all 102 folds in \cref{sec:supp-source-seeds}.
Each condition uses the same $K400$ indices, support mean and centered
image keys $\tanh(\bar z/4)$. No generator or FT parameters are retrained.
All fits precede evaluation on the complete 200-image official test.

\paragraph{Readout definitions.}
Native-response Ridge maps keys directly to centered target responses.
Native-error Ridge instead maps keys to remaining errors after the fixed
FT prediction, without a response subspace.
Each uses one global penalty chosen by exact, intercept-aware
leave-one-out PRESS over $\exp(\operatorname{linspace}(-5,5,21))$,
including an infinite-penalty zero predictor.
The $Q$-global control projects errors into the learned response coordinates
and selects one shared penalty. Coordinate RM selects penalties separately
for each coordinate. Both $Q$ controls retain the same $.05$ projection
stabilizer and null option; native-output fits use standard Ridge without
that projection stabilizer.
Thus the global-versus-coordinate $Q$ comparison isolates penalty
selection, whereas native-versus-$Q$ comparisons also change output
representation and projection regularization.

\paragraph{Results and aggregation.}
\Cref{tab:supp-frozen-readouts} reports means over the three source
seeds, with subjects equally weighted within each seed.
On EEG, Coordinate RM exceeds $Q$-global Ridge by $.0049$
(95\% subject-bootstrap interval $[.0035,.0064]$; 10/10 subjects).
On Alljoined, the difference is $.0021$ ($[.0008,.0037]$; 14/20);
MEG's $.0019$ difference has an interval spanning zero.
For these paired intervals, seeds are averaged within subject before
10,000 subject resamples (seed 20261110).
Native-error Ridge selects the zero correction in all 102 folds, retaining
FT-only predictions under this global selection rule.
These results support coordinate-wise regularization on EEG and Alljoined;
raw correlation, centered correlation and MSE retain their distinct trade-offs.

\subsection{Learned, PCA and random response coordinates}

We compare three 64-dimensional spans on all ten EEG subjects (source
seed 42), refitting identical Coordinate PRESS RM with fixed FT, keys and
$K400$ supports. PCA fits centered image-wise mean training responses
from the nine source subjects; target and test responses do not fit its
basis. Randomized PCA uses rank 80, four power iterations and seed
$730000+s$; the random orthonormal control uses seed $740000+s$.
Projection stabilization and penalty candidates match across conditions.

\Cref{tab:supp-pca-response-coordinates} shows residual $.1091$ for learned
$Q$, $.0892$ for PCA and $.0979$ for random coordinates.
Both comparisons favor learned coordinates in all ten subjects:
gains are $.0199$ over PCA (95\% interval $[.0161,.0237]$) and
$.0112$ over random ($[.0081,.0147]$).
The learned response span therefore supplies useful error coordinates
beyond leading source-response variance directions in this EEG protocol.

\begin{table}[!htbp]
    \centering
    \tableformat
    \setlength{\tabcolsep}{2pt}
    \caption{\textbf{Response-coordinate choice for EEG memory.}
    Fixed FT, source seed 42, $K400$, all ten subjects and official tests.
    Each RM span has 64 dimensions; PCA uses source training responses only.}
    \label{tab:supp-pca-response-coordinates}
    \begin{tabular*}{\columnwidth}{@{\extracolsep{\fill}}lrrrr@{}}
        \toprule
        Coordinates & Raw & Residual & Centered & MSE \\
        \midrule
        FT only & .6496 & .0979 & .1558 & .021705 \\
        PCA $Q$ & .6407 & .0892 & .1256 & .022192 \\
        Random $Q$ & .6496 & .0979 & .1553 & .021709 \\
        Learned $Q$ & .6508 & .1091 & .1649 & .021655 \\
        \bottomrule
    \end{tabular*}
\end{table}

\subsection{Rotating axes within the same response span}

To isolate coordinate orientation from subspace choice, we replace
$Q$ with $OQ$ using five fixed orthogonal matrices $O$ (seeds
811101--811105). All ten EEG subjects retain the same source seed 42,
$K400$ supports, support mean, image keys and frozen FT prediction.
Each readout is refitted with the same projection stabilizer, penalty
grid and null option; neither the generator nor FT is retrained.

\Cref{tab:supp-same-span-rotations} shows that Coordinate RM benefits from
the learned axes even when the response span is unchanged.
Its residual advantage over the five-rotation mean is $.0055$
(95\% subject-bootstrap interval $[.0043,.0068]$), positive in all ten
subjects. Rotations are averaged within subject before 10,000 subject
resamples (seed 20261111). Global PRESS remains invariant: decoded
corrections differ by at most $5.96{\times}10^{-7}$.
This control supports the alignment of learned response axes
with coordinate-wise shrinkage, beyond the choice of response subspace.

\begin{table}[!htbp]
    \centering
    \tableformat
    \caption{\textbf{Same-span rotation control for EEG memory.}
    Residual Pearson on all ten subjects and official tests.
    Orthogonal rotations preserve the 64-dimensional response span;
    only the coordinate axes change.}
    \label{tab:supp-same-span-rotations}
    \begin{tabular*}{\columnwidth}{@{\extracolsep{\fill}}lrr@{}}
        \toprule
        Coordinates & Coordinate PRESS & Global PRESS \\
        \midrule
        Learned $Q$ & .1091 & .1039 \\
        Rotation 1 & .1030 & .1039 \\
        Rotation 2 & .1041 & .1039 \\
        Rotation 3 & .1038 & .1039 \\
        Rotation 4 & .1036 & .1039 \\
        Rotation 5 & .1036 & .1039 \\
        \midrule
        Five-rotation mean & .1036 & .1039 \\
        \bottomrule
    \end{tabular*}
\end{table}

\clearpage
\clearpage
\clearpage
\clearpage
\clearpage

\section{Generation and calibration controls}
\label{sec:supp-additional-controls}

\subsection{Diffusion sample averaging}
\label{sec:supp-diffusion-averaging}

To determine whether calibration gains persist as stochastic prediction
variance is reduced, we average 1, 8, or 16 independently initialized
diffusion samples before fitting FT and Coordinate PRESS RM.
This experiment reuses the residual-diffusion replacement head in
\cref{tab:supp-generator-ft-rm}: all ten strict LOSO EEG folds retain
their seed-42 source checkpoints, trained for 5,000 updates on the
complete official source training set. Each image uses 20-step DDIM
with $\eta=0$. Sample sets are nested, and the first sample reuses the
original initialization. Generator weights remain frozen.

Every sample count uses the same 400 target-training images and support
mean. FT is fitted for 300 steps; RM is independently refitted with FT
off and on. Fully shuffled support refits the entire calibration chain.
All conditions are fitted and their predictions saved before scoring
the complete 200-image official test set. The no-module condition
retains response-only support statistics. Metrics follow
\cref{sec:supp-protocol}, including its fixed preprocessing.

\begin{table*}[!t]
\centering
\tableformat
\setlength{\tabcolsep}{2.5pt}
\caption{\textbf{Diffusion sample averaging and individualized calibration.}
(a) Ten-subject macro averages on THINGS-EEG2 with identical $K400$
support. Centered denotes test-image-centered Pearson; MSE uses five
decimal places. (b) Paired residual-correlation gains with subject-level
95\% bootstrap intervals from 10,000 resamples. Counts indicate subjects
with positive gains. The diffusion head is the matched replacement
generator in \cref{tab:supp-generator-ft-rm}.}
\label{tab:supp-diffusion-averaging}
\begin{tabular*}{\textwidth}{@{\extracolsep{\fill}}lcc*{8}{r}@{}}
\toprule
\multicolumn{11}{l}{\textit{(a) Complete metrics and support-pairing controls}} \\
& & & \multicolumn{4}{c}{Correct pairs}
& \multicolumn{4}{c}{Shuffled pairs} \\
\cmidrule(lr){4-7}\cmidrule(lr){8-11}
Samples & FT & RM & Raw & Resid. & Centered & MSE
& Raw & Resid. & Centered & MSE \\
\midrule
1 & $\times$ & $\times$
& .3575 & .0173 & .0185 & .06078 & .3575 & .0173 & .0185 & .06078 \\
& $\checkmark$ & $\times$
& .3604 & .0226 & .0241 & .06072 & .3536 & .0105 & .0112 & .06082 \\
& $\times$ & $\checkmark$
& .3600 & .0219 & .0233 & .06067 & .3502 & .0055 & .0059 & .06111 \\
& $\checkmark$ & $\checkmark$
& .3614 & \textbf{.0245} & .0260 & .06065 & .3497 & .0046 & .0050 & .06110 \\
\midrule
8 & $\times$ & $\times$
& .5594 & .0401 & .0491 & .02750 & .5594 & .0401 & .0491 & .02750 \\
& $\checkmark$ & $\times$
& .5620 & .0560 & .0673 & .02750 & .5563 & .0249 & .0313 & .02755 \\
& $\times$ & $\checkmark$
& .5620 & .0505 & .0610 & .02739 & .5512 & .0125 & .0163 & .02783 \\
& $\checkmark$ & $\checkmark$
& .5632 & \textbf{.0577} & .0693 & .02739 & .5508 & .0105 & .0139 & .02783 \\
\midrule
16 & $\times$ & $\times$
& .5915 & .0487 & .0652 & .02512 & .5915 & .0487 & .0652 & .02512 \\
& $\checkmark$ & $\times$
& .5935 & .0671 & .0867 & .02512 & .5889 & .0308 & .0431 & .02518 \\
& $\times$ & $\checkmark$
& .5940 & .0609 & .0799 & .02502 & .5836 & .0155 & .0226 & .02545 \\
& $\checkmark$ & $\checkmark$
& .5950 & \textbf{.0693} & .0898 & .02502 & .5832 & .0132 & .0196 & .02546 \\
\bottomrule
\end{tabular*}

\vspace{4pt}
\begin{tabular*}{\textwidth}{@{\extracolsep{\fill}}llrrr@{}}
\toprule
\multicolumn{5}{l}{\textit{(b) Subject-level calibration gains}} \\
Samples & Contrast & Mean gain & 95\% interval & Improved \\
\midrule
1 & FT+RM minus no modules & .0072 & [.0053, .0092] & 10/10 \\
  & RM added after FT & .0019 & [.0011, .0027] & 10/10 \\
  & FT added after RM & .0026 & [.0021, .0032] & 10/10 \\
\midrule
8 & FT+RM minus no modules & .0176 & [.0137, .0219] & 10/10 \\
  & RM added after FT & .0017 & [.0002, .0033] & 7/10 \\
  & FT added after RM & .0072 & [.0061, .0084] & 10/10 \\
\midrule
16 & FT+RM minus no modules & .0206 & [.0162, .0256] & 10/10 \\
   & RM added after FT & .0022 & [.0004, .0042] & 7/10 \\
   & FT added after RM & .0084 & [.0072, .0098] & 10/10 \\
\bottomrule
\end{tabular*}
\end{table*}

Sample averaging improves the uncalibrated diffusion prediction, while
FT and RM provide further image-dependent gains
(\cref{tab:supp-diffusion-averaging}). With 16 samples, residual
correlation increases from $.0487$ without either module to $.0693$
with both; the paired gain is $.0206$ with a 95\% interval
$[.0162,.0256]$, improving all ten subjects.
Adding RM after FT retains a positive mean gain at every sample count,
with intervals above zero; seven of ten subjects improve at 8 and
16 samples. The corresponding fully shuffled FT+RM score is $.0132$
at 16 samples. Thus, calibration gains persist after reducing sampling
variance and depend on the image--response correspondence in support.

\subsection{Structure under common waveform regularization}
\label{sec:supp-common-regularization}

To isolate the effect of response-bank structure from the placement
of temporal regularization, we compare separable and full waveform
banks with no temporal penalty or one common penalty applied to their
decoded bases. The within-subject study covers all ten EEG and four
MEG subjects with initialization seeds 42, 43, and 44.
It uses the factorized backbone without a Response Refiner, as in the
structure comparison in \cref{tab:core-mechanisms}.
Both banks retain 64 factors. Within each seed, visual projectors,
initial decoded waveforms, and minibatch order match.
The fixed budgets are five EEG epochs and eleven MEG epochs on the
complete official training sets, using final-epoch checkpoints and
all 200 official test images.

For the common penalty, each decoded basis is normalized by its
Frobenius norm before computing squared first and second temporal
differences. These are summed over sensors and averaged over bases
and valid time points, with coefficient $.001$ for both banks.
Response-bank weight decay is zero; visual-projector weight decay
is $.0001$, and the correlation-loss weight is $.05$ in all conditions.
This matches the training budget and regularization rule across banks
with different parameter counts. We average the three seeds within
each subject before computing macro scores and paired subject-bootstrap
intervals.

\begin{table*}[!t]
\centering
\tableformat
\setlength{\tabcolsep}{3pt}
\caption{\textbf{Response-bank structure under common temporal regularization.}
(a) Complete-test subject macro averages after averaging seeds 42/43/44
within each subject. Penalty applies the same temporal-difference
function to normalized decoded waveforms for both banks.
(b) Separable-minus-full residual-correlation gains with paired
subject-bootstrap 95\% intervals. Subjects, rather than seed--subject
combinations, are the sampling units.}
\label{tab:supp-common-regularization}
\begin{tabular*}{\textwidth}{@{\extracolsep{\fill}}lllrrrr@{}}
\toprule
\multicolumn{7}{l}{\textit{(a) Complete metrics}} \\
Dataset & Bank & Penalty & Raw & Resid. & Centered & MSE \\
\midrule
THINGS-EEG2 & Separable & None & .6773 & \textbf{.2013} & .2107 & .02033 \\
            & Full waveform & None & .6555 & .1634 & .1707 & .02146 \\
            & Separable & Common & .6773 & \textbf{.2016} & .2110 & .02033 \\
            & Full waveform & Common & .6555 & .1635 & .1707 & .02146 \\
\midrule
THINGS-MEG & Separable & None & .5618 & \textbf{.2038} & .2034 & .54504 \\
           & Full waveform & None & .5281 & .1718 & .1714 & .59833 \\
           & Separable & Common & .5618 & \textbf{.2038} & .2034 & .54504 \\
           & Full waveform & Common & .5281 & .1718 & .1714 & .59833 \\
\bottomrule
\end{tabular*}

\vspace{4pt}
\begin{tabular*}{\textwidth}{@{\extracolsep{\fill}}llrrr@{}}
\toprule
\multicolumn{5}{l}{\textit{(b) Subject-level structural gains}} \\
Dataset & Penalty & Mean gain & 95\% interval & Improved \\
\midrule
THINGS-EEG2 & None & .0379 & [.0339, .0425] & 10/10 \\
            & Common & .0381 & [.0341, .0428] & 10/10 \\
THINGS-MEG & None & .0320 & [.0217, .0429] & 4/4 \\
           & Common & .0320 & [.0217, .0428] & 4/4 \\
\bottomrule
\end{tabular*}
\end{table*}

The separable bank improves residual correlation in every subject
under either penalty setting (\cref{tab:supp-common-regularization}).
With common regularization, its EEG and MEG gains are $.0381$ and
$.0320$, with both subject-bootstrap intervals above zero.
Raw correlation, test-image-centered correlation, and MSE also favor
the separable bank. Applying the common penalty changes the scores
little relative to the structural gap, supporting a benefit from
separable response structure beyond the choice of temporal penalty.

\clearpage

\begingroup
\raggedbottom
\section{Additional comparisons with unified encoding--decoding models}
\label{sec:supp-neuroflow}

\paragraph{Image-to-response evaluation.}
We additionally evaluate the image-to-response generation of
NeuroFlow~\cite{mai2026neuroflow} and Mind-Omni~\cite{lu2026mindomni},
two unified encoding--decoding frameworks, using released weights.
For NeuroFlow, we use subject-specific Flow EMA and NeuroVAE checkpoints.
Inference uses the authors' released
CLIP image features, 20 forward Euler steps without Heun correction,
and seed 0. Released training-set voxel means and standard deviations
restore predictions to native response units. Generation receives image
features only, without measured test responses or additional calibration.
We evaluate all 1,000 official test images for each of subjects 1, 2, 5,
and 7, retaining every native voxel and averaging three measured repeats.
Raw Pearson, training-mean-subtracted residual Pearson, and MSE use the
same evaluator as \cref{tab:within}; macro averages weight subjects equally.

For Mind-Omni, image-only inference uses the released stage-1.2 model
and fMRI tokenizer with 64 sampling steps, guidance scale 9,
temperature $(2,0)$, float32 precision, and batches of 16.
The generator seed is reset to 42 for each batch.
No captions or measured brain responses enter generation.
The 16,127-voxel MNI predictions are mapped to each subject's native ROI
using the released anatomical registrations, C-order voxel indexing,
and trilinear interpolation. Per-voxel means and standard deviations
are matched using all 9,000 training images per subject, with three
native repeats averaged per image. Voxels outside the MNI mask's coverage
use the native training mean. Evaluation retains all native voxels.

\begin{table}[!htbp]
    \centering
    \tableformat
    \setlength{\tabcolsep}{3pt}
    \caption{\textbf{Additional NSD comparisons with unified
    encoding--decoding models.}
    All entries use 1,000 test images per subject, native voxels,
    and three-repeat-mean targets.
    SENSA is the model in \cref{tab:within}.
    Bold marks the best score within each subject or macro average.}
    \label{tab:supp-neuroflow}
    \begin{tabular*}{\columnwidth}{@{\extracolsep{\fill}}llrrr@{}}
        \toprule
        Subject & Method & Raw $\uparrow$ & Resid.\ $\uparrow$ & MSE $\downarrow$ \\
        \midrule
        S1 & NeuroFlow & \textbf{.8534} & \textbf{.3509} & \textbf{.0306} \\
           & Mind-Omni & .7504 & .1306 & .0575 \\
           & \method & .8519 & .3452 & .0307 \\
        \midrule
        S2 & NeuroFlow & \textbf{.8127} & \textbf{.3391} & .0528 \\
           & Mind-Omni & .6967 & .1265 & .0984 \\
           & \method & .8105 & .3333 & \textbf{.0527} \\
        \midrule
        S5 & NeuroFlow & .8670 & .3709 & .0733 \\
           & Mind-Omni & .7689 & .1334 & .1330 \\
           & \method & \textbf{.8715} & \textbf{.4138} & \textbf{.0688} \\
        \midrule
        S7 & NeuroFlow & .8212 & .2587 & .0515 \\
           & Mind-Omni & .7049 & .0959 & .0938 \\
           & \method & \textbf{.8262} & \textbf{.2995} & \textbf{.0505} \\
        \midrule
        Macro & NeuroFlow & .8386 & .3299 & .0520 \\
              & Mind-Omni & .7303 & .1216 & .0956 \\
              & \method & \textbf{.8400} & \textbf{.3480} & \textbf{.0507} \\
        \bottomrule
    \end{tabular*}
\end{table}

\paragraph{Comparison across subjects.}
\Cref{tab:supp-neuroflow} extends the generation evaluation to unified
encoding--decoding models. SENSA has the highest four-subject raw and
residual correlations and the lowest MSE among the three methods.
Its mean residual correlation is $.3480$, compared with $.3299$ for
NeuroFlow and $.1216$ for Mind-Omni. NeuroFlow has higher correlations
on S1 and S2, whereas SENSA is higher on S5 and S7.
\clearpage
\endgroup

\end{document}